\documentclass{article}

\usepackage{environ}
\newif\ifconferencemode
\NewEnviron{movable}[2]{%
  \ifconferencemode
    #2 
    \expandafter\xdef\csname stored#1\endcsname{\unexpanded\expandafter{\BODY}} 
  \else
    \BODY 
  \fi
}
\newcommand{\showstored}[1]{\csname stored#1\endcsname}

\newif\ifconferencemode
\conferencemodefalse

\ifconferencemode
    \newcommand{\NeurIPS}[1]{#1}
    \newcommand{\Arxiv}[1]{}
    \newcommand{\arxvFtnt}[1]{}
    \newcommand{\mainThmAsym}{thm:analBnd}
    \newcommand{\mainThmRec}{thm:analBnd}
\else
    \newcommand{\NeurIPS}[1]{}
    \newcommand{\Arxiv}[1]{#1}
    \newcommand{\arxvFtnt}[1]{\footnote{#1}}
    \newcommand{\mainThmAsym}{thm:asymBnd}
    \newcommand{\mainThmRec}{thm:recBnd}
\fi

\ifconferencemode

    \usepackage[final]{./style/NeurIPS_2025}

\else
\usepackage{natbib}
\usepackage[margin=1in]{geometry}
\usepackage{amssymb}
\usepackage{bbm}
\usepackage[textsize=tiny]{todonotes}
\fi

\usepackage[utf8]{inputenc} 
\usepackage[T1]{fontenc}    
\usepackage{hyperref}       
\usepackage{url}            
\usepackage{booktabs}       
\usepackage{amsfonts, amsthm, amsmath, mathtools}       
\usepackage{nicefrac}       
\usepackage{microtype}      
\usepackage{xcolor}         
\usepackage{graphicx}
\graphicspath{ {./plots/} }
\usepackage{float}
\usepackage{graphicx}
\usepackage{subcaption}

\theoremstyle{plain}
\newtheorem{theorem}{Theorem}[section]

\newtheorem{lemma}[theorem]{Lemma}
\newtheorem{claim}[theorem]{Claim}
\newtheorem{corollary}[theorem]{Corollary}
\theoremstyle{definition}
\newtheorem{definition}[theorem]{Definition}

\newtheorem{remark}[theorem]{Remark}

\newcommand{\eps}{\varepsilon}
\newcommand{\reals}{\mathbb{R}}
\newcommand{\naturals}{\mathbb{N}}
\newcommand{\eqExp}[2]{\overset{(#1)}{#2}}
\newcommand{\lvec}[1]{\reflectbox{$\vec{\reflectbox{$#1$}}$}}

\newcommand{\out}{y}
\newcommand{\outRV}{Y}
\newcommand{\outDom}{\mathcal{Y}}
\newcommand{\alg}{M}
\newcommand{\rand}{R}
\newcommand{\domain}{\mathcal{X}}
\newcommand{\dataset}{\boldsymbol{s}}
\newcommand{\view}{\boldsymbol{v}}
\newcommand{\viewRV}{\boldsymbol{V}}

\newcommand{\prob}[2]{\underset{#1}{\mathbb{P}} \left(#2 \right)}
\newcommand{\expect}[2]{\underset{#1}{\mathbb{E}} \left[#2 \right]}
\newcommand{\loss}[3]{\ell\left(#1; #2, #3 \right)}

\newcommand{\lossAlg}[4]{\ell_{#1}\left(#2; #3, #4 \right)}

\newcommand{\Renyi}[1]{\boldsymbol{R}_{#1}}
\newcommand{\RenyiDiv}[3]{\Renyi{#1}\left(#2 \Vert #3 \right)}
\newcommand{\HockeyStick}[1]{\boldsymbol{H}_{#1}}
\newcommand{\HockeyStickDiv}[3]{\HockeyStick{#1}\left(#2 ~\left\Vert~ #3 \right. \right)}

\newcommand{\Pois}[2]{\mathcal{P}_{#1}\left(#2\right)}
\newcommand{\PoisFunc}[3]{\mathcal{P}_{#1}\left(#2; #3\right)}
\newcommand{\shuf}[2]{\mathcal{S}_{#1}\left(#2\right)}

\newcommand{\alloc}[2]{\mathcal{A}_{#1}\left(#2\right)}
\newcommand{\allocFunc}[3]{\mathcal{A}_{#1}\left(#2; #3\right)}
\newcommand{\post}[2]{\mathcal{T}_{#1}\left(#2\right)}
\newcommand{\postFunc}[3]{\mathcal{T}_{#1}\left(#2; #3\right)}

\newcommand{\add}[1]{\lvec{#1}}
\newcommand{\rem}[1]{\vec{#1}}
\newcommand{\simAdd}{\mathrel{\substack{\sim\\[-0.8ex] \scalebox{0.6}{$\leftarrow$}}}}
\newcommand{\simRem}{\mathrel{\substack{\sim\\[-0.8ex] \scalebox{0.6}{$\rightarrow$}}}}
\newcommand{\privProf}[1]{\delta_{#1}}
\newcommand{\privProfAdd}[1]{\add{\delta}_{#1}}
\newcommand{\privProfRem}[1]{\rem{\delta}_{#1}}

\title{Privacy amplification by random allocation}

\ifconferencemode
\author{
    Vitaly Feldman \\
    Apple\\
    \texttt{vitaly.edu@gmail.com} \\
    \And
    Moshe Shenfeld\thanks{Work partially done while author was an intern at Apple} \\
    The Hebrew university of Jerusalem \\
    \texttt{moshe.shenfeld@mail.huji.ac.il} \\
}
\else
\author{
    Vitaly Feldman\\
    Apple \\
    \and
    Moshe Shenfeld\footnote{Work partially done while author was an intern at Apple.}\\
    The Hebrew University of Jerusalem
}
\fi

\begin{document}
\maketitle

\begin{abstract}
    We consider the privacy amplification properties of a sampling scheme in which a user's data is used in $k$ steps chosen randomly and uniformly from a sequence (or set) of $t$ steps. This sampling scheme has been recently applied in the context of differentially private optimization \citep{CGHLKKMSZ24, CCGHST25} and is also motivated by communication-efficient high-dimensional private aggregation \citep{AFKRT25}. Existing analyses of this scheme either rely on privacy amplification by shuffling which leads to overly conservative bounds or require Monte Carlo simulations that are computationally prohibitive in most practical scenarios.
    
    We give the first theoretical guarantees and numerical estimation algorithms for this sampling scheme. In particular, we demonstrate that the privacy guarantees of random $k$-out-of-$t$ allocation can be upper bounded by the privacy guarantees of the well-studied independent (or Poisson) subsampling in which each step uses the user's data with probability $(1+o(1))k/t$. Further, we provide two additional analysis techniques that lead to numerical improvements in several parameter regimes. Altogether, our bounds give efficiently-computable and nearly tight numerical results for random allocation applied to Gaussian noise addition.
\end{abstract}

\section{Introduction}
One of the central tools in the analysis of differentially private algorithms are so-called {\em privacy amplification} guarantees, where amplification results from sampling of the inputs. In these results one starts with a differentially private algorithms (or a sequence of such algorithms) and a randomized selection (or sampling) to determine which of the $n$ elements in a dataset to run each of the $t$ algorithms on. Importantly, the random bits of the sampling scheme and the selected data elements are not revealed. For a variety of sampling schemes this additional uncertainty is known to lead to improved privacy guarantees of the resulting algorithm, that is, privacy amplification. 

In the simpler, single step case a DP algorithm is run on a randomly chosen subset of the dataset. As first shown by \citet{KLNRS11}, if each element of the dataset is included in the subset with probability $\lambda$ (independently of other elements) then the privacy of the resulting algorithm is better (roughly) by a factor $\lambda$. This basic result has found numerous applications, most notably in the analysis of the differentially private stochastic gradient descent (DP-SGD) algorithm \citep{BST14}. In DP-SGD gradients are computed on randomly chosen batches of data points and then privatized through clipping and Gaussian noise addition. Privacy analysis of this algorithm is based on the so called Poisson sampling: elements in each batch and across batches are chosen randomly and independently of each other. The absence of dependence implies that the algorithm can be analyzed relatively easily as a direct composition of single step amplification results. The downside of this simplicity is that such sampling is less efficient and harder to implement within the standard ML pipelines. As a result, in practice some form of shuffling is used to define the batches in DP-SGD leading to a well-recognized discrepancy between the implementations of DP-SGD and their analysis \citep{CGKKMSZ24a, CGKKMSZ24b, ABDCH24}.

\begin{figure}[ht]
  \centering
  \includegraphics[width=1\linewidth]{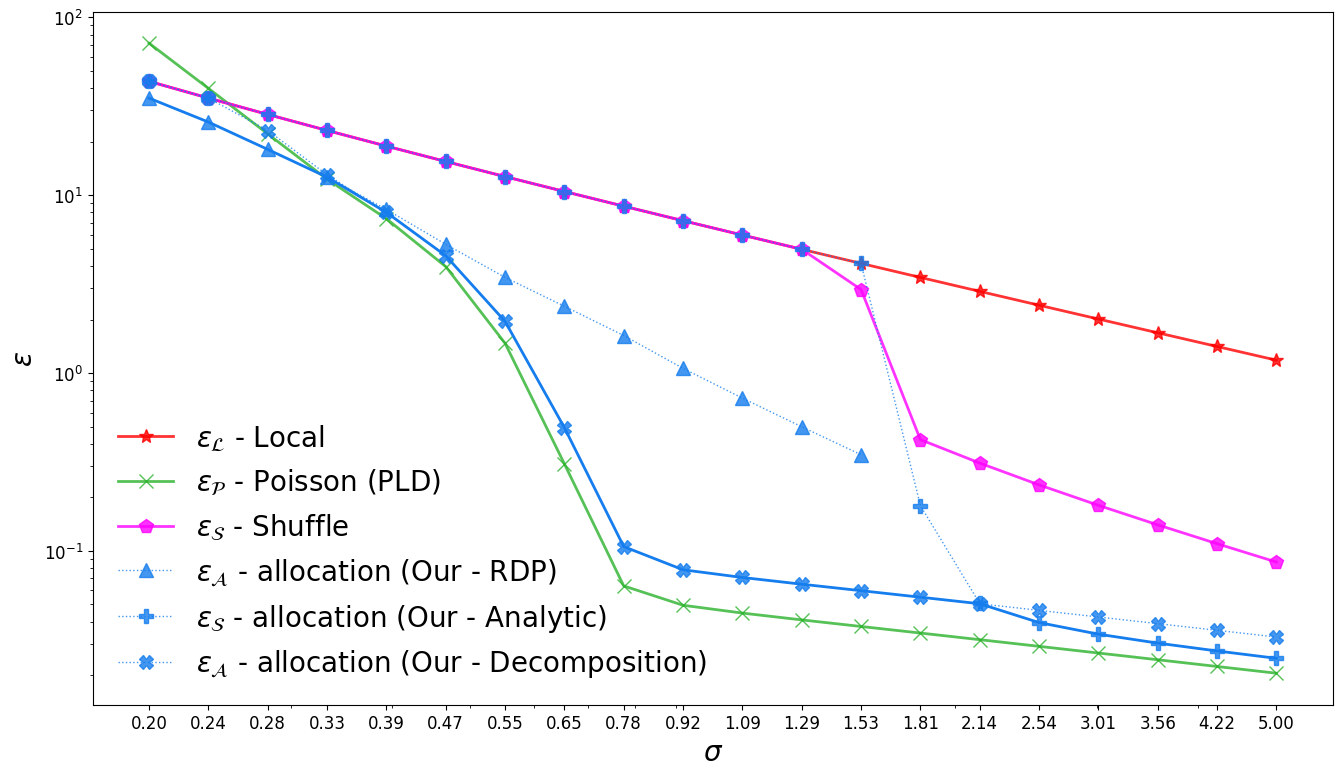}
  \caption{\small Upper bounds on privacy parameter $\eps$ as a function of the noise parameter $\sigma$ for various schemes and the local algorithm (no amplification), all using the Gaussian mechanism with fixed parameters $\delta = 10^{-10}$, $t = 10^{6}$. In the Poisson scheme $\lambda = 1/t$. The "flat" part of the RDP based calculation is due to computational limitations, which was computed for the range $\alpha \in[2, 60]$.}
  \label{fig:main}
\end{figure}

Motivated by the shuffle model of federated data analysis \citep{BEMMRLRKTS17}, \citet{CSUZZ19, EFMRTT19} have studied the privacy amplification of the shuffling scheme. In this scheme, the $n$ elements are randomly and uniformly permuted and the $i$-th element in the permuted order is used in the $i$-th step of the algorithm. This sampling scheme can be used to analyze the implementations of DP-SGD used in practice \citep{EFMRTT19, FMT21}. However, the analysis of this sampling scheme is more involved and nearly tight results are known only for relatively simple pure DP ($\delta =0$) algorithms \citep{FMT21,FMT23, GDDKS21}. 
In particular, applying these results to Gaussian noise addition requires using $(\eps,\delta)$-guarantees of the Gaussian noise. This leads to an additional $\sqrt{\ln(1/\delta)}$ factor in the asymptotic analysis and significantly worse numerical results (see Fig.~\ref{fig:main} for comparison and discussion in Section \ref{sec:asym}).

Note that shuffling differs from Poisson subsampling in that participation of elements is dependent both in each step (or batch) and across the steps. If the participation of elements in each step is dependent (by fixing the total number of participating elements) but the steps are independent then the sampling scheme can be tightly analyzed as a direct composition of fixed subset size sampling steps (e.g., using bound in \citet{BBG18,ZDW22}).
However, a more problematic aspect of Poisson sampling is the stochasticity in the number of times each element is used in all steps. For example, using Poisson sampling with sampling rate $1/t$ over $t$ batches will result in a roughly $1/e$ probability of not using the sample which implies dropping approximately $37\%$ of the data, and the additional sampling randomness may increase the resulting variance as demonstrated in Appendix \ref{apd:util}. In a distributed setting it is also often necessary to limit the maximum number of times a user participates in the analysis due to time or communication constraints on the protocol \citep{CSOK24,AFKRT25}. Poisson sampling does not allow to fully exploit the available limit which hurts the utility. 

Motivated by the privacy analysis of DP-SGD and the problem of communication-efficient high-dimensional private aggregation with two servers \citep{AFKRT25}, we analyze sampling schemes where each element participates in exactly $k$ randomly chosen steps out of the total $t$, independently of other elements. We refer to this sampling as $k$-out-of-$t$ {\em random allocation}.
For $k=1$, this scheme is a special case of the {\em random check-in} model of defining batches for DP-SGD in \citep{BKMT20}. Their analysis of this variant relies on the amplification properties of shuffling and thus does not lead to better privacy guarantees than those that are known for shuffling. Very recently, \citet{CGHLKKMSZ24} have studied such sampling (referring to it as {\em balls-and-bins sampling}) in the context of training neural networks via DP-SGD. Their main results show that from the point of view of utility (namely, accuracy of the final model) random allocation with $k=1$ is essentially identical to shuffling and is noticeably better than Poisson sampling. Concurrently, \citet{CCGHST25} considered the same sampling scheme for the matrix mechanism in the context of DP-FTRL. The privacy analysis in these two works reduces the problem to analyzing the divergence of a specific pair of distributions on $\reals^t$. They then used Monte Carlo simulations to estimate the privacy parameters of this pair. Their numerical results suggest that privacy guarantees of $1$-out-of-$t$ random allocation are similar to those of the Poisson sampling with rate of $1/t$. While very encouraging, such simulations have several limitations which we discuss in Appendix \ref{apd:monte-carlo}, most notably, achieving high-confidence estimates for small $\delta$ and supporting composition appear to be computationally impractical. This approach also does not lead to provable privacy guarantees and does not lend itself to asymptotic analysis (such as the scaling of the privacy guarantees with $t$).

\subsection{Our contribution}
We provide three new analyses for the random allocation setting that result in provable guarantees that nearly match or exceed those of the Poisson subsampling at rate $k/t$. The analyses rely on different techniques and lead to incomparable numerical results. We describe the specific results below and illustrate the resulting bounds in Fig.~\ref{fig:main}. 

In our main result we show that the privacy of random allocation is upper bounded by that of the Poisson scheme with sampling probability $\approx k/t$ up to lower order terms which are asymptotically vanishing in $t/k$. Specifically, we upper bound it by the $k$-wise composition of Poisson subsampling with rate $(1+\gamma) k/t$ applied to a dominating pair of distributions for each step of the original algorithm (Def.~\ref{def:domPair})  with an additional $t \delta_0 + \delta'$ added to the $\delta$ parameter. Here,  $\gamma=O\left(e^{\eps_0} \sqrt{\frac{k \ln(k/\delta')}{t}}\right)$ and $\eps_0,\delta_0$ are the privacy parameters of the original algorithm. The formal statement of this result that includes all the constants can be found in Thm.~\ref{\mainThmAsym}. Additionally, we show in Thm.~\ref{\mainThmRec} this lower order term can be recursively bounded using $(\eps',\delta')$ parameters of the same algorithm for some $\eps'>\eps$. This leads to significant numerical improvements in our results.

\Arxiv{
We note that our result scale with $\eps_0,\delta_0$ parameters of the original algorithm. This may appear to lead to the same overheads as the results based on full shuffling analysis. However in our case these parameters only affect the lower order term, whereas for shuffling they are used as the basis for privacy amplification (Lemma \ref{lem:GaussAsymBnd}).
}

Our analysis relies on several simplification steps. Using a dominating pair of distributions for the steps of the original algorithm, we first derive an explicit dominating pair of distributions for random allocation (extending a similar result for Gaussian noise in \citep{CGHLKKMSZ24,CCGHST25}). Equivalently, we reduce the allocation for general multi-step adaptive algorithms to the analysis of random allocation for a single (non-adaptive) randomizer on two inputs.
We also analyze only the case of $k=1$ and then use a reduction from general $k$ to $k=1$.
Finally, our analysis of the non-adaptive randomizer for $k=1$ relies on a decomposition of the allocation scheme into a sequence of posterior sampling steps for which we then prove a high-probability bound on subsampling probability in each step.

We note that, in general, the privacy of the composition of subsampling of the dominating pair of distributions can be worse than the privacy of the sampling scheme of a concrete algorithm, even if this pair tightly dominates it. This is true for both Poisson and random allocation schemes. However, all existing analyses of the Poisson sampling are effectively based on composition of subsampling for a dominating pair of distributions. Moreover, if the algorithm has a pair of neighboring datasets inducing this dominating pair, then our upper bound can be stated directly in terms of the Poisson subsampling scheme with respect to this pair. Such {\em dominating input} exists for many standard algorithms including those based on Gaussian and Laplace noise addition.

While our result shows asymptotic equivalence of allocation and Poisson subsampling, it may lead to suboptimal bounds for small values of $t/k$ and large $\eps_0$. We address this using two additional techniques which are also useful as starting points for the recursive version of our main result.

We first show that $\eps$ of random allocation with $k=1$ is at most a constant ($\approx 1.6$) factor times larger than $\eps$ of the Poisson sampling with rate $1/t$ for the same $\delta$ (see Theorem \ref{thm:dcmpBnd}). This upper bound does not asymptotically approach Poisson subsampling but applies in all parameter regimes. To prove this upper bound  we observe that Poisson subsampling is essentially a mixture of random allocation schemes with various values of $k$. We then prove a monotonicity property of random allocations showing that increasing $k$ leads to worse privacy. Combining these results with the advanced joint convexity property \cite{BBG18} gives the upper bound.

Finally, we give a direct analysis of the divergence for the dominating pair of distributions. Due to the asymmetric nature of the add/remove privacy our bounds require different techniques for each of the directions. In the remove direction we derive a closed form expression for the R\'{e}nyi DP \citep{Mironov17} of the dominating pair of distributions for allocation in terms of the RDP parameters of the original algorithm (Theorem \ref{thm:dirRemBnd}). This method has two important advantages. First, it gives a precise bound on the RDP parameters of integer order (as opposed to just an upper bound). Secondly, it is particularly easy to use in the typical setting where composition is used in addition to a sampling scheme (for example, when $k>1$ or in multi-epoch DP-SGD). The primary disadvantage of this technique is that the conversion from RDP bounds to the regular $(\eps,\delta)$ bounds is known to be somewhat lossy (typically within $10$-$20\%$ range in multi-epoch settings). The same loss is also incurred when Poisson sampling is analyzed via RDP (referred to as moment accounting \citep{ACGMMT16}). Two more limitations of this technique result from the restriction to the range $\alpha \ge 2$, and the computational complexity when $\alpha$ is in the high tens.

For the add direction we give an approximate upper bound in terms of the usual composition of a different, explicitly defined randomizer over the same domain. While this bound is approximate, the divergence for the add direction is typically significantly lower than the one for the remove direction and therefore, in our evaluations, this approximation has either minor or no effect on the maximum.  
Overall, in our evaluations of this method for Gaussian distribution in most regimes the resulting bounds are almost indistinguishable from those obtained via RDP for Poisson distribution (see Fig.~\ref{fig:multi-epoch} for examples). In fact, in some regimes
it is better than Poisson sampling (Figure \ref{fig:RDP-dom}). 

\noindent{\bf Numerical evaluation:} In Section \ref{sec:numEval} we provide numerical evaluation and comparisons of our bounds to those for Poisson sampling as well as other relevant bounds.\footnote{Python implementation of all methods is available in a \href{https://github.com/moshenfeld/random_allocation.git}{GitHub project} and in \href{https://pypi.org/project/random-allocation/}{a package}.} Our evaluations across many parameter regimes give bounds on the privacy of random allocation that are very close, typically within 10\% of those for the Poisson subsampling with rate $k/t$. This means that random allocation can be used to replace Poisson subsampling with only a minor loss in privacy. At the same time, in many cases, the use of random allocation can improve utility. In the context of training neural networks via DP-SGD this was shown in \citep{CGHLKKMSZ24}. Application of our bounds also lead to improvement over Poisson subsampling in \citep{AFKRT25}. We demonstrate that even disregarding some practical disadvantages of Poisson subsampling, random allocation has a better privacy-utility trade-off for mean estimation in low-dimensional regime. This improvement stems from the fact that random allocation computes the sum exactly whereas Poisson subsampling introduces additional variance. At the same time in the high-dimensional regime noise due to privacy dominates the final error and thus the trade-off boils down to the difference in the privacy bounds.

\subsection{Related work}
Our work builds heavily on tools and ideas developed for analysis of privacy amplification by subsampling, composition and shuffling. We have covered the work directly related to ours earlier and will describe some of the tools and their origins in the preliminaries. A more detailed technical and historical overview of subsampling and composition for DP can be found in the survey by \citet{Steinke22}.  The shuffle model was first proposed by \citet{BEMMRLRKTS17}. The formal analysis of the privacy guarantees in this model was initiated in \citep{EFMRTT19, CSUZZ19}. The sequential shuffling scheme that we discuss here was defined by \citet{EFMRTT19} who proved the first general privacy amplification results for this scheme albeit only for pure DP algorithms. Improved analyses and extensions to approximate DP were given in \citep{BBGN19, BKMT20, FMT21, FMT23, GDDKS21, KHH22}.

DP-SGD was first defined and theoretically analyzed in the convex setting by \citet{BST14}. Its use in machine learning was spearheaded by the landmark work of 
\citet{ACGMMT16} who significantly improved the privacy analysis via the moments accounting technique and demonstrated the practical utility of the approach. In addition to a wide range of practical applications, this work  has motivated the development of more advanced techniques for analysis of sampling and composition. At the same time most analyses used in practice still assume Poisson subsampling when selecting batches whereas some type of shuffling is used in implementation. It was recently shown that it results in an actual difference between the reported and true privacy level in some regimes \citep{CGKKMSZ24a, CGKKMSZ24b, ABDCH24}.

In a concurrent and independent work \citet{DCO25} considered the same sampling method (referring to it as \emph{Balanced Iteration Subsampling}). Their results are closest in spirit to our direct bounds. Specifically, they provide RDP-based bounds for the same dominating pair of distributions in the Gaussian case for both add and remove directions. Their bound for general $k$ is incomparable to ours as it is based on a potentially loose upper bound for divergences of order $\alpha > 2$, while using an exact extension of their approximation to $k > 1$. In contrast, our RDP-based bound uses a reduction from general $k$ to $k=1$ that is potentially loose but our computation for the $k=1$ case is exact (for the remove direction which is typically larger than the add direction). In our numerical comparisons, the bounds in \cite{DCO25} are comparable or worse than our direct bounds and are often significantly worse than the bounds from our main result. We discuss these differences in more detail and provide numerical comparison in Appendix \ref{apd:comp_loose_RDP}.

\section{Preliminaries}
We denote the domain of \emph{elements} by $\domain$ and the set of possible \emph{outputs} by $\outDom$.
Given a dataset $\dataset \in \domain^{*}$ and an output $\out \in \outDom$, we denote by $P_{\alg}(\out \vert \dataset) \coloneqq \prob{\outRV \sim \alg(\dataset)}{\outRV = \out}$ the probability of observing the output $\out$ as the output of some randomized algorithm $\alg$ which was given dataset $\dataset$ as input.\arxvFtnt{In case of measurable spaces, this quantity represents the probability density function rather than the probability mass function} \Arxiv{We omit the subscript when the algorithm is clear from the context.}

\subsection{Sampling schemes}\label{sec:schm}
In this work, we consider \emph{$t$-step algorithms} defined using an algorithm $\alg$ that takes some subset of the dataset and a sequence of previous outputs as input. Formally, let $\outDom^{<t}=\bigcup_{i<t} \outDom^{i}$. $\alg$ takes a dataset in $\domain^{*}$ and a view $\view\in\outDom^{<t}$ as its inputs and outputs a value in $\outDom$. A $t$-step algorithm defined by $\alg$ first uses some scheme to define $t$ subsets $\dataset^{1}, \ldots, \dataset^{t} \subseteq \dataset$, then sequentially computes $\out_{i} = \alg \left(\dataset^{i}, \view^{i - 1} \right)$, where $\view^{i} \coloneqq (\out_{1}, \ldots, \out_{i})$ are the intermediate \emph{views} consisting of the outputs produced so far, and $\view^{0} = \emptyset$.
Such algorithms include DP-SGD, where each step consists of a call to the Gaussian mechanism (\ref{lem:gaussPriv}), with gradient vectors adaptively defined as a function of previous outputs.

The assignment of the elements in $\dataset$ to the various subsets can be done in a deterministic manner (e.g., $\dataset^{1} = \ldots = \dataset^{t} = \dataset)$, or randomly using a \emph{sampling scheme}. We consider the following three sampling schemes.
\begin{enumerate}
    \item 
    \emph{Poisson scheme} parametrized by sampling probability $\lambda \in [0, 1]$, where each element is added to each subset $\dataset^{i}$ with probability $\lambda$ independent of the other elements and other subsets,
    \item \emph{Shuffling scheme} which uniformly samples a permutation $\pi$ over $[n]$ where $n$ is the sample size, and sets $\dataset^{i} = \{x_{\pi(i)}\}$ (in this case, the sample size and number of steps must match).
   \item \emph{Random allocation scheme} parameterized by a number of selected steps $k \in [t]$, which uniformly samples $k$ indices $\boldsymbol{i} = (i_{1}, \ldots, i_{k}) \subseteq [t]$ for each element and adds them to the corresponding subsets $\dataset^{i_{1}}, \ldots, \dataset^{i_{k}}$.
\end{enumerate}

For a $t$-step algorithm defined by $\alg$, we denote by $\Pois{t, \lambda}{\alg} : \domain^{*} \rightarrow \outDom^{t}$ an algorithm using $\alg$ with the Poisson sampling scheme, $\shuf{n}{\alg} : \domain^{n} \rightarrow \outDom^{n}$ for the shuffling scheme, and $\alloc{t, k}{\alg} : \domain^{*} \rightarrow \outDom^{t}$ when $\alg$ is used with the random allocation scheme. When $k = 1$ we omit it from the notation for clarity.

\subsection{Privacy notions}
We consider the standard add/remove adjacency notion of privacy in which datasets $\dataset, \dataset' \in \domain^{*}$ are neighboring if $\dataset$ can be obtained from $\dataset'$ via adding or removing one of the elements. To appropriately define sampling schemes that operate over a fixed number of elements we augment the domain with a ``null'' element $\bot$, that is, we define $\domain' \coloneqq \domain \cup \{\bot\}$.
When a $t$-step algorithm assigns $\bot$ to $\alg$ we treat it as an empty set, that is, for any $\dataset \in \domain^{*}$, $\view \in \outDom^{*}$ we have $\alg(\dataset, \view) = \alg((\dataset, \bot), \view)$.
We say that two datasets $\dataset, \dataset' \in \domain^{n}$ are \emph{neighbors} and denote it by $\dataset \simeq \dataset'$, if one of the two can be created by replacing a single element in the other dataset by $\bot$.

We rely on the hockey-stick divergence to quantify the privacy loss. 
\begin{definition}\Arxiv{[Hockey-stick divergence \cite{BKOZB12}]}\label{def:HSdiv}
    Given $\kappa \ge 0$ and two distributions $P, Q$ over some domain $\Omega$, the \emph{hockey-stick divergence} between them is defined to be
    \[
        \HockeyStickDiv{\kappa}{P}{Q} \coloneqq \int_{\Omega} \left[P(\omega) - \kappa Q(\omega) \right]_{+} d\omega = \expect{\omega \sim Q}{\left[e^{\loss{\omega}{P}{Q}} - \kappa \right]_{+}} = \expect{\omega \sim P}{\left[1 - \kappa e^{\loss{\omega}{Q}{P}} \right]_{+}},
    \]
    where $\loss{\omega}{P}{Q} \coloneqq \ln \left(\frac{P(\omega)}{Q(\omega)}\right)$; $\frac{P(\omega)}{Q(\omega)}$ is the ratio of the probabilities for countable domain or the Radon Nikodym derivative in the continuous case, and $\left[x \right]_{+} \coloneqq \max\{0, x\}$.\arxvFtnt{Despite its name, the hockey-stick divergence is actually not a true divergence under the common definition, since it does not satisfy part of the positivity condition which requires that the divergence is equal to $0$ only for two distributions that are identical almost everywhere, because it is not strictly convex at $1$. This has no effect on our results, since we don't use any claim that is based on this property of divergences.}
    When $P, Q$ are distributions induced by neighboring datasets $\dataset \simeq \dataset'$ and an algorithm $\alg$, we refer to the log probability ratio as the \emph{privacy loss random variable} and denote it by $\lossAlg{\alg}{\out}{\dataset}{\dataset'}$. We omit $\alg$ from the notation when it is clear from the context.
\end{definition}

\begin{definition}[\Arxiv{Privacy profile }\citep{BBG18}]\label{def:privProf}
    Given an algorithm $\alg : \domain^{*} \rightarrow \outDom$, the privacy profile $\privProf{\alg} : \reals \rightarrow [0, 1]$ is defined to be the maximal hockey-stick divergence between the distributions induced by any neighboring datasets and past view. Formally, $\privProf{\alg}(\eps) \coloneqq \underset{\dataset \simeq \dataset' \in \domain^{*}, \view \in \outDom^{*}}{\sup} \left( \HockeyStickDiv{e^{\eps}}{\alg(\dataset, \view) }{ \alg(\dataset', \view)} \right)$.
    
    Since the hockey-stick divergence is asymmetric in the general case, we use $\privProfRem{\alg}$ and $\simRem$ to denote the \emph{remove} direction where $\bot \in \dataset'$ and $\privProfAdd{\alg}$, $\simAdd$ to denote the \emph{add} direction when $\bot \in \dataset$. Notice that $\privProf{\alg}(\eps) = \max\{\privProfRem{\alg}(\eps), \privProfAdd{\alg}(\eps) \}$. 
\end{definition}

Another useful divergence notion is the \emph{R\'{e}nyi divergence}.
\begin{definition}\Arxiv{[R\'{e}nyi divergence]}\label{def:renDiv}
    Given $\alpha > 1$ and two distributions $P, Q$ over some domain $\Omega$, the \emph{R\'{e}nyi divergence} between them is defined to be $\RenyiDiv{\alpha}{P}{Q} \coloneqq \frac{1}{\alpha-1} \ln \left(\expect{\omega  \sim Q}{e^{\alpha \cdot \loss{\omega}{P}{Q}}} \right)$.\arxvFtnt{The cases of $\alpha = 1$ and $\alpha = \infty$ are defined by continuity which results in $\Renyi{1} = \boldsymbol{D}_{KL}$ - the KL divergence, and $\Renyi{\infty} = \boldsymbol{D}_{\infty}$ - the max divergence.}
\end{definition}

\begin{movable}{mov:RDPimpDP}{}
Since R\'{e}nyi divergence is effectively a bound on the moment generating function, it can be used to bound the hockey-stick divergence which is effectively a tail bound.
\begin{lemma}[R\'{e}nyi bounds Hockey-stick, Prop. 12 in \citet{CKS20}]\label{lem:RDPtoDP}
    Given two distributions $P, Q$, if $\RenyiDiv{\alpha}{P}{Q} \le \rho$ then $\HockeyStickDiv{e^{\eps}}{P}{Q} \le \frac{1}{\alpha-1}e^{(\alpha-1)(\rho - \eps)}\left(1 - \frac{1}{\alpha} \right)^{\alpha}$.
\end{lemma}
\end{movable}

We can now formally define our privacy notions.

\begin{definition}[\Arxiv{Differential privacy }\citep{DKMMN06}]
    Given $\eps > 0$; $\delta \in [0, 1 ]$, an algorithm $\alg$ will be called \emph{$(\eps, \delta)$-differentially private (DP)}, if $\privProf{\alg}(\eps) \le \delta$.
\end{definition}

\begin{definition}[\Arxiv{R\'{e}nyi differential privacy }\citep{Mironov17}]
    Given $\alpha \ge 1$; $\rho > 0$, an algorithm $\alg$ will be called \emph{$(\alpha, \rho)$-R\'{e}nyi differentially private (RDP)}, if $\underset{\dataset \simeq \dataset' \in \domain^{*}, \view \in \outDom^{*}}{\sup} \left( \RenyiDiv{\alpha}{\alg(\dataset, \view)}{\alg(\dataset', \view)} \right) \le \rho$.
\end{definition}

One of the most common algorithms is the Gaussian mechanism $N_{\sigma}$, which simply reports the sum of (some function of) the elements in the dataset with the addition of Gaussian noise.
One of its main advantages  is that we have closed form expressions of its privacy
\begin{movable}{mov:GaussMech}{(Lemma \ref{lem:gaussPriv}).}
.

\begin{lemma}[Gaussian mechanism DP guarantees, \citep{BW18,Mironov17}] \label{lem:gaussPriv}
     Given $d \in \naturals$; $\sigma > 0$, let $\domain = \outDom \coloneqq \reals^{d}$. The \emph{Gaussian mechanism} $N_{\sigma}$ is defined as $N_{\sigma}(\dataset) \coloneqq \mathcal{N}(\sum_{x \in \dataset} x, \sigma^{2} I_{d})$.

    If the domain of $N_{\sigma}$ is the unit ball in $\reals^{d}$, we have $\privProf{N_{\sigma}}(\eps) = \Phi \left(\frac{1}{2 \sigma} - \eps \sigma \right) - e^{\eps} \Phi \left(-\frac{1}{2 \sigma} - \eps \sigma \right)$, where $\Phi$ is the CDF of the standard Normal distribution, and for any $\alpha \ge 1$ $N_{\sigma}$ is $(\alpha, \alpha / (2 \sigma^{2})$-RDP.
\end{lemma}
\end{movable}

\subsection{Dominating pair of distributions}
A key concept for characterizing the privacy guarantees of an algorithm is that of a \emph{dominating pair} of distributions. 
\begin{definition}[\Arxiv{Dominating pair }\citep{ZDW22}] \label{def:domPair}
    Given distributions $P, Q$ over some domain $\Omega$, and $P', Q'$ over $\Omega'$, we say $(P', Q')$ \emph{dominate} $(P, Q)$ if for all $\kappa \ge 0$ we have $\HockeyStickDiv{\kappa}{P}{Q} \le \HockeyStickDiv{\kappa}{P'}{Q'}$.\arxvFtnt{The $\alpha \in [0, 1]$ regime does not correspond to useful values of $\eps$, but yet is crucial for the following guarantees, as demonstrated by \citet{LRKS24}.} If $\privProf{\alg}(\eps) \le \HockeyStickDiv{e^{\eps}}{P}{Q}$ for all $\eps \in \reals$, we say $(P, Q)$ is a \emph{dominating pair} of distributions for $\alg$. If the inequality can be replaced by an equality for all $\eps$, we say it is a \emph{tightly dominating pair}.
    If there exist some $\dataset \simeq \dataset' \in \domain^{*}$ such that $P = \alg(\dataset)$, $Q = \alg(\dataset')$ we say $(\dataset, \dataset')$ are the dominating pair of datasets for $\alg$. By definition, a dominating pair of input datasets is tightly dominating. \end{definition}

\Arxiv{
\citet{ZDW22} provide several useful properties of dominating pairs; A tightly dominating pair $(P, Q)$ always exists (Proposition 8), if $(P, Q)$ dominate $\privProfRem{\alg}$, then $(Q, P)$ dominate $\privProfAdd{\alg}$, which implies either $(P, Q)$ or $(Q, P)$ dominate $\privProf{\alg}$ (Lemma 28), and if $(P'_{i}, Q'_{i})$ dominates $(P_{i}, Q_{i})$ for $i \in \{1, 2\}$ then $(P'_{1} \times P'_{2}, Q'_{1} \times Q'_{2})$ dominates $(P_{1} \times P_{2}, Q_{1} \times Q_{2})$, where $\times$ denotes composition of the two distributions. This last result holds even when the $P_{2}, Q_{2}$, etc. are conditioned on the output of the first distribution (Theorem 10).
}
We use the notion of dominating pair to define a dominating randomizer, which captures the privacy guarantees of the algorithm independently of its algorithmic adaptive properties. 
\begin{definition}\Arxiv{[Dominating randomizer]}\label{def:domRand}
    Given a $t$-step algorithm defined by $\alg$, we say that $\rand : \{\bot, *\} \rightarrow \outDom$ is a \emph{dominating randomizer} for $\alg$ and set $\rand(*) = P$ and $\rand(\bot) = Q$, where $(P, Q)$ is the tightly dominating pair of $\alg(\cdot,\cdot)$ w.r.t. $\simRem$ over all indexes $i \in [t]$ and input partial views $\view^{i-1}$.\NeurIPS{\footnote{Such a pair always exists \citep[Proposition 8]{ZDW22}}.}
\end{definition}

The definition of the Poisson and random allocation schemes naturally extends to the case where the internal algorithm is a randomizer. In this case $\Pois{t, \lambda}{\rand} : \{*, \bot\} \rightarrow \outDom^{t}$ and $\alloc{t, \lambda}{\rand} : \{*, \bot\} \rightarrow \outDom^{t}$.

\Arxiv{
From the definition, the privacy profile of $\alg$ is upper bounded by that of the $\rand$, and equality is achieved only if $\alg$ has a dominating pair of datasets (for some view). When it comes to sampling schemes, it might be the case that even if $\alg$ has a dominating pair of datasets, this pair does not dominate the Poisson or allocation schemes defined by this algorithm, and in fact such pair might not exist. For example, while the Gaussian mechanism is dominated by the pair $(1, 0)$ (Claim \ref{clm:domPairGauss}), a pair of datasets dominate the DP-SGD algorithm \citep{ACGMMT16} which is essentially a Poisson scheme using the Gaussian mechanism, only if the norm of the difference between the gradients they induce exceeds the clipping radius at all iterations. Since known state-of-the-art bounds rely on the properties of the randomizer rather than leveraging the properties of the specific algorithm, this gap is not specific to our privacy bounds.
}

\begin{movable}{mov:postProc}{}
An important property of domination is its equivalence to existence of postprocessing.
\begin{lemma}[Post processing, Thm.~II.5 \citep{KOV15}]\label{lem:domPostProc}
    Given distributions $P, Q$ over some domain $\Omega$, and $P', Q'$ over $\Omega'$, $(P, Q)$ dominate $(P', Q')$ if and only if there exists a randomized function $\varphi : \Omega \rightarrow \Omega'$ such that $P' = \varphi(P)$ and $Q' = \varphi(Q)$.
\end{lemma}
\end{movable}

\Arxiv{
\begin{remark}
We note that an alternative way to frame our results is using the local randomizer perspective used in the privacy analysis of shuffling (e.g.~\citep{EFMRTT19}). In this perspective, the local randomizer $\rand$ is fixed first and the goal is to analyze the privacy of the sequence of $t$ applications of $\rand$, where the given data element $x$ is used as an input in randomly chosen $k$ steps and $\bot$ is used as an input in all the other steps (with view being an additional input in the adaptive case). Our analysis corresponds more naturally to this local perspective. The definition of the dominating randomizer effectively allows us to reduce the central setting to the local one.
\end{remark}
}

\section{General reduction} \label{sec:renRed}
We first prove two general claims which reduce the bound on arbitrary algorithms, datasets, and number of allocations, to the case of a single allocation ($k=1$) of a simple non-adaptive randomizer receiving a single element. Missing proofs can be found in Appendix \ref{apd:renRed}

From the definition of the dominating randomizer, for any $\eps \in \reals$ we have $\privProf{\alg}(\eps) \le \privProf{\rand}(\eps)$. We now prove that this is also the case for allocation scheme, that is $\privProf{\alloc{t, k}{\alg}}(\eps) \le \privProf{\alloc{t, k}{\rand}}(\eps)$, and that the supremum over neighboring datasets for $\alloc{t, k}{\rand}$ is achieved by the pair of datasets containing a single element, that is $\dataset = \{*\}$, $\dataset' = \{\bot\}$.
This results from the fact that random allocation can be viewed as a two steps process, where first all elements but one are allocated, then the remaining one is allocated and the algorithm is ran for $t$ steps. From the convexity of the hockey-stick divergence we can upper bound the privacy profile of the random allocation scheme by the worst case allocation of all elements but the removed one.
From Lemma \ref{lem:domPostProc}, each intermediate call to the mechanism is a post process of the randomizer, which can be used to recursively define a randomized mapping from the random allocation over the randomizer to the allocation over the mechanism. Using the same lemma, this mapping implies that $\privProf{\alloc{t, k}{\alg}}(\eps) \le \privProf{\alloc{t, k}{\rand}}(\eps)$.

\begin{theorem}\Arxiv{[Reduction to randomizer]}\label{lem:singElem}
    Given $t \in \naturals$; $k \in [t]$ and a $t$-step algorithm defined by $\alg$ dominated by a randomizer $\rand$, we have $\privProf{\alloc{t, k}{\alg}}(\eps) \le \privProf{\alloc{t, k}{\rand}}(\eps)$.
\end{theorem}
\Arxiv{
\begin{proof}[Proof outline]
    Given $n \in \naturals$, a dataset $\dataset \in \domain^{n-1}$ and element $x \in \domain$, the random allocation scheme $\allocFunc{t, k}{\alg}{(\dataset, x)}$ can be decomposed into two steps. First all elements in $\dataset$ are allocated, then $x$ is allocated and the outputs are sampled based on the allocations. From the quasi-convexity of the hockey-stick divergence, this implies
    \[
        \HockeyStickDiv{\kappa}{\allocFunc{t, k}{\alg}{(\dataset, x)}}{\allocFunc{t, k}{\alg}{(\dataset, \bot)}} \le \underset{a \in \boldsymbol{a}^{t, k}(n-1)}{\max} \left(\HockeyStickDiv{\kappa}{\mathcal{A}_{t, k}^{a}(\alg; (\dataset, x))}{\mathcal{A}_{t, k}^{a}(\alg; (\dataset, \bot))}\right),
    \]
    where $\boldsymbol{a}^{t, k}(n-1)$ is the set of all possible allocations of $n-1$ elements into $k$ out of $t$ steps, and for any $a \in \boldsymbol{a}^{t, k}(n-1)$ we denote by $\mathcal{A}_{t, k}^{a}(\alg; (\dataset, x))$ the allocation scheme conditioned on the allocation of $\dataset$ according to $a$. 

    Fixing $a$ to be the allocation that maximizes the right-hand side of the inequality, and denoting $\dataset_{1}, \ldots, \dataset_{t}$ the subsets it induces, for any index $i \in [t]$ and input prefix view $\view^{i-1}$ we have $(\rand(*), \rand(\bot))$ dominates $(\alg((\dataset^{i}, x), \view^{i-1}), \alg((\dataset^{i}, \bot), \view^{i-1}))$, so from Lemma \ref{lem:domPostProc}, there exists a randomized mapping $\varphi_{\view^{i-1}}$ such that $\alg((\dataset^{i}, x), \view^{i-1}) = \varphi_{\view^{i-1}}(\rand(*))$ and $\alg((\dataset^{i}, \bot), \view^{i-1}) = \varphi_{\view^{i-1}}(\rand(\bot))$. Using these mappings, we recursively define another randomized mapping $\varphi$ by sequentially sampling $\out'_{i} \sim \varphi_{\view'^{i-1}}(\out_{i})$ for $i = 1, \ldots ,t$, and prove that $\mathcal{A}_{t, k}^{a}(\alg; (\dataset, x)) = \varphi(\allocFunc{t, k}{\rand}{*})$ and $\mathcal{A}_{t, k}^{a}(\alg; (\dataset, \bot)) = \varphi(\allocFunc{t, k}{\rand}{\bot})$, which by invoking Lemma \ref{lem:domPostProc} again, implies $\alloc{t, k}{\alg}$ is dominated by $\alloc{t, k}{\rand}$ and completes the proof.
\end{proof}
}
A special case of this result for Gaussian noise addition and $k=1$ was given by \citet[Theorem 1]{CGHLKKMSZ24}, and in the context of the matrix mechanism by \citet[Lemma 3.2]{CCGHST25}. The same bound for the Poisson scheme is a direct result from the combination of Claim \ref{clm:domPairComp} and \citet[Theorem 11]{ZDW22}.

Next we show how to translate any bound on the privacy profile of the random allocation with $k=1$ to the case of $k > 1$ by decomposing it to $k$ calls to a $1$ out of $t/k$ steps allocation process.

\begin{lemma}\Arxiv{[Reduction to a single allocation]}\label{lem:multAlloc}
    For any $k \in \naturals$, $\eps > 0$ we have $\privProf{\alloc{t, k}{\rand}}(\eps) \le \privProf{\alloc{\lfloor t/k \rfloor}{\rand}}^{\otimes k}(\eps)$, where ${\otimes k}$ denotes the composition of $k$ runs of the algorithm or scheme which in our case is $\alloc{\lfloor t/k \rfloor}{\rand}$.
\end{lemma}

\begin{movable}{mov:multAlloc}{}
\begin{proof}\NeurIPS{[Proof of Lemma \ref{lem:multAlloc}]}
    Notice that the random allocation of $k$ indexes out of $t$ can be described as a two steps process, first randomly splitting $t$ into $k$ subsets of size $t/k$, \footnote{For simplicity we assume that $t$ is divisible by $k$.} then running $\alloc{t/k, 1}{\rand}$ on each  of the $k$ copies of the scheme. Using the same convexity argument as in the proof of Lemma \ref{lem:singElem}, the privacy profile of $\alloc{t, k}{\rand}$ is upper bounded by the composition of $k$ copies of $\alloc{t/k, 1}{\rand}$.
\end{proof}
We remark that this lemma holds for arbitrary $t$-step algorithms (and not just non-adaptive randomizers) but in the adaptive case the usual  sequential composition should be replaced by concurrent composition \citep{VW21}, which was recently proven to provide the same privacy guarantees \citep{lyu22, VZ23}.
\end{movable}

Combining these two results, the privacy profile of the random allocation scheme is bounded by the (composition of the) hockey-stick divergence between $\allocFunc{t}{\rand}{*}$ and $\allocFunc{t}{\rand}{\bot} = \rand^{\otimes t}(\bot)$ in both directions, which we bound in three different ways in the following section.

\section{Privacy bounds}
\subsection{Truncated Poisson bound} \label{sec:truncPoisBnd}
Roughly speaking, our main theorem states that random allocation is asymptotically identical to the Poisson scheme with sampling probability $\approx k/t$ up to lower order terms. Formal proofs and missing details of this section can be found in Appendix \ref{apd:truncPoisBnd}.

\ifconferencemode
\begin{theorem}\label{\mainThmAsym}
    Given $\eps_{0} > 0$; $\delta_{0} \in [0, 1]$ and a $(\eps_{0}, \delta_{0})$-DP randomizer $\rand$, for any $\eps, \delta > 0$ we have $\privProf{\alloc{t}{\rand}}(\eps) \le \privProf{\Pois{t, \eta}{\rand}}(\eps) + t \delta_{0} + \delta$, where $\eta \coloneqq \frac{1}{t(1-\gamma)}$ and $\gamma \coloneqq \min \left\{\cosh(\eps_{0}) \cdot \sqrt{\frac{2}{t} \ln \left(\frac{1}{\delta} \right)}, 1 -\frac{1}{t} \right\}$.
    
    Furthermore, for any $\eps, \eps' > 0$ and randomizer $\rand$ we have $\privProfRem{\alloc{t}{\rand}}(\eps) \le \privProfRem{\Pois{t, \eta}{\rand}}(\eps) + \tau \cdot \privProfAdd{\alloc{t}{\rand}}\left(\eps' \right)$ and $\privProfAdd{\alloc{t}{\rand}}(\eps) \le \privProfAdd{\Pois{t, \eta}{\rand}}(\eps) + \tau e^{2 \eps'} \cdot \privProfAdd{\alloc{t}{\rand}}\left(\eps' \right)$, where $\eta \coloneqq \frac{e^{2 \eps'}}{t}$ and $\tau \coloneqq \frac{1}{e^{\eps'}(e^{\eps'}-1)}$.
\end{theorem}
\else
\begin{theorem}\label{\mainThmAsym}
        Given $\eps_{0} > 0$; $\delta_{0} \in [0, 1]$ and a $(\eps_{0}, \delta_{0})$-DP randomizer $\rand$, for any $\eps, \delta > 0$ we have $\privProf{\alloc{t}{\rand}}(\eps) \le \privProf{\Pois{t, \eta}{\rand}}(\eps) + t \delta_{0} + \delta$, where $\eta \coloneqq \frac{1}{t(1-\gamma)}$ and $\gamma \coloneqq \min \left\{\cosh(\eps_{0}) \cdot \sqrt{\frac{2}{t} \ln \left(\frac{1}{\delta} \right)}, 1 -\frac{1}{t} \right\}$.
\end{theorem}
\fi

Since $\eta$ corresponds to a sampling probability of $\frac{1}{t}$ up to a lower order term in $t$, this implies that the privacy of random allocation scheme is asymptotically upper-bounded by the Poisson scheme. While this holds for sufficiently large value of $t$, in many practical parameter regimes
\ifconferencemode
the second part of the theorem provides tighter bounds.
\else
the following theorem provides tighter bounds.
\begin{theorem}[Recursive bound]\label{\mainThmRec}
    For any $\eps, \eps' > 0$ we have $\privProfRem{\alloc{t}{\rand}}(\eps) \le \privProfRem{\Pois{t, \eta}{\rand}}(\eps) + \tau \cdot \privProfAdd{\alloc{t}{\rand}}\left(\eps' \right)$ and $\privProfAdd{\alloc{t}{\rand}}(\eps) \le \privProfAdd{\Pois{t, \eta}{\rand}}(\eps) + \tau e^{2 \eps'} \cdot \privProfAdd{\alloc{t}{\rand}}\left(\eps' \right)$,
    where $\eta \coloneqq \frac{e^{2 \eps'}}{t}$ and $\tau \coloneqq \frac{1}{e^{\eps'}(e^{\eps'}-1)}$.
\end{theorem}
\fi

While the recursive expression might seem to lead to a vacuous loop, it is in fact a useful tool. Notice that $\privProfAdd{\alloc{t}{\rand}}\left(\eps' \right) / \privProfAdd{\alloc{t}{\rand}}(\eps)$ quickly diminishes as $\eps' / \eps$ grows, so it suffices to set $\eps' = C \eps$ for some constant $1 \ll C$ for the second term to become negligible.
\ifconferencemode
Both parts of this theorem follow from Lemma \ref{lem:analBnd}  which bounds the privacy profile of the random allocation scheme by that of the corresponding Poisson scheme with sampling probability $\eta$, with an additional term roughly corresponding to a tail bound on the privacy loss of the allocation scheme.
\else
Both of these theorems rely on the next lemma that bounds the privacy profile of the random allocation scheme by that of the corresponding Poisson scheme with sampling probability $\eta$, with an additional term roughly corresponding to a tail bound on the privacy loss of the allocation scheme. 
\fi
\begin{movable}{mov:analBnd}{}
\begin{lemma}[Analytic bound]\label{lem:analBnd}
    For any $\eps > 0$ and $\eta \in [1/t, 1]$ we have $\privProfRem{\alloc{t}{\rand}}(\eps) \le \privProfRem{\Pois{t, \eta}{\rand}}(\eps) + \rem{\beta}_{\alloc{t}{\rand}}(\eta)$ and $\privProfAdd{\alloc{t}{\rand}}(\eps) \le \privProfAdd{\Pois{t, \eta}{\rand}}(\eps) + \add{\beta}_{\alloc{t}{\rand}}(\eta)$,
    where $\rem{\beta}_{\alloc{t}{\rand}} (\eta) \coloneqq P_{\allocFunc{t}{\rand}{*}}(B(\eta))$, $\add{\beta}_{\alloc{t}{\rand}} (\eta) \coloneqq P_{\allocFunc{t}{\rand}{\bot}}(B(\eta))$, and $B(\eta) \coloneqq \left\{\view \in \outDom^{t-1} ~\left\vert~ \underset{i \in [t-1]}{\max} \left(\loss{\view^{i}}{\bot}{*} \right.\right) > \ln(t \eta) \right\}$.
\end{lemma}
Theorem \ref{\mainThmAsym} follows from this lemma by identifying the total loss with a sum of the independent losses per step and using maximal Azuma-Hoeffding inequality. Theorem\ref{\mainThmRec} follows from this lemma by using a simple relationship between the privacy loss tail bound and the privacy profile.

The proof of this lemma consists of a sequences of reductions.
\begin{proof}
    Following \citep{EFMRTT19}, we introduce the posterior sampling scheme (Definition \ref{def:postSchem}), where the sampling probability depends on the previous outputs. Rather than selecting in advance a single step, at each step the scheme chooses to include the element with posterior probability induced by the previously released outputs $\lambda_{\view^{i}} \coloneqq P_{\allocFunc{t}{\rand}{*}}\left(i+1 \in \boldsymbol{I} \vert \view^{i} \right)$, where $\boldsymbol{I}$ is the subset of chosen steps. 
    
    Though this scheme seems like a variation of the Poisson scheme, we prove (Lemma \ref{lem:postEqAlloc}) that in fact its output is distributed like the output of random allocation, which implies they share the same privacy guarantees. The crucial difference between these two schemes is the fact that unlike random allocation, the distribution over the outputs of any step of the posterior scheme is independent of the distribution over output of previous steps given the view and the dataset, since there is no shared randomness (such as the chosen allocation).

    We then define a truncated variant of the posterior distribution (Definition \ref{def:trncPostScm}), where the sampling probability is capped by some threshold, and bound the difference between the privacy profile of the truncated and original posterior distributions, by the probability that the posterior sampling probability will exceed the truncation threshold (Lemma \ref{lem:truncPostBndPost}).

    Finally, we bound the privacy profile of the truncated posterior scheme by the privacy profile of the Poisson scheme with sampling probability corresponding to the truncation threshold, using the fact the privacy loss is monotonically increasing in the sampling probability (Lemma \ref{lem:truncBndPois}), which completes the proof. Part of these last two lemmas is a special case of the tail bound that recently proved in \citep[Theorem 3.1]{CCGST23}.

    Formally,
    \[
        \privProfRem{\alloc{t}{\rand}}(\eps) \eqExp{1}{=} \privProfRem{\post{t}{\rand}}(\eps) \eqExp{2}{\le} \privProfRem{\post{t, \eta}{\rand}}(\eps) + \rem{\beta}_{\alloc{t}{\rand}}(\eta) \eqExp{3}{\le} \privProfRem{\Pois{t, \eta}{\rand}}(\eps) + \rem{\beta}_{\alloc{t}{\rand}}(\eta),
    \]
	where (1) results from Lemma \ref{lem:postEqAlloc}, (2) from Lemma \ref{lem:truncPostBndPost}, and (3) from Lemma \ref{lem:truncBndPois}.

    The same proof can be repeated as is for the add direction. 
\end{proof}

\begin{remark}
    Repeating the previous lemmas while changing the direction of the inequalities and the sign of the lower order terms, we can similarly prove that the random allocation scheme upper bounds the Poisson scheme up to lower order terms, which implies they are asymptotically identical.
\end{remark}    
\end{movable}

\subsection{ Asymptotic analysis}\label{sec:asym}
Combining Theorem \ref{\mainThmAsym} with Lemma \ref{lem:multAlloc} and applying it to the Gaussian mechanism results in the next corollary.\arxvFtnt{Extending Lemma \ref{lem:analBnd} and Theorem \ref{thm:asymBnd} to directly account for allocation of $k$ steps might improve some lower order terms, but requires a more involved version of Lemma \ref{lem:truncPostBndPost}, specifically Claim \ref{clm:postProbEqLoss} on which its proof relies. We leave this for future work.}

\begin{corollary} \label{cor:GaussAnalBnd}
    Given $\eps, \delta > 0$, for any $\sigma > 8 \cdot \max \left\{\sqrt{\ln(t/\delta)}, \sqrt{\frac{k}{t}} \ln(t/\delta) \right\}$, we have $\privProf{\alloc{t, k}{N_{\sigma}}}(\eps) \le \privProf{\Pois{t, 2 k / t}{N_{\sigma}}}(\eps) + 2 \delta$, where $N_{\sigma}$ is the Gaussian mechanism.
\end{corollary}

Using this Corollary we can derive asymptotic bounds on the privacy guarantees of the Gaussian mechanism amplified by random allocation. Since the Gaussian mechanism is dominated by the one-dimensional Gaussian randomizer (Claim \ref{clm:domPairGauss}) where $\rand(*) = \mathcal{N}(1, \sigma^{2})$ and $\rand(\bot) = \mathcal{N}(0, \sigma^{2})$, this corollary implies that for sufficiently large $\sigma$, the random allocation scheme with the Gaussian mechanism  $\alloc{t, k}{N_{\sigma}}$ is $(\eps, \delta)$-DP for any $\eps > C \cdot \max\left\{\frac{k \sqrt{\ln(t/\delta)}}{\sigma \sqrt{t}}, \frac{k^{2} \sqrt{\ln(t/\delta)}}{t^{1.5}}\right\}$ for some universal constant $C$ (Lemma \ref{lem:GaussAsymBnd}). We note that the dependence of $\eps$ on $\sigma$; $\delta$; $k$; and $t$ matches that of the Poisson scheme for $\lambda = k / t$ (Lemma \ref{lem:poisPriv}) up to an additional logarithmic dependence on $t$ (Poisson scales with $\ln(1/\delta)$),
unlike the shuffle scheme which acquires an additional $\sqrt{\ln(1/\delta)}$ by converting approximate the DP randomizer to pure DP first, resulting in the bound $\eps \ge C' \cdot \frac{k \cdot \ln(1/\delta)}{\sigma \sqrt{t}}$ \citep{FMT21}. A detailed comparison can be found in Appendix \ref{apd:asym}.

The recursive bound (\NeurIPS{second part of }Theorem \ref{\mainThmRec}) provides similar asymptotic guarantees for arbitrary mechanisms, when $\eps_{0} \le 1$ for the local mechanism $\alg$. In this case, the privacy parameter of its corresponding Poisson scheme $\eps_{\mathcal{P}}$ is approximately linear in the sampling probability. Setting the sampling probability to $1/t$ and combining amplification by subsampling with advanced composition implies $\eps_{\mathcal{P}} = O\left(\eta \eps_{0} \sqrt{t \cdot \ln(1/\delta)} \right) = O\left(\eps_{0} \sqrt{\frac{\ln(1/\delta)}{t}} \right)$. Setting $\eps' = C \eps_{\mathcal{P}}$ for some constant $1 \ll C < 1/\eps_{\mathcal{P}}$, we get $\eps_{\mathcal{A}} = O\left(\eta \eps_{0} \sqrt{t \cdot \ln(1/\delta)} \right) =  O\left((1 + \eps_{\mathcal{P}})\eps_{0} \sqrt{\frac{\ln(1/\delta)}{t}} \right) = O(\eps_{\mathcal{P}} + \eps_{\mathcal{P}}^{2})$.

While Theorem \ref{\mainThmAsym} provides a full asymptotic characterization of the random allocation scheme, the bounds it induces could be suboptimal for small $t$ or large $\eps_{0}$. In the following section we provide several  bounds that hold in all parameter regimes. We also use these to ``bootstrap'' the recursive bound\Arxiv{ (Theorem \ref{\mainThmRec})}. 
\subsection{Poisson Decomposition}\label{sec:PoisDecom}

\begin{theorem}\Arxiv{[Decomposition bound]}\label{thm:dcmpBnd}
    For any $\eps > 0$ we have $\privProfRem{\alloc{t}{\rand}}(\eps) \le \rem{\gamma} \cdot \privProfRem{\Pois{t, \lambda}{\rand}}(\rem{\eps})$ and $\privProfAdd{\alloc{t}{\rand}}(\eps) \le \add{\gamma} \cdot \privProfAdd{\Pois{t, \lambda}{\rand}}(\add{\eps})$, where $\rem{\gamma} \coloneqq \frac{1}{1 - \left(1 - \lambda \right)^{t}}$, $\add{\gamma} \coloneqq 1  + e^{\eps}(\rem{\gamma}-1)$, $\rem{\eps} \coloneqq \ln (1 + (e^{\eps} - 1)/\rem{\gamma})$, and $\add{\eps} \coloneqq -\ln\left(1 - (1-e^{-\eps})/\rem{\gamma}\right)$.
\end{theorem}

We remark that while this theorem provides separate bounds for the add and remove adjacency notions\footnote{An earlier version of this work has mistakenly stated that an upper bound for the remove direction applies to both directions.}, numerical analysis seems to indicate that the bound on the remove direction is always larger than the one for the add direction.

Setting $\lambda \coloneqq 1/t$ yields $\rem{\gamma} \approx e/(e-1) \approx 1.6$, which bounds the difference between these two sampling methods up to this factor in $\eps$ in the $\eps < 1$ regime.

Formal proofs and missing details can be found in Appendix \ref{apd:PoisDecom}.
\begin{movable}{mov:dcmpBnd}{}
\begin{proof}\NeurIPS{[Proof of Theorem \ref{thm:dcmpBnd}]}
    The proof of this theorem consists of several key steps. First, we show (Lemma \ref{lem:allocMontn}) that increasing the number of allocations can only harm the privacy, that is, for any sequence of integers $k \le k_{1} < \ldots < k_{j} \le t$, and non-negative $\lambda_{1}, \ldots, \lambda_{j}$ s.t. $\lambda_{1} + \ldots + \lambda_{j} = 1$, the privacy profile of $\alloc{t, k}{\rand}$ is upper-bounded by the privacy profile of $\lambda_{1} \alloc{t, k_{1}}{\rand} + \ldots + \lambda_{j} \alloc{t, k_{j}}{\rand}$, where we use convex combinations of algorithms to denote an algorithm that randomly chooses one of the algorithms with probability given in the coefficient.

    Next, we notice (Lemma \ref{lem:PoisDecom}) that the Poisson scheme can be decomposed into a sequence of random allocation schemes, by first sampling the number of steps in which the element will participate from the Binomial distribution and then running the random allocation scheme for the corresponding number of steps,
        \[
	        \PoisFunc{t, \lambda}{\rand}{x} = \sum_{k = 0}^{t} B_{t, \lambda}(k) \cdot \allocFunc{t, k}{\rand}{x}, 
    	\]
    	where $B_{t, \lambda}$ is the PDF of the binomial distribution with parameters $t, \lambda$ and $\allocFunc{t, 0}{\rand}{x} \coloneqq \rand^{\otimes t}(\bot)$.

	We then define the Poisson scheme conditioned on allocating the element at least once
	    \[
    	    \mathcal{P}_{t, \lambda}^{+}(\rand; x) = \rem{\gamma} \sum_{k \in [t]} B_{t, \lambda}(k) \cdot \allocFunc{t, k}{\rand}{x},
	    \]
		and use a generalized version of the advanced joint convexity (Claim~\ref{clm:advJntCvx}) to relate its privacy profile that of the Poisson scheme (Lemma~\ref{lem:PoisUpper}). 
		
    Formally, 
    \begin{align*}
        \privProfRem{\alloc{t}{\rand}}(\eps) & = \HockeyStickDiv{e^{\eps}}{\allocFunc{t, k}{\rand}{*}}{\allocFunc{t, k}{\rand}{\bot}}
        \\ & \eqExp{1}{=} \HockeyStickDiv{e^{\eps}}{\rem{\gamma} \sum_{k \in [t]} B_{t, \lambda}(k) \cdot \allocFunc{t, 1}{\rand}{*}}{\allocFunc{t, k}{\rand}{\bot}}
        \\ & \eqExp{2}{\le} \HockeyStickDiv{e^{\eps}}{\rem{\gamma} \sum_{k \in [t]} B_{t, \lambda}(k)\cdot \allocFunc{t, k}{\rand}{*}}{\allocFunc{t, k}{\rand}{\bot}}
        \\ & \eqExp{3}{=} \HockeyStickDiv{e^{\eps}}{\mathcal{P}_{t, \lambda}^{+}(\rand; *)}{\mathcal{P}_{t, \lambda}^{+}(\rand; \bot)}
        \\ & \eqExp{4}{=} \rem{\gamma} \HockeyStickDiv{e^{\rem{\eps}}}{\PoisFunc{t, \lambda}{\rand}{*}}{\PoisFunc{t, \lambda}{\rand}{\bot}}
        \\ & = \rem{\gamma} \cdot \privProfRem{\Pois{t, \lambda}{\rand}}(\rem{\eps}),
    \end{align*}
    where (1) results from the fact $\sum_{k \in [t]} B_{t, \lambda}(k) = 1/\rem{\gamma}$, (2) from Lemma \ref{lem:allocMontn}, (3) from the definition of $\mathcal{P}_{t, \lambda}^{+}$ and the fact that $\PoisFunc{t, \lambda}{\rand}{\bot} = \rand^{\otimes t}(\bot) = \allocFunc{t, \lambda}{\rand}{\bot}$, and (4) from the first part of Lemma \ref{lem:PoisUpper}.
    
    Repeating the same proof using the second part of Lemma \ref{lem:PoisUpper} proves the bound on $\privProfAdd{\alloc{t}{\rand}}$.
\end{proof}

Combining the Poisson decomposition perspective shown in Lemma \ref{lem:PoisDecom} with the monotonicity in number of allocations shown in Lemma \ref{lem:allocMontn}, additionally implies the following corollary.
\begin{corollary}\label{cor:TruncPois}
    For any $\lambda \in [0, 1]$; $k \in [t]$, we have $\privProf{\Pois{t, \lambda, k}{\rand}}(\eps) \le \privProf{\Pois{t, \lambda}{\rand}}(\eps)$, where $\Pois{t, \lambda, k}{\rand}$ denotes the Poisson scheme where the number of allocations is upper bounded by $k$.
\end{corollary}
\end{movable}

\subsection{Direct analysis}\label{sec:dirAnals}
The previous bounds rely on a reduction to Poisson scheme. In this section we bound the privacy profile of the random allocation scheme directly, which is especially useful in the low privacy regime where the privacy profile of random allocation is lower than that of Poisson. Formal proofs and missing details of this section can be found in Appendix \ref{apd:dirAnals}.

Our main result expresses the RDP of the random allocation scheme in the remove direction in terms of the RDP parameters of the randomizer, and provides an approximate bound in the the add direction\footnote{An earlier version of this work has mistakenly stated that an upper bound for the remove direction applies to both directions.}. While the privacy bounds induced by RDP are typically looser than those relying on full analysis and composition of the privacy loss distribution (PLD), the gap nearly vanishes as the number of composed calls to the randomizer grows, as depicted in Figure \ref{fig:multi-epoch}.

\begin{theorem}\Arxiv{[Direct remove bound]}\label{thm:dirRemBnd}
    Given two integers $t, \alpha \in \naturals$, we denote by $\boldsymbol{\Pi}_{t}(\alpha)$ the set of integer partitions of $\alpha$ consisting of $\le t$ elements.\arxvFtnt{If $t \ge \alpha$, $\boldsymbol{\Pi}_{t}(\alpha) = \boldsymbol{\Pi}(\alpha)$ is simply the set of all integer partitions.} Given a partition $\Pi \in \boldsymbol{\Pi}_{t}(\alpha)$, we denote by $\binom{\alpha}{\Pi} = \frac{\alpha!}{\prod_{p \in \Pi} p!}$, and denote by $C(\Pi)$  the list of counts of unique values in $P$ (e.g. if $\alpha = 9$ and $\Pi = [1, 2, 3, 3]$ then $C(\Pi) = [1, 1, 2]$).
        For any $\alpha \ge 2$ and randomizer $\rand$ we have\arxvFtnt{The first version of this work stated an incorrect combinatorial coefficient in this expression. The numerical comparisons were based on the correct expression.}
    \[
        \RenyiDiv{\alpha}{\allocFunc{t}{\rand}{*}}{\allocFunc{t}{\rand}{\bot}} = \frac{1}{\alpha - 1} \ln \left( \frac{1}{t^{\alpha}} \sum_{\Pi \in \boldsymbol{\Pi}_{t}(\alpha)} \binom{t}{C(\Pi)} \binom{\alpha}{\Pi} \prod_{p \in \Pi} e^{(p-1)\RenyiDiv{p}{\rand(*)}{\rand(\bot)}}\right).
    \]
\end{theorem}
For the add direction we use a different bound.
\begin{theorem}\Arxiv{[Direct add bound]}\label{thm:dirAddBnd}
    Given $\gamma \in [0,1]$ and a randomizer $\rand$, we define a new randomizer $\rand_{\gamma}$ which given an input $x$ samples $\out \propto \frac{P_{\rand}(\out \vert x)^{\gamma} \cdot P_{\rand}(\out \vert \bot)^{1-\gamma}}{Z_{\gamma}}$, where $Z_{\gamma}$ is the normalizing factor.
    
    For any $\eps > 0$ we have $\privProfAdd{\alloc{t}{\rand}}(\eps) \le \HockeyStickDiv{e^{\eps'}}{\rand^{\otimes t}(\bot)}{\rand^{\otimes t}_{1/t}(*)}$, where $\eps' \coloneqq \eps -t \cdot \ln(Z_{1/t})$.
\end{theorem}
These two theorems follow from Lemmas \ref{lem:dirRemBnd} and \ref{lem:dirAddBnd} for $P = \rand(*)$ and $Q = \rand(\bot)$.

As is the case in Theorem \ref{thm:dcmpBnd}, numerical analysis seems to indicate the bound on the remove direction always dominates the one for the add direction.

Since we have an exact expression for the hockey-stick and R\'{e}nyi divergences of the Gaussian mechanism, these two theorems immediately imply the following corollary.
\begin{corollary}\label{cor:dirGauss}
    Given $\sigma > 0$, and a Gaussian mechanism $N_{\sigma}$, we have for any integer $\alpha \ge 2$
    \[
        \RenyiDiv{\alpha}{\allocFunc{t}{N_{\sigma}}{1}}{N_{\sigma}^{\otimes t}(0)} = \frac{1}{\alpha - 1} \left(\ln \left(\sum_{\Pi \in \boldsymbol{\Pi}_{t}(\alpha)} \binom{t}{C(\Pi)} \binom{\alpha}{\Pi} e^{\underset{p \in \Pi}{\sum}\frac{p^{2}}{2 \sigma^{2}}} \right) - \alpha \left(\frac{1}{2 \sigma^{2}} + \ln (t) \right) \right),
    \]
    and $\privProfAdd{\alloc{t}{N_{\sigma}}}(\eps) \le \privProf{N_{\sigma'}}(\eps')$ where $\sigma' \coloneqq \sqrt{t} \sigma$ and $\eps' \coloneqq \eps - \frac{1-1/t}{2 \sigma^{2}}$ for all $\eps \in \reals$.
\end{corollary}

Corollary \ref{cor:dirGauss} gives a simple way to exactly compute integer RDP parameters of random allocation with Gaussian noise in the remove direction. Interestingly, they closely match RDP parameters of the Poisson scheme with rate $1/t$ in most regimes (e.g. Fig.~\ref{fig:multi-epoch}). In fact, in some (primarily large $\eps$) parameter regimes the bounds based on RDP of allocation are lower than the PLD-based bounds for Poisson subsampling (Fig.~\ref{fig:RDP-dom}). The restriction to integer values has negligible effect, which can be further mitigated using \citep[Corollary 10]{WBK19}, which upper bounds the fractional R\'{e}nyi divergence by a linear combination of the R\'{e}nyi divergence of its rounded integer values. We also note that $\vert \boldsymbol{\Pi}_{t}(\alpha) \vert$ is sub-exponential in $\alpha$ which leads to performance issues in the very high privacy ($\eps \ll 1$) regime (Large $\sigma$ values in Fig~\ref{fig:main}). Since the typical value of $\alpha$ used for accounting is in the low tens, this quantity can be efficiently computed using several technical improvements which we discuss in Appendix \ref{apd:RDPimp}. On the other hand, in the very low privacy regime ($\eps \gg 1$), the $\alpha$ that leads to the best bound on $\eps$ is typically in the range $[1,2]$ which cannot be computed exactly using our method. Finally, we remark that while this result is stated only for $k=1$, it can be extended to $k>1$ using the same argument as in Lemma \ref{lem:multAlloc}. In fact, RDP based bounds are particularly convenient for subsequent composition which is necessary to obtain bounds for $k>1$ or multi-epoch training algorithms.

\section{Numerical evaluation}\label{sec:numEval}
\begin{figure}[ht]
  \centering
  \includegraphics[width=1\linewidth]{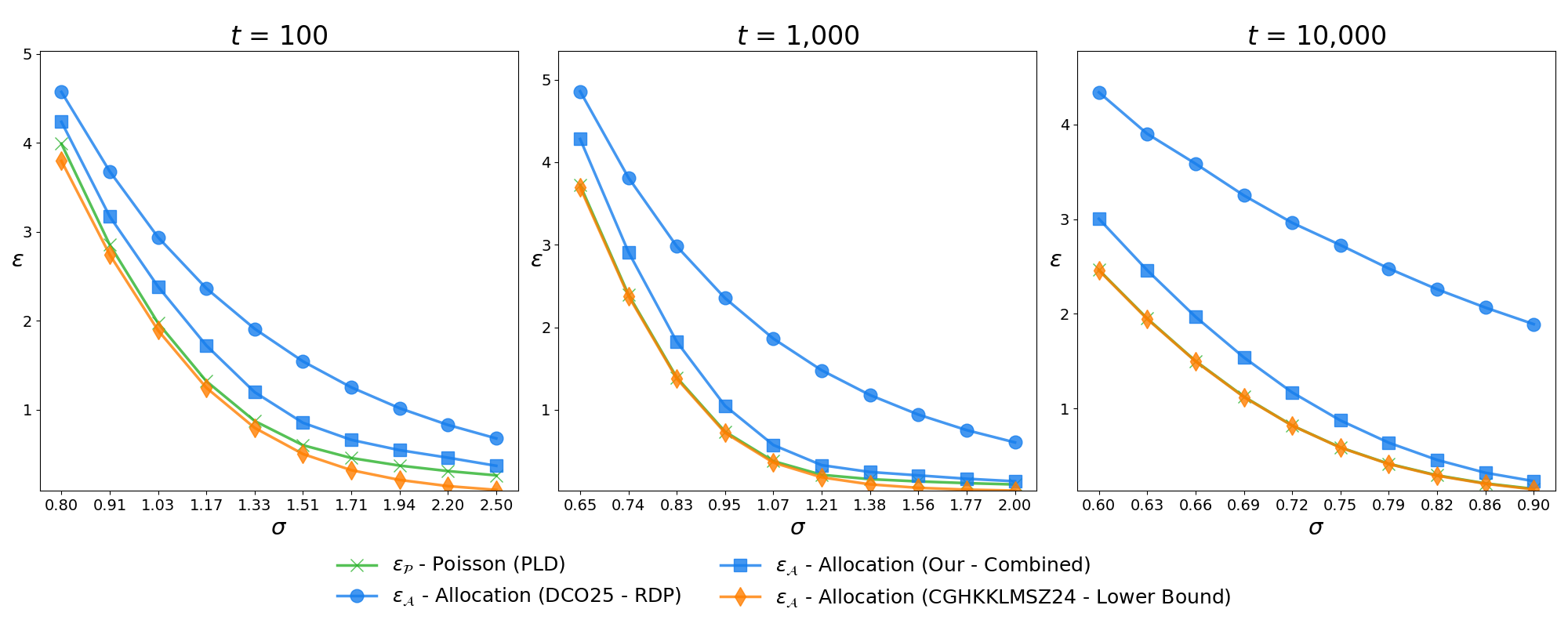}
  \caption{\small Bounds on privacy parameter $\eps$ as a function of the noise parameter $\sigma$ for various values of $t$, all using the Gaussian mechanism with $\delta = 10^{-10}$. We compare the minimum over all our methods to the independent results in \citet{DCO25}, lower bound by \citet{CGHLKKMSZ24}, and to the Poisson scheme with $\lambda = 1/t$.}
  \label{fig:multi-comp}
\end{figure}

In this section we demonstrate that numerical implementations of our results give the first nearly-tight and provable bounds on privacy amplification of random allocation with Gaussian noise, notably showing (Fig.~\ref{fig:main}, \ref{fig:multi-comp}) that they nearly match bounds known for Poisson subsampling. Compared to the Monte Carlo-based technique by \citet{CGHLKKMSZ24} (\ref{apd:monte-carlo}), we show in many regimes our results match these bounds up to constants in $\delta$ (logarithmic in $\eps$), and the computational limitation of the MC technique in the low $\delta$ and high confidence level regime. We additionally compare our results to the RDP-based approximation by \citet{DCO25} (\ref{apd:comp_loose_RDP}), and demonstrate the advantage of our tight RDP analysis in the regime where $k \ll t$. Their bound is tighter than our direct analysis in the $k \approx t$ regime, where the effect of amplification is small and $\eps$ is prohibitively large.

\Arxiv{\subsection{Privacy-utility tradeoff}\label{apd:util}}
\begin{movable}{mov:util}
{
We demonstrate the utility degradation induced by Poisson subsampling relative to random allocation using the simple setting of estimating the mean of a Bernoulli distribution from a sampled dataset (App.~\ref{apd:util}). We derive theoretical approximations for the mean square error of the two schemes and match them with numerical simulations, that demonstrate random allocation always has lower error for sufficiently large sample size. Together with the results of \citet{CGHLKKMSZ24}, our results imply that random allocation (or balls-and-bins sampling) has the utility benefits of shuffling while having the privacy benefits of Poisson subsampling.  This provides a (reasonably) practical way to reconcile a long-standing and concerning discrepancy between the practical implementations of DP-SGD and its commonly-used privacy analyses. 
}
The results of \citet{CGHLKKMSZ24} show that in the context of training DP-SGD, random allocation (or balls-and-bins sampling) has the utility benefits of shuffling while having the privacy benefits of Poisson subsampling. Here we investigate the privacy-utility trade-off in a simple-to-analyze setting of mean estimation over a Boolean hypercube, that illustrates one possible source of this relative advantage.

We start with the one-dimensional setting. Consider a dataset $\dataset \in \{0,1\}^{n}$ sampled iid from a Bernoulli distribution with expectation $p \in [0,1]$, where $p$ is estimated from the data elements using one of the two schemes. Formally, at each iteration, the algorithm reports a noisy sum of the elements in the corresponding subset $\out_{i}$, and the estimated expectation is $\hat{p} \coloneqq \frac{1}{n} \sum_{i \in [t]} \out_{i}$.

Since $\hat{p}$ is averaged over the various steps, in the case of random allocation with the Gaussian mechanism we have $\hat{p}_{\mathcal{A}} = \frac{1}{n}\left(\sum_{x \in \dataset} x + \sum_{i \in [t]} \xi_{i}\right)$ where $\xi_{i}$ is the noise added at step $i$. From the property of the Gaussian mechanism $\sum_{i \in [t]} \xi_{i}$ is a Gaussian random variable with variance $t \sigma^{2}$, and from the definition of the distribution, $\sum_{x \in \dataset} x \sim \text{Bin}(n, p)$, so $\hat{p}_{\mathcal{A}} \sim \frac{1}{n}\left(\text{Bin}(n, p) + \mathcal{N}(0, t \sigma^{2})\right)$. In particular this implies $\mathbb{E}[\hat{p}_{\mathcal{A}}] = p$ and $\text{Var}(\hat{p}_{\mathcal{A}}) = \frac{p(1-p)}{n} + \frac{t \sigma^{2}}{n^{2}}$, where the first term is the sampling noise and the second is the privacy noise.

Poisson subsampling adds some complexity to the analysis, but can be well approximated for large sample size. The estimation $\hat{p}_{\mathcal{P}}$ follows a similar distribution to that of $\hat{p}_{\mathcal{A}}$, with an additional step. First we sample $u \sim \text{Bin}(n, p)$, then - following the insight introduced in Lemma \ref{lem:PoisDecom} - we sample $v_{i} \sim \text{Bin}(t, 1/t)$ for all $i \in [u]$, which amounts to sampling $m \sim \text{Bin}(u \cdot t, 1/t)$. We note that $\mathbb{E}[m] = u$ and $\text{Var}(m) = u \cdot t \cdot \frac{1}{t}\left(1-\frac{1}{t}\right) \approx u $. Since w.h.p. $u \approx p \cdot n$, we get $\mathbb{E}[\hat{p}_{\mathcal{P}}] = p$ and $\text{Var}(\hat{p}_{\mathcal{A}}) \approx \frac{p(1-p)}{n} + \frac{p}{n} + \frac{t \sigma^{2}}{n^{2}}$, where the first term is the sampling noise, the second is the Poisson sampling noise, and the third is the privacy noise.

The noise scale required to guarantee some fixed privacy parameters using the random allocation scheme is typically larger than the one required by the Poisson scheme, as shown in Figures \ref{fig:main}. But following the asymptotic analysis discussed in Section \ref{sec:asym} we have $t \sigma^{2} \approx \frac{\ln(1/\delta)}{\eps^{2}}$ for both schemes, with constants differing by $\approx 10\%$ in most practical parameter regimes, as shown in Figures \ref{fig:main} and \ref{fig:multi-comp}, which implies the privacy noise $\frac{t \sigma^{2}}{n^{2}}$ is typically slightly larger for the random allocation scheme. On the other hand, for $p \rightarrow 1$, the Poisson sampling noise is arbitrarily larger than the sampling noise. Since the privacy bound becomes negligible as $n$ increases, we get that the random allocation scheme asymptotically (in $n$) dominates the Poisson scheme, as illustrated in Figure~\ref{fig:util}. In the $\eps = 1, d = 1$ case, the induced $\sigma$ is sufficiently small, so the gap is dominated by the additional Poisson sampling noise, when $\eps = 0.1$, this effect becomes dominate only for relatively large sample size.

In the high-dimensional setting the privacy noise dominates the sampling noise and therefore the privacy-utility tradeoff is dominated by the difference in the (known) privacy guarantees of the two schemes. In  Figure~\ref{fig:util} we give an example of this phenomenon for $d=1000$.

\begin{figure}[ht]
    \centering
    \includegraphics[width=1\linewidth]{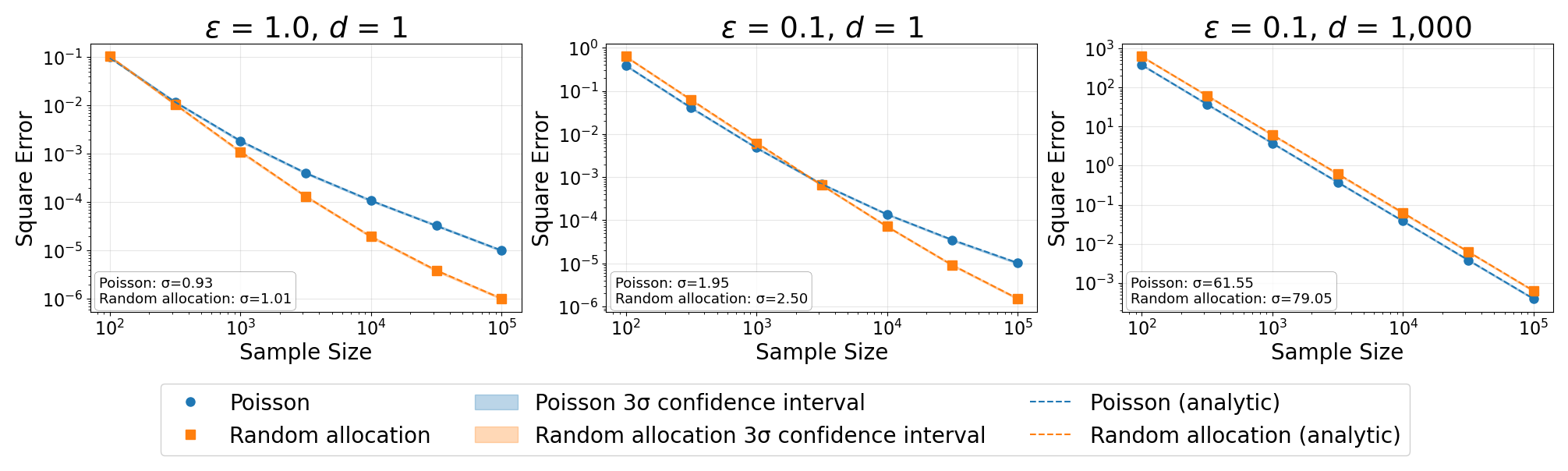}
    \caption{\small Analytical and empirical square error for the Poisson and random allocation scheme for the setting discussed in Appendix \ref{apd:util}, for various values of $\eps$ and $d$ (which corresponds to an increase in sensitivity). We set $p=0.9$, $t=10^{3}$, $\delta = 10^{-10}$. The experiment was carried $10^{4}$ times, so the $3$-std confidence intervals are barely visible.}
    \label{fig:util}
\end{figure}    
\end{movable}

\section{Discussion}
This work provides the first theoretical guarantees and numerical estimation algorithms for the random allocation sampling scheme. Its main analysis shows that its privacy guarantees are asymptotically identical to those of the Poisson scheme. We provide two additional analyses which lead to tighter bounds in some setting (Fig. \ref{fig:main}). The resulting combined bound of the random allocation remains close to that of the Poisson scheme in many practical regimes (Fig. \ref{fig:multi-comp}, \ref{fig:Chau_et_al_delta}), and even exceeds it in some. Unlike the Poisson scheme, our bounds are analytical and do not rely on numerical PLD analysis, which results in some remaining slackness. Further, unlike PLD-based bounds, our $(\eps,\delta)$ bounds do not lend themselves for tight privacy accounting of composition. Both of these limitations are addressed in our subsequent work \citep{FS25b} where we show that PLD of random allocation can be approximated efficiently, leading to tighter and more general numerical bounds. 

\Arxiv{\subsection*{Acknowledgments}}
\begin{movable}{mov:ack}{}
We are grateful to Kunal Talwar for proposing the problem that we investigate here as well as suggesting the decomposition-based approach for the analysis of this problem (established in Theorem \ref{thm:dcmpBnd}).  We thank Thomas Steinke and Christian Lebeda for pointing out that an  earlier version of our work mistakenly did not analyze the add direction of our direct bound. We also thank Hilal Asi, Hannah Keller, Guy Rothblum and Katrina Ligett for thoughtful comments and motivating discussions of these results and their application in \citep{AFKRT25}. 
Shenfeld’s work was supported by the Apple Scholars in AI/ML PhD Fellowship.
\end{movable}

\ifconferencemode
\newpage
\begin{ack}
\showstored{mov:ack}
\end{ack}
\else
\newpage
\fi

\bibliography{bibliography}
\bibliographystyle{plainnat}

\newpage
\appendix
\ifconferencemode
\section{Missing definitions and claims} \label{apd:prem}

\showstored{mov:RDPimpDP}

\showstored{mov:GaussMech}

\showstored{mov:postProc}

\fi

\section{Missing proofs from Section \ref{sec:renRed}} \label{apd:renRed}

\begin{proof}[Proof of Lemma \ref{lem:singElem}]
    Given $n \in \naturals$, a dataset $\dataset \in \domain^{n-1}$ and element $x \in \domain$, the random allocation scheme $\allocFunc{t, k}{\alg}{(\dataset, x)}$ can be decomposed into two steps. First all elements in $\dataset$ are allocated, then $x$ is allocated and the outputs are sampled based on the allocations. We denote by $\boldsymbol{a}^{t, k}(n-1)$ the set of all possible allocations of $n-1$ elements into $k$ out of $t$ steps, and for any $a \in \boldsymbol{a}^{t, k}(n-1)$ denote by $\mathcal{A}_{t, k}^{a}(\alg; (\dataset, x))$ the allocation scheme conditioned on the allocation of $\dataset$ according to $a$. Given the neighboring datasets $(\dataset, x)$ and $(\dataset, \bot)$ we have,
    \begin{align*}
        \HockeyStick{\kappa} & \left(\allocFunc{t, k}{\alg}{(\dataset, x)} \Vert \allocFunc{t, k}{\alg}{(\dataset, \bot)} \right) 
        \\ & = \HockeyStickDiv{\kappa}{\frac{1}{\left\vert \boldsymbol{a}^{t, k}(n-1) \right\vert} \sum_{a \in \boldsymbol{a}^{t, k}(n-1)} \mathcal{A}_{t, k}^{a}(\alg; (\dataset, x))}{\frac{1}{\left\vert \boldsymbol{a}^{t, k}(n-1) \right\vert} \sum_{a \in \boldsymbol{a}^{t, k}(n-1)} \mathcal{A}_{t, k}^{a}(\alg; (\dataset, \bot))}
        \\ & \le \underset{a \in \boldsymbol{a}^{t, k}(n-1)}{\max} \left(\HockeyStickDiv{\kappa}{\mathcal{A}_{t, k}^{a}(\alg; (\dataset, x))}{\mathcal{A}_{t, k}^{a}(\alg; (\dataset, \bot))}\right),
    \end{align*}
    and similarly,
    \[
        \HockeyStickDiv{\kappa}{\allocFunc{t, k}{\alg}{(\dataset, \bot)}}{\allocFunc{t, k}{\alg}{(\dataset, x)}} \le \underset{a \in \boldsymbol{a}^{t, k}(n-1)}{\max} \left(\HockeyStickDiv{\kappa}{\mathcal{A}_{t, k}^{a}(\alg; (\dataset, \bot))}{\mathcal{A}_{t, k}^{a}(\alg; (\dataset, x))}\right),
    \]
    where the inequality results from the quasi-convexity of the hockey-stick divergence.

    Fixing $a$ to be the allocation that maximizes the right-hand side of the inequality, we denote by $\dataset_{1}, \ldots, \dataset_{t}$ the subsets defined by $a$.\footnote{Conceptually, this is equivalent to considering the random allocation scheme over a single element $x$, with a sequence of mechanisms $\alg_{\dataset^{j}}$ defined by the various subsets.} From the definition of the randomizer, for any index $j \in [t]$ and input prefix view $\view^{j-1}$ we have $(\rand(*), \rand(\bot))$ dominates $(\alg((\dataset^{j}, x), \view^{j-1}), \alg((\dataset^{j}, \bot), \view^{j-1}))$, so from Lemma \ref{lem:domPostProc}, there exists a randomized mapping $\varphi_{\view^{j-1}}$ such that $\alg((\dataset^{j}, x), \view^{j-1}) = \varphi_{\view^{j-1}}(\rand(*))$ and $\alg((\dataset^{j}, \bot), \view^{j-1}) = \varphi_{\view^{j-1}}(\rand(\bot))$.\footnote{We note that $\varphi$ depends on $\dataset^{j}$ and $x$ as well. We omit them from notations for simplicity, since they are fixed at this point of the argument.}
    Using these mappings, we will recursively define another randomized mapping $\varphi$. Given an output view $\view \in \outDom^{t}$, we define $\view' \sim \varphi(\view)$ by sequentially sampling $\out'_{j} \sim \varphi_{\view'^{j-1}}(\out_{j})$ for $j = 1, \ldots ,t$, where $\view'^{j} \coloneqq (\out_{1}, \ldots, \out_{j})$ and $\view'^{0} \coloneqq \emptyset$.

    We will now prove that $\mathcal{A}_{t, k}^{a}(\alg; (\dataset, x)) = \varphi(\allocFunc{t, k}{\rand}{*})$ and $\mathcal{A}_{t, k}^{a}(\alg; (\dataset, \bot)) = \varphi(\allocFunc{t, k}{\rand}{\bot})$, which by invoking Lemma \ref{lem:domPostProc} again, implies $\alloc{t, k}{\alg}$ is dominated by $\alloc{t, k}{\rand}$ and completes the proof.

    From the law of total probability we have,
    \[
        P_{\mathcal{A}_{t, k}^{a}(\alg; (\dataset, x))}(\view) = \frac{1}{\binom{t}{k}} \underset{\boldsymbol{i} \subseteq [t], \vert \boldsymbol{i} \vert = k}{\sum} P_{\mathcal{A}_{t, k}^{a}(\alg; (\dataset, x))}(\view \vert \boldsymbol{i}) = \frac{1}{\binom{t}{k}} \underset{\boldsymbol{i} \subseteq [t], \vert \boldsymbol{i} \vert = k}{\sum} \prod_{j \in [t]} P_{\mathcal{A}_{t, k}^{a}(\alg; (\dataset, x))}(\view^{j} \vert \view^{j-1}, \boldsymbol{i})
    \]
    and
    \[
        P_{\allocFunc{t, k}{\rand}{*}}(\view) = \frac{1}{\binom{t}{k}} \underset{\boldsymbol{i} \subseteq [t], \vert \boldsymbol{i} \vert = k}{\sum} P_{\allocFunc{t, k}{\rand}{*}}(\view \vert \boldsymbol{i}) = \frac{1}{\binom{t}{k}} \underset{\boldsymbol{i} \subseteq [t], \vert \boldsymbol{i} \vert = k}{\sum}  \prod_{j \in [t]} P_{\allocFunc{t, k}{\rand}{*}}(\view^{j} \vert \view^{j-1}, \boldsymbol{i}),
    \]
    where $\boldsymbol{i} \subseteq [t]$ denots the allocation of the single element $x$.

    Using these identities, it suffices to prove that for any subset of indexes $\boldsymbol{i}$, index $j \in [t]$, and input prefix view $\view^{j-1}$, we have $P_{\mathcal{A}_{t, k}^{a}(\alg; (\dataset, x))}(\view^{j} \vert \view^{j-1}, \boldsymbol{i}) = P_{\varphi(\allocFunc{t, k}{\rand}{*})}(\view^{j} \vert \view^{j-1}, \boldsymbol{i})$ and $P_{\mathcal{A}_{t, k}^{a}(\alg; (\dataset, \bot))}(\view^{j} \vert \view^{j-1}, \boldsymbol{i}) = P_{\varphi(\allocFunc{t, k}{\rand}{\bot})}(\view^{j} \vert \view^{j-1}, \boldsymbol{i})$.

    From the definition,
    \begin{align*}
        P_{\mathcal{A}_{t, k}^{a}(\alg; (\dataset, x))}(\view^{j} \vert \view^{j-1}, \boldsymbol{i}) & = P_{\mathcal{A}_{t, k}^{a}(\alg; (\dataset, x))}(\out_{j}~\vert~\view^{j-1}, \boldsymbol{i})
        \\ & = \begin{cases}
            P_{\alg((\dataset^{j}, x), \view^{j-1})}(\out_{j}) & j \in \boldsymbol{i} \\
            P_{\alg((\dataset^{j}, \bot), \view^{j-1})}(\out_{j}) & j \notin \boldsymbol{i}          
        \end{cases}
        \\ & = \begin{cases}
            P_{\varphi_{\view^{j-1}}(\rand(*))}(\out_{j}) & j \in \boldsymbol{i} \\
            P_{\varphi_{\view^{j-1}}(\rand(\bot))}(\out_{j}) & j \notin \boldsymbol{i}
        \end{cases}
        \\ & = P_{\varphi(\allocFunc{t, k}{\rand}{*})}(\out_{j}~\vert~\view^{j-1}, \boldsymbol{i})
        \\ & = P_{\varphi(\allocFunc{t, k}{\rand}{*})}(\view^{j}~\vert~\view^{j-1}, \boldsymbol{i}).
    \end{align*}
    In the case of the null element, the allocation doesn't have any effect so we have,
    \[
        P_{\mathcal{A}_{t, k}^{a}(\alg; (\dataset, \bot))}(\view^{j} \vert \view^{j-1}) = P_{\alg((\dataset^{j}, \bot), \view^{j-1})}(\out_{j}) = P_{\varphi_{\view^{j-1}}(\rand(\bot))}(\out_{j}) = P_{\varphi(\allocFunc{t, k}{\rand}{\bot})}(\view^{j}~\vert~\view^{j-1})
    \]
    which completes the proof.
\end{proof}

\showstored{mov:multAlloc}

\section{Missing proofs from Section \ref{sec:truncPoisBnd}}\label{apd:truncPoisBnd}

\showstored{mov:analBnd}

Throughout the rest of the section, claims will be stated in terms of $\rand$ and $*$ for simplicity, but can be generalized to $\alg$ and $x$. We note that $\allocFunc{t, k}{\rand}{\bot} = \rand^{\otimes t}(\bot)$ where $\rand^{\otimes t}(\bot)$ denotes $t$ sequential calls to $\rand(\bot)$, an identity which will be used several times throughout the next section

\subsection{Posterior scheme}
We start by introducing the posterior sampling scheme, where the sampling probability depends on the previous outputs. 
\begin{definition}[Posterior probability and scheme] \label{def:postSchem}
    Given a subset size $k \in [t]$, an index $i \in \left[t-1 \right]$, a view $\view^{i} \in \outDom^{i}$, and a randomizer $\rand$, the $i+1$ \emph{posterior probability} of the $k$ allocation out of $t$ given $\view^{i}$ is the probability that the index $i+1$ was one of the $k$ steps chosen by the random allocation scheme, given that the view $\view^{i}$ was produced by the first $i$ rounds of $\allocFunc{t}{\rand}{*}$. Formally, $\lambda_{\view^{i}, k} \coloneqq P_{\allocFunc{t, k}{\rand}{*}}\left(i+1 \in \boldsymbol{I} \vert \view^{i} \right)$, where $\boldsymbol{I}$ is the subset of chosen steps.

    The \emph{posterior scheme} is a function $\post{t, k}{\rand}: \{*, \bot\} \rightarrow \outDom^{t}$ parametrized by a randomizer $\rand$, number of steps $t$, and number of selected steps $k$, which given $*$, sequentially samples
    \[
        \out_{i+1} \sim \left(\lambda_{\view^{i}, k} \cdot \rand(*) + (1-\lambda_{\view^{i}, k}) \cdot \rand(\bot) \right), 
    \]
    where $\lambda_{\view^{0}, k} = k/t$, and $\postFunc{t, k}{\rand}{\bot} = \allocFunc{t, k}{\rand}{\bot}$.  As before, we omit $k$ from the notations where $k=1$.
\end{definition}

Though this scheme seems like a variation of the Poisson scheme, the following lemma shows that in fact its output is distributed like the output of random allocation.
\begin{lemma} \label{lem:postEqAlloc}
    For any subset size $k \in [t]$ and randomized $\rand$, $\allocFunc{t, k}{\rand}{*}$ and $\postFunc{t, k}{\rand}{*}$ are identically distributed, which implies $\privProfRem{\alloc{t, k}{\rand}}(\eps) = \privProfRem{\post{t, k}{\rand}}(\eps)$ and $\privProfAdd{\alloc{t, k}{\rand}}(\eps) = \privProfAdd{\post{t, k}{\rand}}(\eps)$ for any randomizer and all $\eps \ge 0$.
\end{lemma}

\begin{proof}
    We notice that for all $j \in [t-1]$ and $\view^{j} \in \outDom^{j}$, 
    \begin{align*}
        P_{\allocFunc{t, k}{\rand}{*}}(\view^{j+1} \vert \view^{j}) & = \frac{P_{\allocFunc{t, k}{\rand}{*}}(\view^{j+1})}{P_{\alloc{t, k}{\rand}}(\view^{j})}
        \\ & \eqExp{1}{=} \frac{\underset{\boldsymbol{i} \subseteq [t], \vert \boldsymbol{i} \vert = k}{\sum} P_{\allocFunc{t, k}{\rand}{*}}(\view^{j+1}, \boldsymbol{I} = \boldsymbol{i})}{P_{\allocFunc{t, k}{\rand}{*}}(\view^{j})}
        \\ & \eqExp{2}{=} \frac{\underset{\boldsymbol{i} \subseteq [t], \vert \boldsymbol{i} \vert = k}{\sum} P_{\allocFunc{t, k}{\rand}{*}}(\view^{j}, \boldsymbol{I} = \boldsymbol{i}) \cdot P_{\allocFunc{t, k}{\rand}{*}}(\out_{j+1} \vert \boldsymbol{I} = \boldsymbol{i}, \view^{j})}{P_{\allocFunc{t, k}{\rand}{*}}(\view^{j})}
        \\ & \eqExp{3}{=} \left(\underset{\boldsymbol{i} \subseteq [t], \vert \boldsymbol{i} \vert = k, j+1 \notin \boldsymbol{i}}{\sum} \frac{P_{\allocFunc{t, k}{\rand}{*}}(\view^{j}, \boldsymbol{I} = \boldsymbol{i})}{P_{\allocFunc{t, k}{\rand}{*}}(\view^{j})} \right) P_{\rand(\bot)}(\out_{j+1})
        \\ & ~~~~ + \left(\underset{\boldsymbol{i} \subseteq [t], \vert \boldsymbol{i} \vert = k, j+1 \in \boldsymbol{i}}{\sum} \frac{P_{\allocFunc{t, k}{\rand}{*}}(\view^{j}, \boldsymbol{I} = \boldsymbol{i})}{P_{\allocFunc{t, k}{\rand}{*}}(\view^{j})} \right) P_{\rand(*)}(\out_{j+1})
        \\ & = P_{\allocFunc{t, k}{\rand}{*}}(j+1 \notin \boldsymbol{I} \vert \view^{j}) \cdot P_{\rand(\bot)}(\out_{j+1} \vert \view^{j}) + P_{\allocFunc{t, k}{\rand}{*}}(j+1 \in \boldsymbol{I} \vert \view^{j}) \cdot P_{\rand(*)}(\out_{j+1})
        \\ & = (1-\lambda_{\view^{j}, k}) \cdot P_{\rand(\bot)}(\out_{j+1}) + \lambda_{\view^{j}, k} \cdot P_{\rand(*)}(\out_{j+1})
        \\ & \eqExp{3}{=}  P_{\postFunc{t, k}{\rand}{*}}(\view^{j+1} \vert \view^{j}), 
    \end{align*}
    where (1) denotes the subset of steps selected by the allocation scheme by $\boldsymbol{I}$ so $\boldsymbol{I} = \boldsymbol{i}$ denotes the selected subset was $\boldsymbol{i}$, (2) results from the definition $\view^{j+1} = (\view^{j}, \out_{j+1})$ and Bayes law, (3) from the fact that if $j+1 \in \boldsymbol{I}$ then $\out_{j+1}$ depends only on a $*$ and if $j+1 \notin \boldsymbol{I}$ then $\out_{j+1}$ depends only on $\bot$, and (4) is a direct result of the posterior scheme definition.

    Since $P(\view \vert *) = \prod_{i\in[t-1]} P(\view^{j+1} \vert *, \view^{j})$ for any scheme, this completes the proof. 
\end{proof}

\subsection{Truncated scheme}
Next we define a truncated variant of the posterior distribution and use it to bound its privacy profile.
\begin{definition}[Truncated scheme]\label{def:trncPostScm}
    The \emph{truncated posterior scheme} is a function $\post{t, k, \eta}{\rand}: \{*, \bot\} \rightarrow \outDom^{t}$ parametrized by a randomized $\rand$, number of steps $t$, number of selected steps $k$, and threshold $\eta \in [0, 1]$, which given $*$, sequentially samples
    \[
        \out_{i+1} \sim \left(\lambda_{\view^{i}, k}^{\eta} \cdot \rand(*) + (1-\lambda_{\view^{i}, k}^{\eta}) \cdot \rand(\bot) \right), 
    \]
    where $\lambda_{\view^{i}, k}^{\eta} \coloneqq \min \{\lambda_{\view^{i}, k},  \eta \}$, and $\postFunc{t, k, \eta}{\rand}{\bot} = \postFunc{t, k}{\rand}{\bot}$.
\end{definition}

Next we relate the privacy profiles of the posterior and truncated schemes.
\begin{lemma}\label{lem:truncPostBndPost}
    Given a randomizer $\rand$, for any $\eta \in [k/t,1]$; $\eps > 0$ we have
    \[
        \privProfRem{\post{t, k}{\rand}}(\eps) \le \privProfRem{\post{t, k, \eta}{\rand}}(\eps) + \rem{\beta}_{\alloc{t}{\rand}}(\eta) ~~~ \text{and} ~~~ \privProfAdd{\post{t, k}{\rand}}(\eps) \le \privProfAdd{\post{t, k, \eta}{\rand}}(\eps) + \add{\beta}_{\alloc{t}{\rand}}(\eta)
    \]
    where $\rem{\beta}_{\alloc{t}{\rand}}(\eta)$ and $\add{\beta}_{\alloc{t}{\rand}}(\eta)$ were defined in Lemma \ref{lem:analBnd}.
\end{lemma}

The proof of this lemma relies on the relation between $\lambda_{\view^{i}}$ and the privacy loss of the random allocation scheme, stated in the next claim.

\begin{claim}\label{clm:postProbEqLoss}
    Given $i \in [t-1]$, a randomizer $\rand$, and a view $\view^{i}$ we have $\lambda_{\view^{i}} = \frac{1}{t} e^{\lossAlg{\alloc{t}{\rand}}{\view^{i}}{\bot}{*}}$.
\end{claim}
\begin{proof}
    From the definition,
    \begin{align*}
        \lambda_{\view^{i}} & = P_{\allocFunc{t}{\rand}{*}} \left(I = i+1 \vert \view^{i} \right)
        \\ & \eqExp{1}{=} \frac{P_{\allocFunc{t}{\rand}{*}} \left(\view^{i} \vert I = i+1 \right) \cdot P\left(I = i+1 \right)}{P_{\allocFunc{t}{\rand}{*}} \left(\view^{i} \right)}
        \\ & \eqExp{2}{=} \frac{1}{t} \cdot \frac{P_{\allocFunc{t}{\rand}{\bot}} \left(\view^{i} \right)}{P_{\allocFunc{t}{\rand}{*}} \left(\view^{i} \right)}
        \\ & = \frac{1}{t} e^{\lossAlg{\alloc{t}{\rand}}{\view^{i}}{\bot}{*}},
    \end{align*}
    where (1) results from Bayes law and (2) from the fact $P\left(I = i+1 \right) = \frac{1}{t}$ and $P_{\allocFunc{t}{\rand}{*}} \left(\view^{i} \vert I = i+1 \right) = P_{\allocFunc{t}{\rand}{\bot}} \left(\view^{i} \right)$.
\end{proof}

\begin{proof}[Proof of Lemma \ref{lem:truncPostBndPost}]
    Denoting $\mathcal{B}_{\eta}^{t} \coloneqq \left\{\view \in \outDom^{t} ~ \big \vert ~ \underset{i \in [t-1]}{\max} \left(\lambda_{\view^{i}}\right) > \eta  \right\}$, for any $\mathcal{C} \subseteq \outDom^{t}$ we have
    \begin{align*}
        \underset{\viewRV \sim \postFunc{t}{\rand}{*}}{\mathbb{P}} & (\viewRV \in \mathcal{C})
        \\ & = \prob{\viewRV \sim \postFunc{t}{\rand}{*}}{\viewRV \in \mathcal{C} / \mathcal{B}_{\eta}^{t}} + \prob{\viewRV \sim \postFunc{t}{\rand}{*}}{\viewRV \in \mathcal{C} \cap \mathcal{B}_{\eta}^{t}}
        \\ & \eqExp{1}{=} \prob{\viewRV \sim \postFunc{t, \eta}{\rand}{*}}{\viewRV \in \mathcal{C} / \mathcal{B}_{\eta}^{t}} + \prob{\viewRV \sim \postFunc{t}{\rand}{*}}{\viewRV \in \mathcal{C} \cap \mathcal{B}_{\eta}^{t}}
        \\ & \eqExp{2}{\le} e^{\eps} \prob{\viewRV \sim \postFunc{t, \eta}{\rand}{\bot}}{\viewRV \in \mathcal{C} / \mathcal{B}_{\eta}^{t}} + \HockeyStickDiv{e^{\eps}}{\postFunc{t, \eta}{\rand}{*}}{\postFunc{t, \eta}{\rand}{\bot}} + \prob{\viewRV \sim \postFunc{t}{\rand}{*}}{\viewRV \in \mathcal{B}_{\eta}^{t}}
        \\ & \le e^{\eps} \prob{\viewRV \sim \postFunc{t, \eta}{\rand}{\bot}}{\viewRV \in \mathcal{C}} + \HockeyStickDiv{e^{\eps}}{\postFunc{t, \eta}{\rand}{*}}{\postFunc{t, \eta}{\rand}{\bot}} + \prob{\viewRV \sim \postFunc{t}{\rand}{*}}{\viewRV \in \mathcal{B}_{\eta}^{t}},
    \end{align*}
    which after reordering the terms implies
    \[
        \HockeyStickDiv{e^{\eps}}{\postFunc{t}{\rand}{*}}{\postFunc{t}{\rand}{\bot}} \eqExp{2}{\le} \HockeyStickDiv{e^{\eps}}{\postFunc{t, \eta}{\rand}{*}}{\postFunc{t, \eta}{\rand}{\bot}} + \prob{\viewRV \sim \postFunc{t}{\rand}{*}}{\viewRV \in \mathcal{B}_{\eta}^{t}},
    \]
    and similarly,
    \[
        \HockeyStickDiv{e^{\eps}}{\postFunc{t}{\rand}{\bot}}{\postFunc{t}{\rand}{*}} \eqExp{2}{\le} \HockeyStickDiv{e^{\eps}}{\postFunc{t, \eta}{\rand}{\bot}}{\postFunc{t, \eta}{\rand}{*}} + \prob{\viewRV \sim \postFunc{t}{\rand}{\bot}}{\viewRV \in \mathcal{B}_{\eta}^{t}},
    \]
    where (1) results from the definition of the truncated posterior scheme and the set $\mathcal{B}_{\eta}^{t}$, (2) from the fact that for any couple of distributions $P, Q$ over some domain $\outDom$
    \[
        \HockeyStickDiv{e^{\eps}}{P}{Q} = \underset{\mathcal{C} \subseteq \outDom^{t}}{\sup} \left(\prob{\outRV \sim P}{\outRV \in \mathcal{C}} - e^{\eps} \prob{\outRV \sim Q}{\outRV \in \mathcal{C}} \right),
    \]
    and (3) from the definition of $\rem{\beta}_{\alloc{t}{\rand}} (\eta)$.

    Combining this with Claim \ref{clm:postProbEqLoss} we get
    \begin{align*}
        \prob{\viewRV \sim \postFunc{t}{\rand}{*}}{\viewRV \in \mathcal{B}_{\eta}^{t}} & = \prob{\viewRV \sim \postFunc{t}{\rand}{*}}{\underset{i \in [t-1]}{\max} \left(\lambda_{\view^{i}}\right) > \eta} 
        \\& = \prob{\viewRV \sim \postFunc{t}{\rand}{*}}{\underset{i \in [t-1]}{\max} \left(\lossAlg{\alloc{t}{\rand}}{\view^{i}}{\bot}{*} \right) > \ln(t \eta)} 
        \\& = \rem{\beta}_{\alloc{t}{\rand}} (\eta),
    \end{align*}
    and similarly
    \[
        \prob{\viewRV \sim \allocFunc{t}{\rand}{\bot}}{\viewRV \in \mathcal{B}_{\eta}^{t}}, = \add{\beta}_{\alloc{t}{\rand}} (\eta).
    \]
\end{proof}

We now relate the truncated scheme's privacy profile to Poisson.

\begin{lemma} \label{lem:truncBndPois}
    Given $\eta \in [0, 1]$ and a randomizer $\rand$, we have $\privProfRem{\post{t, \eta}{\rand}}(\eps) \le \privProfRem{\Pois{t, \eta}{\rand}}(\eps)$ and $\privProfAdd{\post{t, \eta}{\rand}}(\eps) \le \privProfAdd{\Pois{t, \eta}{\rand}}(\eps)$ for all $\eps > 0$.
\end{lemma}

The proof makes use of the following result.
\begin{claim}[Theorem 10 in \citep{ZDW22}] \label{clm:domPairComp}
    If a pair of distributions $(P, Q)$ dominates a randomizer $\rand$ and $(P', Q')$ dominate $\rand'$, then $(P \times P', Q \times Q')$ dominate the composition of $\rand$ and $\rand'$.
\end{claim}

\begin{proof} [Proof of Lemma \ref{lem:truncBndPois}]
    We first notice that the the hockey-stick divergence between a mixture distribution $\lambda P + (1-\lambda) Q$ and its second component $Q$ is monotonically increasing in its mixture parameter $\lambda$. For any $0 \le \lambda \le \lambda' \le 1$ and two distributions $P_{0}, P_{1}$ over some domain, denoting $Q_{\lambda} \coloneqq (1-\lambda) P_{0} + \lambda P_{1}$ we have, $Q_{\lambda'} = \frac{1-\lambda'}{1-\lambda} Q_{\lambda} + \frac{\lambda'-\lambda}{1-\lambda} P_{1}$. From the quasi-convexity of the hockey-stick divergence, for any $\kappa \ge 0$ we have 
    \[
        \HockeyStickDiv{\kappa}{Q_{\lambda'}}{P_{1}} = \HockeyStickDiv{\kappa}{\frac{1-\lambda'}{1-\lambda} Q_{\lambda} + \frac{\lambda'-\lambda}{1-\lambda} P_{1}}{P_{1}} \le \HockeyStickDiv{\kappa}{Q_{\lambda}}{P_{1}},
    \]
    and similarly, $\HockeyStickDiv{\kappa}{P_{1}}{Q_{\lambda'}} \le \HockeyStickDiv{\kappa}{P_{1}}{Q_{\lambda}}$.
    
    Using this fact we get that the privacy profile of a single call to a Poisson subsampling algorithm is monotonically increasing in its sampling probability w.r.t. both $\simRem$ and $\simAdd$, so the privacy profile of every step of $\post{t, \eta}{\rand}$ is upper bounded by that of $\Pois{1, \eta}{\rand}$, and from Claim \ref{clm:domPairComp} its $t$ times composition is the dominating pair of $\Pois{t, \eta}{\rand}$, which completes the proof.
\end{proof}

\subsection{Proof of Lemma \ref{lem:analBnd}}
\begin{proof}
    The proof directly results from combining the previous lemmas.
    \[
        \privProfRem{\alloc{t}{\rand}}(\eps) \eqExp{1}{=} \privProfRem{\post{t}{\rand}}(\eps) \eqExp{2}{\le} \privProfRem{\post{t, \eta}{\rand}}(\eps) + \rem{\beta}_{\alloc{t}{\rand}}(\eta) \eqExp{3}{\le} \privProfRem{\Pois{t, \eta}{\rand}}(\eps) + \rem{\beta}_{\alloc{t}{\rand}}(\eta),
    \]
	where (1) results from Lemma \ref{lem:postEqAlloc}, (2) from Lemma \ref{lem:truncPostBndPost}, and (3) from Lemma \ref{lem:truncBndPois}.

    The same proof can be repeated as is for the add direction. 
\end{proof}

\label{apd:analBnd}
\ifconferencemode
\subsection{Proof of the first part of Theorem \ref{\mainThmAsym}}
\else
\subsection{Proof of Theorem \ref{\mainThmAsym}}
\fi

We first reduce the analysis of general approximate-DP algorithms to that of pure-DP ones, paying an additional $t \delta_{0}$ term in the probability.
\begin{claim}\label{clm:approxToPure}
    Given $\eps_{0} > 0$; $\delta_{0} \in [0, 1]$ and a $(\eps_{0}, \delta_{0})$-DP randomizer $\rand$, there exists a randomized $\hat{\rand}$ which is $\eps_{0}$-DP, such that $\rem{\beta}_{\alloc{t, k}{\rand}}(\eta) \le \rem{\beta}_{\alloc{t, k}{\hat{\rand}}}(\eta) + t \delta_{0}$ and $\add{\beta}_{\alloc{t, k}{\rand}}(\eta) \le \add{\beta}_{\alloc{t, k}{\hat{\rand}}}(\eta) + t \delta_{0}$, where $\beta_{\alloc{t, k}{\rand}}(\eta)$ was defined in Lemma \ref{lem:analBnd}.
\end{claim}

\begin{proof}[Proof of Claim \ref{clm:approxToPure}]
    From Lemma 3.7 in \citep{FMT21}, there exists a randomizer $\hat{\rand}$ which is $\eps_{0}$-DP, such that $\hat{\rand}(\bot) = \rand(\bot)$ and $D_{TV}(\rand(*) \Vert \hat{\rand}(*)) \le \delta_{0}$.
    
    For any $i \in [t]$ consider the posterior scheme $\post{t, k, (i)}{\hat{\rand}}$ which $\forall j < i$ returns 
    \[
        \out_{j+1} \sim \left(\lambda_{\view^{j}, k} \cdot \rand(*) + (1-\lambda_{\view^{j}, k}) \cdot \rand(\bot) \right), 
    \]
    and $\forall j \ge i$ returns
    \[
        \out_{j+1} \sim \left(\lambda_{\view^{i}, k} \cdot \hat{\rand}(*) + (1-\lambda_{\view^{j}, k}) \cdot \hat{\rand}(\bot) \right).
    \]
    
    Notice that $\post{t, k, (0)}{\hat{\rand}} = \post{t, k}{\rand}$ and $\post{t, k, (t)}{\hat{\rand}} = \post{t, k}{\hat{\rand}}$. From the definition, for any $i \in [t]$ we have $D_{TV}\left(\postFunc{t, k, (i-1)}{\hat{\rand}}{*} \Vert \postFunc{t, k, (i)}{\hat{\rand}}{*} \right) \le \delta_{0}$, which implies $D_{TV}\left(\postFunc{t, k}{\rand}{*} \Vert \postFunc{t, k}{\hat{\rand}}{*} \right) \le t \delta_{0}$.

    Combining this inequality with the fact that for any two distributions $P, Q$ over domain $\Omega$ and a subset $\mathcal{C} \subseteq \Omega$ we have $P(\mathcal{C}) \le Q(\mathcal{C}) + D_{TV}(P \Vert Q)$ completes the proof.
\end{proof}

Next we provide a closed form expression for the privacy loss of the random allocation scheme.
\begin{claim}\label{clm:randAlocPrivLoss}
    Given an index $i \in [t]$ and a view $\view^{i} \in \outDom^{i}$ we have,
    \[
        \lossAlg{\alloc{t}{\rand}}{\view^{i}}{*}{\bot} = \ln\left(\frac{1}{t} \left(t-i + \sum_{j \in [i]} e^{\lossAlg{\rand}{\out_{j}}{*}{\bot}}\right) \right).
    \]
\end{claim}

\begin{proof}
    From the definition,
    \[
        P_{\allocFunc{t}{\rand}{*}}(\view^{i} \vert I=j) = \begin{cases}
            \left( \prod_{m = 0}^{i-1} P_{\rand(\bot)}(\out_{j}) \right) P_{\rand(*)}(\out_{i}) \left( \prod_{j = i+1}^{t} P_{\rand(\bot)}(\out_{j}) \right) & j \le i
            \\ \prod_{m \in [i]} P_{\rand(\bot)}(\out_{m}) & j > i
        \end{cases}
    \]
    which implies,
    \[
        \frac{P_{\allocFunc{t}{\rand}{*}}(\view^{i} \vert I=j)}{P_{\allocFunc{t}{\rand}{\bot}}(\view^{i})} = \begin{cases}
            \frac{P_{\rand(*)}(\out_{i})}{P_{\rand(\bot)}(\out_{i})} & j \le i
            \\1 & j > i
        \end{cases}.
    \]
    Using this identity we get,
    \begin{align*}
        \lossAlg{\alloc{t}{\rand}}{\view^{i}}{*}{\bot} & = \ln\left(\frac{P_{\allocFunc{t}{\rand}{*}}(\view^{i})}{P_{\allocFunc{t}{\rand}{\bot}}(\view^{i})} \right)
        \\ & = \ln\left(\frac{\frac{1}{t} \sum_{j=[t]} P_{\allocFunc{t}{\rand}{*}}(\view^{i} \vert I=j)}{P_{\allocFunc{t}{\rand}{\bot}}(\view^{i} \vert \bot)} \right)
        \\ & = \ln\left(\frac{1}{t} \left(t-i + \sum_{j \in [i]} \frac{P_{\rand(*)}(\out_{j} \vert \view^{j-1}) }{P_{\rand(\bot)}(\out_{j} \vert \view^{j-1})} \right) \right)
        \\ & = \ln\left(\frac{1}{t} \left(t-i + \sum_{j \in [i]} e^{\lossAlg{\rand}{\out_{j}}{*}{\bot}} \right) \right).
    \end{align*}
\end{proof}
We are now ready to prove the corollary.
\begin{proof}[Proof of Theorem \ref{\mainThmAsym}]
    Using Claim \ref{clm:approxToPure} we can limit our analysis to a $\eps_{0}$-pure DP randomizer. We have,
    \begin{align*}
        \rem{\beta}_{\alloc{t}{\rand}}\left(\eta \right) & = \prob{\viewRV \sim \allocFunc{t}{\rand}{*}}{\underset{i \in [t-1]}{\max} \left(\lossAlg{\alloc{t}{\rand}}{\viewRV^{i}}{\bot}{*} \right)> \ln(t \eta)}
        \\ & \eqExp{1}{=} \prob{\viewRV \sim \allocFunc{t}{\rand}{*}}{\underset{i \in [t]}{\max}\left(\frac{1}{t-i + \sum_{j \in [i]} e^{\lossAlg{\rand}{\outRV_{j}}{*}{\bot}}} \right) > \eta}
        \\ & = \frac{1}{t} \sum_{l \in [t]} \prob{\viewRV \sim \allocFunc{t}{\rand}{*}}{\underset{i \in [t]}{\max}\left(\frac{1}{t + \sum_{j \in [i]} \left(e^{\lossAlg{\rand}{\outRV_{j}}{*}{\bot}} - 1\right)}\right) > \eta ~ \vert ~ I=l}
        \\ & = \frac{1}{t} \sum_{l \in [t]} \prob{\viewRV \sim \allocFunc{t}{\rand}{*}}{\underset{i \in [t]}{\max}\left(\sum_{j \in [i]} \left(1 - e^{\lossAlg{\rand}{\outRV_{j}}{*}{\bot}}\right)\right) > t\left(1 - \frac{1}{t \eta}\right) ~ \vert ~ I=l},
    \end{align*}
    where (1) results from Claim \ref{clm:randAlocPrivLoss}, and similarly,
    \[
        \add{\beta}_{\alloc{t}{\rand}}\left(\eta \right) \le \prob{\viewRV \sim \allocFunc{t}{\rand}{\bot}}{\underset{i \in [t]}{\max}\left(\sum_{j \in [i]} (1 - e^{\lossAlg{\rand}{\outRV_{j}}{*}{\bot}})\right) > t\left(1 - \frac{1}{t \eta}\right)}.
    \]

    We can now define the following martingale; $D_{0} \coloneqq 0$, $\forall j \in [t-1] : D_{j} \coloneqq 1 - e^{\lossAlg{\rand}{\outRV_{j}}{*}{\bot}}$, and $S_{i} \coloneqq \sum_{j = 0}^{i} D_{j}$. Notice that this is a sub-martingale since for any $j \in [t-1]$
    \[
        \expect{\outRV \sim \rand(\bot)}{1 - e^{\lossAlg{\rand}{\outRV}{*}{\bot}}} = 1 - \expect{\outRV \sim \rand(\bot)}{\frac{P_{\rand(*)}(\outRV)}{P_{\rand(\bot)}(\outRV)}} = 0
    \]
    and 
    \[
        \expect{\outRV \sim \rand(*)}{1 - e^{\lossAlg{\rand}{\outRV}{*}{\bot}}} = 1 - \exp\left(\RenyiDiv{2}{\rand(*)}{\rand(\bot)} \right) \le 0, 
    \]
    where $\Renyi{\alpha}$ is the $\alpha$-R\'{e}nyi divergence (Definition \ref{def:renDiv}).
    
    From the fact $\rand$ is $\eps_{0}$-DP we have $1 - e^{-\eps_{0}} \le D_{j} \le 1 - e^{\eps_{0}}$ almost surely, so the range of $D_{j}$ is bounded by $e^{\eps_{0}} - e^{-\eps_{0}} = 2 \cosh(\eps_{0})$, and we can invoke the Maximal Azuma-Hoeffding inequality and get for any $l \in [t]$, 
    \begin{align*}
        \underset{\viewRV \sim \allocFunc{t}{\rand}{*}}{\mathbb{P}} & \left(\underset{i \in [t]}{\max}\left(\sum_{j \in [i]} \left(1 - e^{\lossAlg{\rand}{\outRV_{j}}{*}{\bot}}\right)\right) > t\left(1 - \frac{1}{t \eta}\right) ~ \vert ~ I=l \right)
        \\ & \le \prob{\viewRV \sim \allocFunc{t}{\rand}{\bot}}{\underset{i \in [t]}{\max}\left(\sum_{j \in [i]} (1 - e^{\lossAlg{\rand}{\outRV_{j}}{*}{\bot}})\right) > t\left(1 - \frac{1}{t \eta}\right)}
        \\ & = \prob{\viewRV \sim \allocFunc{t}{\rand}{\bot}}{\underset{i \in [t]}{\max}\left(C_{i} \right) > t\left(1 - \frac{1}{t \eta}\right)}
        \\ & \le \exp \left(- \frac{t}{2} \left(\frac{t \eta - 1}{t \eta \cosh(\eps_{0})}\right)^{2} \right).
        \\ & \le \exp \left(- \frac{t}{2} \left(\frac{\gamma}{\cosh(\eps_{0})}\right)^{2} \right).
        \\ & \le \delta,
    \end{align*}
    where the last two steps result from the definition of $\eta$ and $\gamma$.
\end{proof}

\ifconferencemode
\subsection{Proof of the second part of Theorem \ref{\mainThmRec}}
\else
\subsection{Proof of Theorem \ref{\mainThmRec}}
\fi

The proof makes use of the following result.
\begin{claim}[Part 2 of lemma 3.3 in \citet{KS14}]\label{clm:DPtoHP}
    Given $\eps > 0$; $\delta \in [0,1]$ and a $(\eps, \delta)$-DP algorithm $\alg$, for any neighboring datasets $\dataset \simeq \dataset'$ we have,
    \[
        \prob{\outRV \sim \alg(\dataset)}{\loss{\outRV}{\dataset}{\dataset'} > 2 \eps} \le \frac{2 \delta}{1 - e^{-\eps}}.\footnote{We note the journal version has a typo in the $\delta$ part of the statement, which does not match the proof. We use the corrected version which appears in the Arxiv version.}
    \]
\end{claim}

\begin{proof}
    Consider the following algorithm based on the randomizer $\rand_{\eta} : \{*, \bot \} \times \outDom^{*} \rightarrow \outDom$ which is defined by $\rand_{\eta}(\bot; \view) = \rand(\bot)$, and $\rand_{\eta}(*; \view) = \begin{cases}
        \rand(*) & \lossAlg{\alloc{t}{\rand}}{\view^{i}}{\bot}{*} < \ln(t \eta)
        \\ \rand(\bot) & \lossAlg{\alloc{t}{\rand}}{\view^{i}}{\bot}{*} \ge \ln(t \eta)
    \end{cases}$.

    Given a view $\view$, denote
    \[
        i_{\view} \coloneqq \begin{cases}
            \underset{i \in [t-1]}{\arg \min} \left(\lossAlg{\alloc{t}{\rand}}{\view^{i}}{\bot}{*} \ge \ln(t \eta) \right) & \underset{i \in [t-1]}{\max} \left(\lossAlg{\alloc{t}{\rand}}{\view^{i}}{\bot}{*} \right) > \ln(t \eta)
            \\ 0 & \underset{i \in [t-1]}{\max} \left(\lossAlg{\alloc{t}{\rand}}{\view^{i}}{\bot}{*} \right) \le \ln(t \eta)
        \end{cases},
    \]
    the first index where the privacy loss exceeds $\ln(t \eta)$ if such index exists and $0$ otherwise, and notice that,
    \[
        \lossAlg{\alloc{t}{\rand}}{\view}{\bot}{*} \eqExp{1}{=} \ln\left(\frac{1}{\frac{1}{t} \sum_{i \in [t]} e^{\lossAlg{\rand}{\out_{i}}{*}{\bot}`}} \right) \eqExp{2}{=} \ln\left(\frac{t}{t - i_{\view} + \sum_{i \in [i_{\view}]} e^{\lossAlg{\rand}{\out_{i}}{*}{\bot}}} \right),
    \]
    where (1) results from Claim \ref{clm:randAlocPrivLoss} and (2) from the definition of $\rand_{\eta}$ which implies the distribution of $\rand_{\eta}(*; \view^{j}) = \rand_{\eta}(\bot; \view^{j})$ for all $j > i_{\view}$.

    Using this fact we get,
    \begin{align*}
        \rem{\beta}_{\alloc{t}{\rand}}\left(\eta \right) & = \prob{\viewRV \sim \allocFunc{t}{\rand}{*}}{\underset{i \in [t-1]}{\max} \left(\lossAlg{\alloc{t}{\rand}}{\view^{i}}{\bot}{*} \right)> \ln(t \eta)}
        \\ & \eqExp{1}{=} \prob{\viewRV \sim \allocFunc{t}{\rand}{*}}{\lossAlg{\alloc{t}{\rand}}{\view^{i_{\view}}}{\bot}{*} > \ln(t \eta)}
        \\ & \eqExp{2}{=} \prob{\viewRV \sim \allocFunc{t}{\rand}{*}}{\frac{1}{t-i_{\view} + \sum_{j \in [i_{\view}]} e^{\lossAlg{\rand}{\out_{j}}{*}{\bot}}} > \eta}
        \\ & \eqExp{3}{=} \prob{\viewRV \sim \allocFunc{t}{\rand_{\eta}}{*}}{\frac{1}{t-i_{\view} + \sum_{j \in [i_{\view}]} e^{\lossAlg{\rand}{\out_{j}}{*}{\bot}}} > \eta}
        \\ & \eqExp{4}{=} \prob{\viewRV \sim \allocFunc{t}{\rand_{\eta}}{*}}{\lossAlg{\alloc{t}{\rand_{\eta}}}{\viewRV}{\bot}{*} > \ln(t \eta)}
        \\ & \eqExp{5}{\le} \frac{1}{t \eta} \cdot \prob{\viewRV \sim \allocFunc{t}{\rand_{\eta}}{\bot}}{\lossAlg{\alloc{t}{\rand_{\eta}}}{\viewRV}{\bot}{*} > \ln(t \eta)}
        \\ & \eqExp{6}{\le} \frac{2}{t \eta - \sqrt{t \eta}} \cdot \privProfAdd{\alloc{t}{\rand_{\eta}}}\left(\frac{\ln(t \eta)}{2}\right)
        \\ & \eqExp{7}{\le} \frac{2}{t \eta - \sqrt{t \eta}} \cdot \privProfAdd{\alloc{t}{\rand}}\left(\frac{\ln(t \eta)}{2}\right)
        \\ & \eqExp{8}{=} \tau \cdot \privProfAdd{\alloc{t}{\rand}}\left(\eps' \right),
    \end{align*}
    and repeating all steps but (5) we similarly get,
    \[
        \add{\beta}_{\alloc{t}{\rand}}\left(\eta \right) \le \tau e^{2 \eps'} \cdot \privProfAdd{\alloc{t}{\rand}}\left(\eps' \right),
    \]
    where (1) results from the definition of $i_{\view}$, (2) from Claim \ref{clm:randAlocPrivLoss}, (3) from the definition of $\rand_{\eta}$, (4) from the previous identity, (5) from the definition of the privacy loss, (6) from Claim \ref{clm:DPtoHP}, (7) from the fact $\rand_{\eta}$ is dominated by $\rand$, and (8) from the definition of $\eta$ and $\tau$.
\end{proof}

\subsection{Separate directions}
For completeness we present the results of figure \ref{fig:main} for the add and remove directions separately for a single and multiple allocations.

\begin{figure}[htbp]
    \centering
    \makebox[\textwidth][c]{
        \begin{subfigure}{0.6\textwidth}
            \centering
            \includegraphics[width=\textwidth]{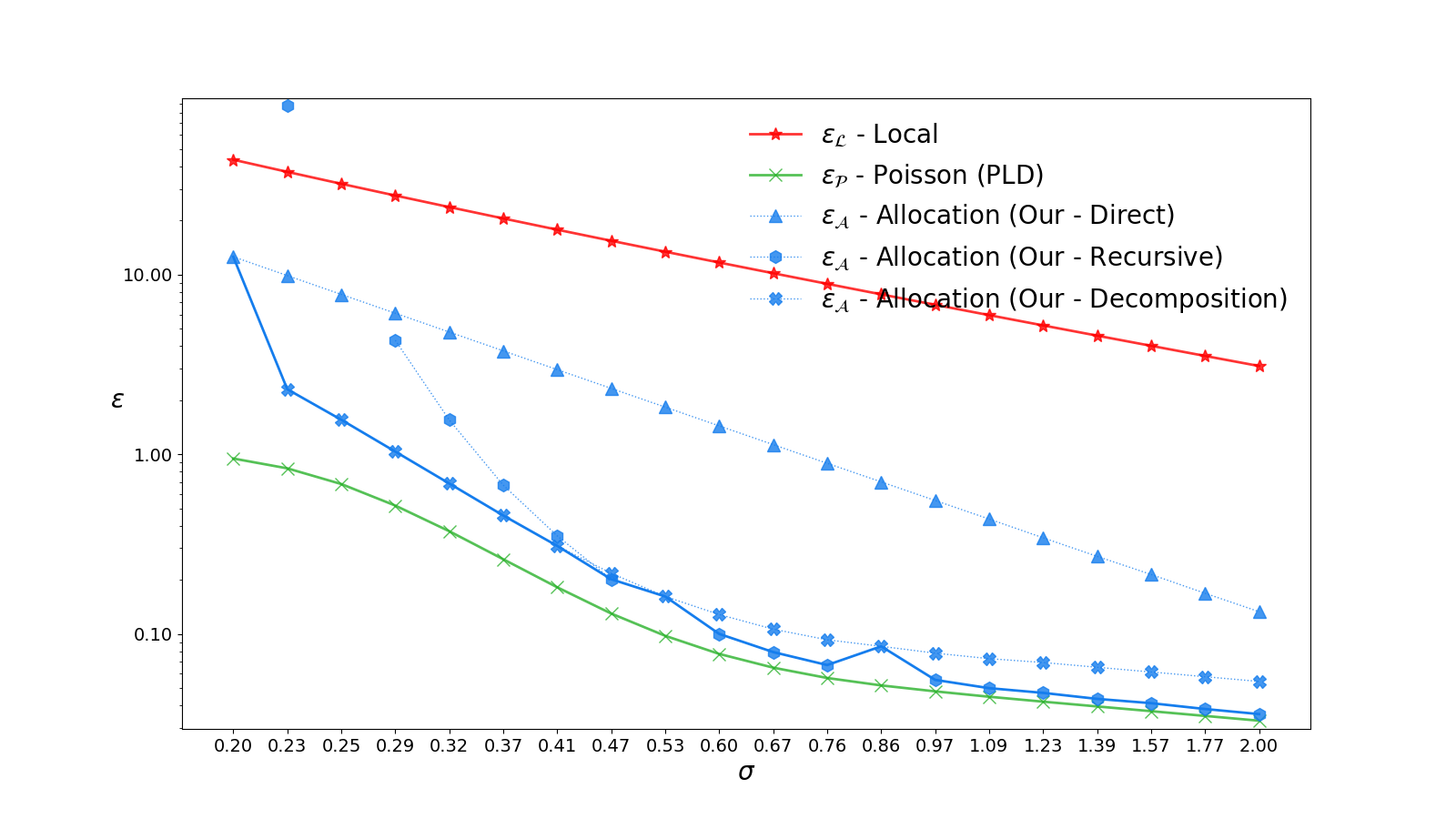}
            \caption{\small Add direction}
            \label{fig:main_add_remove_sub1}
        \end{subfigure}
        \begin{subfigure}{0.6\textwidth}
            \centering
            \includegraphics[width=\textwidth]{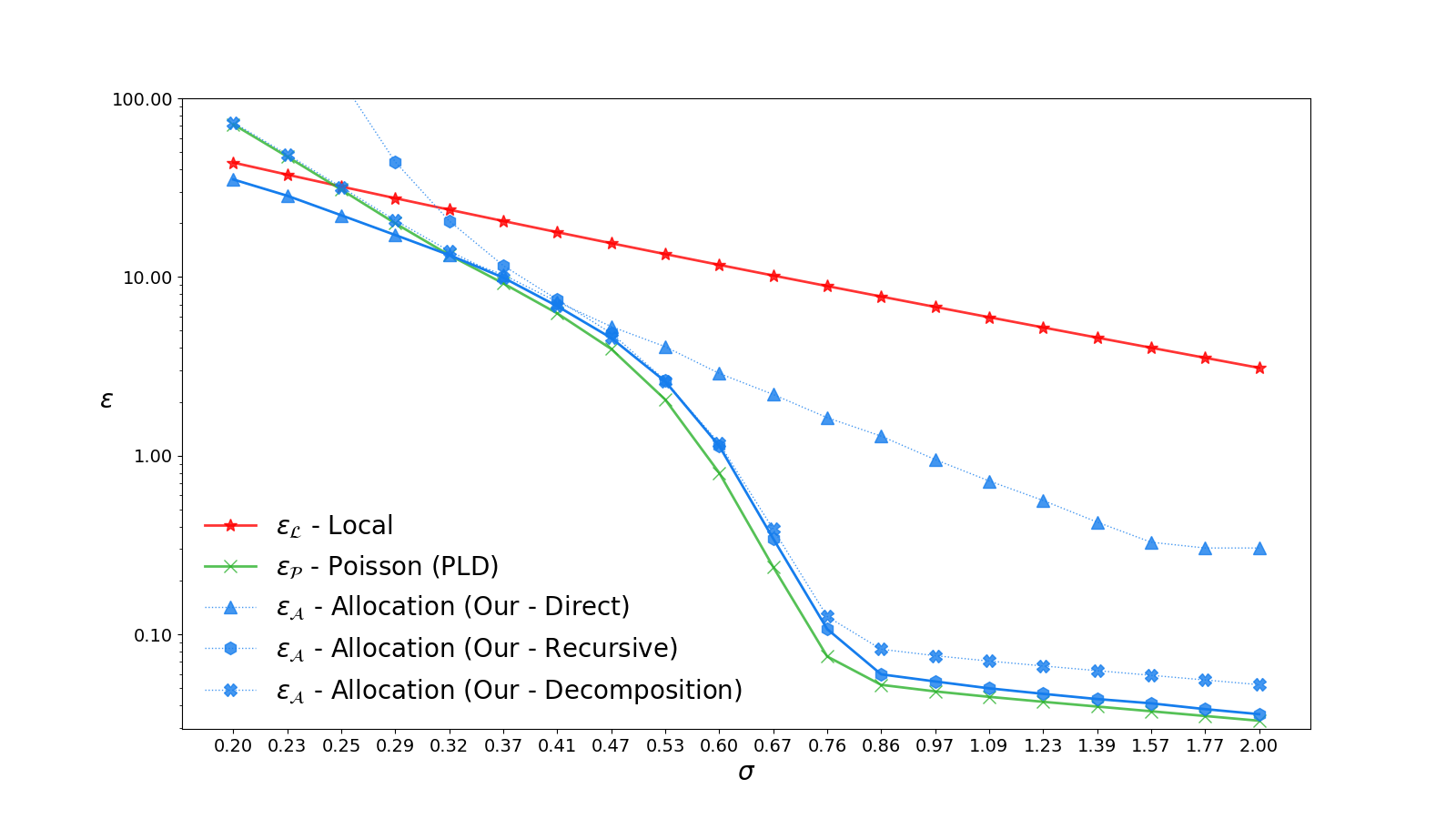}
            \caption{\small Remove direction}
            \label{fig:main_add_remove_sub2}
        \end{subfigure}
    }
    \caption{\small Upper bounds on privacy parameter $\eps$ or the add  and remove directions as a function of the noise parameter $\sigma$ for various schemes, all using the Gaussian mechanism with fixed parameters $\delta = 10^{-10}$, $t = 10^{6}$, the same setting as Figure \ref{fig:main}}
    \label{fig:main_add_remove}
\end{figure}

\begin{figure}[htbp]
    \centering
    \makebox[\textwidth][c]{
        \begin{subfigure}{0.6\textwidth}
            \centering
            \includegraphics[width=\textwidth]{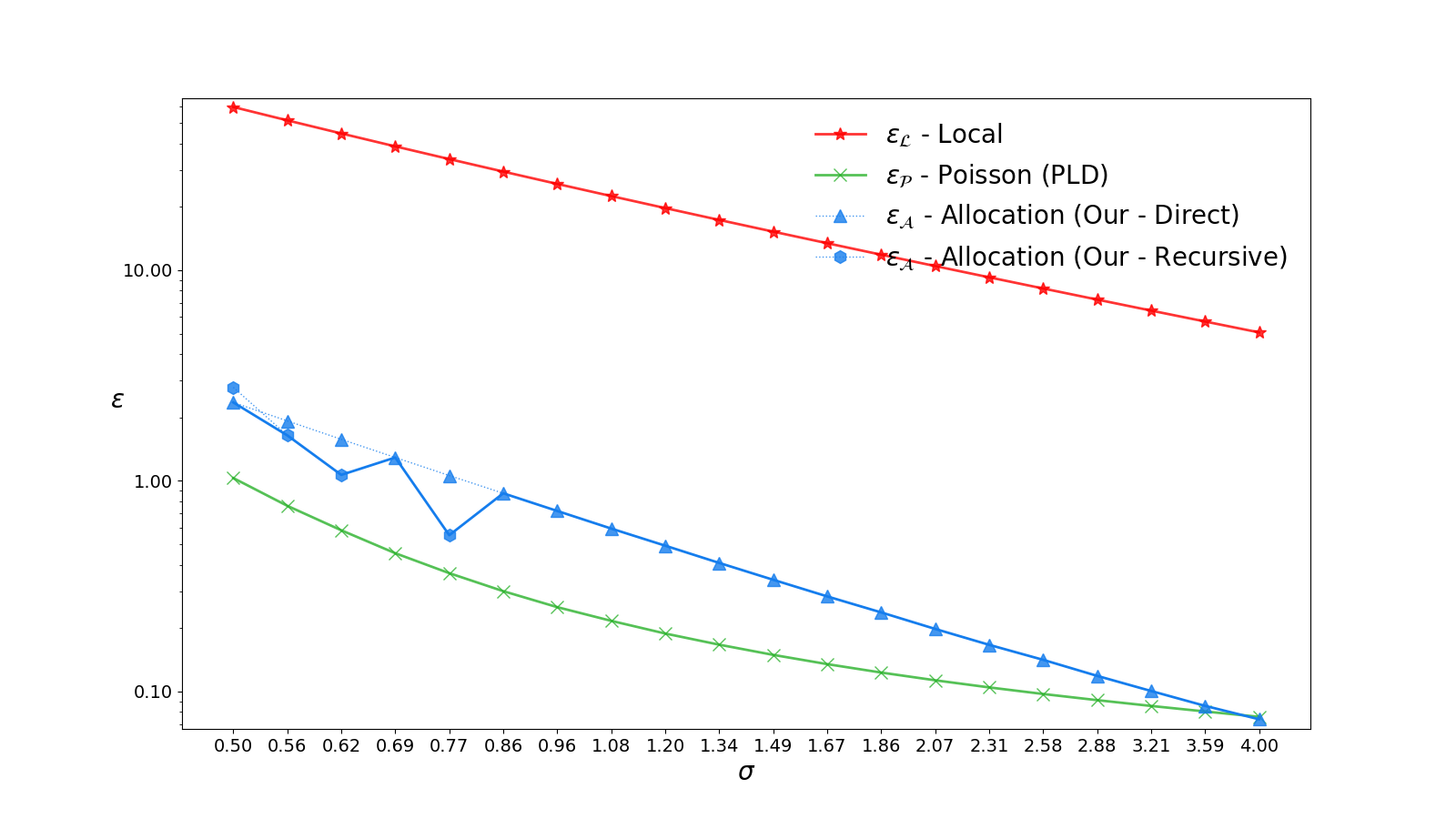}
            \caption{\small Add direction}
            \label{fig:main_several_selected_add_remove_sub1}
        \end{subfigure}
        \begin{subfigure}{0.6\textwidth}
            \centering
            \includegraphics[width=\textwidth]{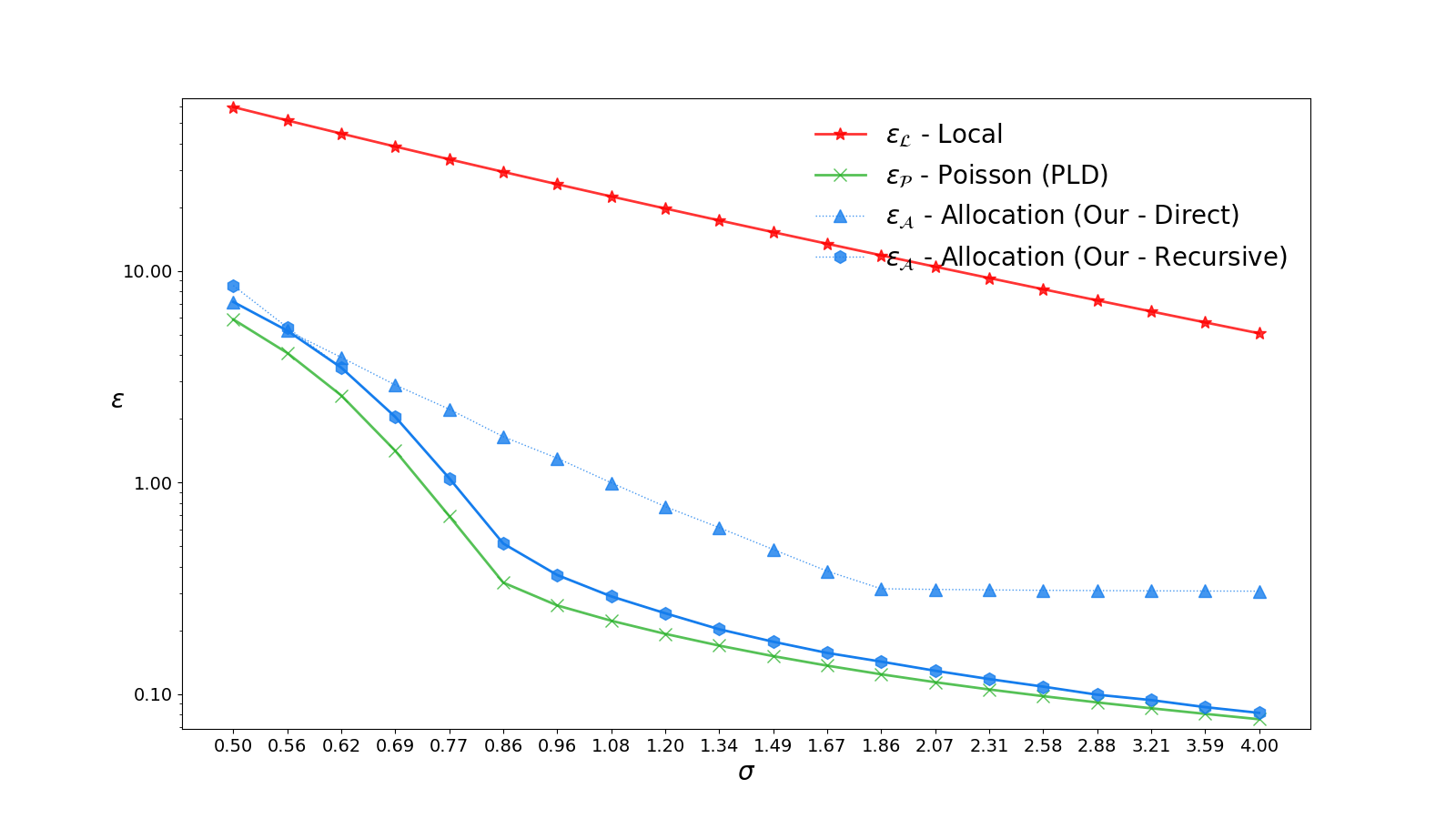}
            \caption{\small Remove direction}
            \label{fig:main_several_selected_add_remove_sub2}
        \end{subfigure}
    }
    \caption{\small Upper bounds on privacy parameter $\eps$ or the add  and remove directions as a function of the noise parameter $\sigma$ for various schemes, all using the Gaussian mechanism with fixed parameters $\delta = 10^{-10}$, $t = 10^{6}$, $k = 10$, the same setting as Figure \ref{fig:main}}
    \label{fig:main_several_selected_add_remove}
\end{figure}

\section{Asymptotic analysis}\label{apd:asym}
We start by recalling the asymptotic bounds for the Poisson scheme due to \citet{ACGMMT16}.\footnote{This is a variant of \citet[Theorem 1]{ACGMMT16} that is better suited for comparison. We prove this version in Appendix \ref{apd:analBnd}.} 
\begin{lemma}[\citep{ACGMMT16}] \label{lem:poisPriv}
    There exists constants $c_{1}, c_{2} > 0$ such that for any $t \in \naturals$; $\lambda \in [0, 1/16]$; $\delta \in [0,1]$, if $t \ge \ln(1/\delta)$ and $\sigma > \max\left\{1, c_{1} \frac{\sqrt{\ln(1/\delta})}{\lambda \sqrt{t}} \right\}$ then the Poisson scheme with the Gaussian mechanism  $\Pois{t, \lambda}{N_{\sigma}}$ is $(\eps, \delta)$-DP for any $\eps \ge c_{2} \max\left\{\frac{\lambda \sqrt{t \cdot \ln(1/\delta)}}{\sigma}, \lambda^{2} \sqrt{t \cdot \ln(1/\delta)} \right\}$.
\end{lemma}

This is a direct result of the fact the Gaussian mechanism is dominated by the one-dimensional Gaussian randomizer (Claim \ref{clm:domPairGauss}) where $\rand(*) = \mathcal{N}(1, \sigma^{2})$ and $\rand(\bot) = \mathcal{N}(0, \sigma^{2})$. Combining this Lemma with Corollary \ref{cor:GaussAnalBnd} implies a similar result for the random allocation scheme.
\begin{lemma}\label{lem:GaussAsymBnd}
	There exist constants $c_{1}, c_{2}$ such that for any $t \in \naturals$; $k \in [t/16]$; $\delta \in [0, 1]$; if    
    \[
        \sigma \ge c_{1} \cdot \max \left\{\sqrt{\ln(t/\delta)}, \sqrt{\frac{k}{t}} \ln(t/\delta), \frac{\sqrt{t \cdot \ln(1/\delta) \cdot \ln(t/k)}}{k} \right\},
    \]
    then the random allocation scheme with the Gaussian mechanism  $\alloc{t, k}{N_{\sigma}}$ is $(\eps, \delta)$-DP for any $\eps \ge c_{2} \max\left\{\frac{k \sqrt{\ln(1/\delta)}}{\sigma \sqrt{t}}, \frac{k^{2} \sqrt{\ln(1/\delta)}}{t^{1.5}}\right\}$.
\end{lemma}

The second term in the bound on $\eps$ is due to the privacy profile of the Poisson scheme, and applies only in the uncommon regime when $\sigma > t/k$. One important difference between the privacy guarantees of the Poisson and random allocation schemes is in the bounds on $\sigma$, which are stricter for random allocation in the $k > \sqrt{t}$ regime (Remark \ref{rem:asymComp}).

The proof of this Lemma is based on the identity of the dominating pair of the Gaussian mechanism.
\begin{claim}[Gaussian randomizer \citep{ACGMMT16}] \label{clm:domPairGauss}
    Given $\sigma > 0$, the Gaussian mechanism $N_{\sigma}$ is tightly dominated by the pair of distributions $(\mathcal{N}(1, \sigma^{2}), \mathcal{N}(0, \sigma^{2}))$, which induce a Gaussian randomizer where $* \coloneqq 1$ and $\bot \coloneqq 0$. This pair can be realized by datasets of arbitrary size $n$ of vectors in dimension $d$ by the pair $((\overset{n-1~\text{times}}{\overbrace{\bar{0}, \ldots, \bar{0}}}, e_{1}), (\overset{n~\text{times}}{\overbrace{\bar{0}, \ldots, \bar{0}}}))$.
\end{claim}
We note that the dominating pair of the Gaussian is one dimensional, regardless of the dimension of the original algorithm.

\begin{proof}[Proof of Lemma \ref{lem:GaussAsymBnd}]
    From Theorem \ref{\mainThmAsym}, each of the schemes has a privacy profile $\privProf{\alloc{t/k}{\rand}}(\eps) \le \privProf{\Pois{t/k, \eta}{\rand}}(\eps) + t/k \delta_{0} + \delta/k$. Applying the union bound to the $t/k \delta_{0}$ and $\delta/k$ terms, and using the fact that the composition of Poisson schemes is a longer Poisson scheme completes the proof of the first part.

    From Lemma \ref{lem:gaussPriv} we have $\eps_0 = \frac{\sqrt{2\ln(1.25/ \delta_{0})}}{\sigma} = \frac{\sqrt{2\ln(1.25 t/ \delta)}}{\sigma}$ (see e.g., \citet{DR14} for exact derivation). From the first bound on $\sigma$ we get $\eps_{0} \le 1$ and therefore $\cosh(\eps_0) = (e^{\eps_0}-e^{\eps_0})/2 \leq 3 \eps_0/2$. Combining this with the second bound on $\sigma$ we get,
    \[
        \gamma \le 3 \eps_0 \sqrt{\frac{k}{2t} \ln \left(\frac{k}{\delta} \right)} \le 3 \frac{\sqrt{2\ln(1.25t/\delta)}}{\sigma} \sqrt{\frac{k}{2t} \ln \left(\frac{k}{\delta} \right)} \le \frac{3\sqrt{k} \ln(1.25t/\delta)}{\sqrt{t}\sigma} \le 1/2,
    \]
    which implies $\eta \leq \frac{2k}{t}$ and $\privProf{\Pois{t,\eta}{N_{\sigma}}}(\eps) \le \privProf{\Pois{t, 2 k / t}{N_{\sigma}}}(\eps)$, since the Poisson scheme's privacy profile is monotonic in the sampling probability as proven in Lemma \ref{lem:truncBndPois}.
\end{proof}

\begin{remark} \label{rem:asymComp}
    While the asymptotic bound on $\eps$ for the Poisson and random allocation schemes is identical up to the additional logarithmic dependence on $t$, only the third bound on $\sigma$ stated for random allocation is required for Poisson. Notice that if $\sqrt{t} > k$ the third term upper bounds the first one, and if additionally $\ln(1/\delta) \le \frac{t^{2}}{k^{3}}$ the second term is bounded by the third one as well. While the first condition might not hold when each element is allocated to many steps, the latter does not hold only when $t < \ln^{2}(1 / \delta)$ which is an uncommon regime of parameters.
\end{remark}

\section{Missing proofs from Section \ref{sec:PoisDecom}} \label{apd:PoisDecom}
\begin{lemma}\label{lem:allocMontn}
        Given $1 \le k \le k' \le t$ and a randomizer $\rand$ dominated by a randomizer $\rand$ we have $\privProfRem{\alloc{t, k}{\rand}}(\eps) \le \privProfRem{\alloc{t, k'}{\rand}}(\eps)$ and $\privProfAdd{\alloc{t, k}{\rand}}(\eps) \le \privProfAdd{\alloc{t, k'}{\rand}}(\eps)$. Furthermore, for any sequence of integers $k \le k_{1} < \ldots < k_{j} \le t$, and non-negative $\lambda_{1}, \ldots, \lambda_{j}$ s.t. $\lambda_{1} + \ldots + \lambda_{j} = 1$, the privacy profile of $\alloc{t, k}{\rand}$ is upper-bounded by the privacy profile of $\lambda_{1} \alloc{t, k_{1}}{\rand} + \ldots + \lambda_{j} \alloc{t, k_{j}}{\rand}$, where we use convex combinations of algorithms to denote an algorithm that randomly chooses one of the algorithms with probability given in the coefficient.
\end{lemma}

\begin{proof}
    To prove this claim, we recall the technique used in the proof of Lemma \ref{lem:analBnd}. We proved in Lemma \ref{lem:postEqAlloc} that $\allocFunc{t, k}{\rand}{*}$ and $\postFunc{t, k}{\rand}{*}$ are identically distributed. From the non-adaptivity assumption, this is just a sequence of repeated calls to the mixture algorithm $\lambda_{\view^{i}, k, *} \cdot \rand(*) + (1-\lambda_{\view^{i}, k, *}) \cdot \rand(\bot)$.
    
    Next we recall the fact proven in Lemma \ref{lem:truncBndPois} that the hockey-stick divergence between this mixture algorithm and $\rand(\bot)$ is monotonically increasing in $\lambda$. Since $\lambda_{\view^{i}, k', *} \ge \lambda_{\view^{i}, k, *}$ for any $k' > k$, this means the pair of distributions $(\lambda_{\view^{i}, k', *} \cdot \rand(*) + (1-\lambda_{\view^{i}, k', *}) \cdot \rand(\bot), \rand(\bot))$ dominates the pair $(\lambda_{\view^{i}, k', *} \cdot \rand(*) + (1-\lambda_{\view^{i}, k', *}) \cdot \rand(\bot), \rand(\bot))$ for any iteration $i$ and view $\view^{i}$ (this domination holds in both directions). Using Claim \ref{clm:domPairComp} this implies we can iteratively apply this for all step and get $\privProf{\alloc{t, k}{\rand}}(\eps) \le \privProf{\alloc{t, k'}{\rand}}(\eps)$ for any $\eps > 0$, thus completing the proof of the first part.

    The proof of the second part is identical, since the posterior sampling probability induced by any mixture of $\alloc{t, k_{1}}{\rand},  \ldots, \alloc{t, k_{j}}{\rand}$ is greater than the one induced by $\alloc{t, k}{\rand}$ the same reasoning follows.
\end{proof}

\begin{lemma}\label{lem:PoisDecom}
    For any $\lambda \in [0, 1]$, element $x \in \domain$, and randomizer $\rand$ we have, 
    \[
        \PoisFunc{t, \lambda}{\rand}{x} = \sum_{k = 0}^{t} B_{t, \lambda}(k) \cdot \allocFunc{t, k}{\rand}{x}, 
    \]
    where $B_{t, \lambda}$ is the PDF of the binomial distribution with parameters $t, \lambda$ and $\allocFunc{t, 0}{\rand}{x} \coloneqq \rand^{\otimes t}(\bot)$.
\end{lemma}

\begin{proof}
    This results from the fact that flipping $t$ coins with bias $\lambda$ can be modeled as first sampling an integer $k \in \{0, 1, \ldots, t\}$ from a binomial distribution with parameters $(t, \lambda)$, then uniformly sampling $i_{1}, \ldots, i_{k} \in [t]$, and setting the coins to $1$ for those indexes.
\end{proof}

The proof of the next claim is a generalization of the advance joint convexity property \citep[Theorem 2]{BBG18}.
\begin{claim}\label{clm:advJntCvx}
    Given $\lambda \in [0, 1]$; $\kappa > 1$ and two distribution $P, Q$ over some domain, we have
    \begin{align*}
        \HockeyStickDiv{\kappa}{(1-\lambda)Q + \lambda P}{Q}  & = \lambda \HockeyStickDiv{\kappa'}{P}{Q}
        \\ \HockeyStickDiv{\kappa}{P}{(1-\lambda)P + \lambda Q} & = \begin{cases}
            \beta \HockeyStickDiv{\kappa''}{P}{Q} & \kappa \le \frac{1}{1-\lambda} \\
            0 & \kappa > \frac{1}{1-\lambda}
        \end{cases},
    \end{align*}
    where $\kappa' \coloneqq 1 + \frac{\kappa - 1}{\lambda}$, $\kappa'' \coloneqq 1 + \frac{\kappa - 1}{1 - \kappa + \kappa \lambda}$, and $\beta \coloneqq 1 - \kappa + \kappa \lambda$.
\end{claim}

\begin{proof}
    The identity for $\HockeyStickDiv{\kappa}{(1-\lambda)P + \lambda Q}{P}$ is a direct result of the advanced joint convexity property \citep[Theorem 2]{BBG18}. For the second part notice that,
    \begin{align*}
        \HockeyStickDiv{\kappa}{P}{(1-\lambda)P + \lambda Q} & = \int_{\Omega} \left[P(\omega) - \kappa \left((1-\lambda)P(\omega)+ \lambda Q(\omega) \right)\right]_{+} d \omega
        \\ & = \int_{\Omega} \left[(1 - \kappa + \kappa \lambda)P(\omega) -  \lambda \kappa Q(\omega) \right]_{+} d \omega
        \\ & = \begin{cases}
            \beta \int_{\Omega} \left[P(\omega) - \kappa'' Q(\omega) \right]_{+} d \omega & \alpha \le \frac{1}{1-\lambda}
            \\ 0 & \kappa > \frac{1}{1-\lambda}
        \end{cases}
        \\ & = \begin{cases}
            \beta \HockeyStickDiv{\kappa''}{P}{Q} & \kappa \le \frac{1}{1-\lambda}
            \\ 0 & \kappa > \frac{1}{1-\lambda}
        \end{cases}.
    \end{align*}
\end{proof}

\begin{lemma}\label{lem:PoisUpper}
    For any $\lambda \in [0, 1]$; $\eps > 0$ and randomizer $\rand$ we have
    \begin{align*}
        \HockeyStickDiv{e^{\eps}}{\mathcal{P}_{t, \lambda}^{+}(\rand; *)}{\mathcal{P}_{t, \lambda}^{+}(\rand; \bot)} & = \rem{\gamma} \HockeyStickDiv{e^{\rem{\eps}}}{\PoisFunc{t, \lambda}{\rand}{*}}{\PoisFunc{t, \lambda}{\rand}{\bot}}
        \\ \HockeyStickDiv{e^{\eps}}{\mathcal{P}_{t, \lambda}^{+}(\rand; \bot)}{\mathcal{P}_{t, \lambda}^{+}(\rand; *)} & = \add{\gamma} \HockeyStickDiv{e^{\add{\eps}}}{\PoisFunc{t, \lambda}{\rand}{\bot}}{\PoisFunc{t, \lambda}{\rand}{*}},
    \end{align*}
    where 
    \[
        \mathcal{P}_{t, \lambda}^{+}(\rand; x) = \rem{\gamma} \sum_{k \in [t]} B_{t, \lambda}(k) \cdot \allocFunc{t, k}{\rand}{x}
    \]
    is the Poisson scheme conditioned on allocating the element at least once, and $\rem{\gamma}$, $\add{\gamma}$, $\rem{\eps}$, and $\add{\eps}$ were defined in Theorem \ref{thm:dcmpBnd}.
\end{lemma}

\begin{proof}
    First notice that, 
    \begin{align*}
        \PoisFunc{t, \lambda}{\rand}{x} & \eqExp{1}{=} \sum_{k = 0}^{t} B_{t, \lambda}(k) \cdot \allocFunc{t, k}{\rand}{x}
        \\ & = B_{t, \lambda}(0) \cdot \allocFunc{t, 0}{\rand}{x} + \sum_{k \in [t]} B_{t, \lambda}(k) \cdot \allocFunc{t, k}{\rand}{x}
        \\ & \eqExp{2}{=} \left(1 - \lambda \right)^{t} \cdot \allocFunc{t, 0}{\rand}{x} + \sum_{k \in [t]} B_{t, \lambda}(k) \cdot \allocFunc{t, k}{\rand}{x}
        \\ & \eqExp{3}{=} (1 - 1/\rem{\gamma}) \cdot \rand^{\otimes t}(\bot) + 1/\rem{\gamma} \cdot \mathcal{P}_{t, \lambda}^{+}(\rand; x),
    \end{align*}
    where (1) results from Lemma \ref{lem:PoisDecom}, (2) from the definition of the binomial distribution, $\lambda'$ and $\mathcal{P}_{t, \lambda}^{+}(\rand)$, and (3) from the definition of $\lambda'$ and the fact $\allocFunc{t, 0}{\rand}{x} = \rand^{\otimes t}(\bot)$.

    Taking the converse of Claim \ref{clm:advJntCvx} we have, 
    \[
        \HockeyStickDiv{e^{\eps}}{\mathcal{P}_{t, \lambda}^{+}(\rand; *)}{\mathcal{P}_{t, \lambda}^{+}(\rand; \bot)} = \rem{\gamma} \HockeyStickDiv{e^{\rem{\eps}}}{\PoisFunc{t, \lambda}{\rand}{*}}{\PoisFunc{t, \lambda}{\rand}{\bot}}.
    \]
    Similarly, from the previous claim we get,
    \[
        \HockeyStickDiv{e^{\eps}}{\mathcal{P}_{t, \lambda}^{+}(\rand; \bot)}{\mathcal{P}_{t, \lambda}^{+}(\rand; *)} = \add{\gamma} \HockeyStickDiv{e^{\add{\eps}}}{\PoisFunc{t, \lambda}{\rand}{\bot}}{\PoisFunc{t, \lambda}{\rand}{*}}.
    \]    
    which completes the proof.
\end{proof}

\NeurIPS{We can now prove the main theorem.}
\showstored{mov:dcmpBnd}

\begin{proof}[Proof Theorem \ref{thm:dcmpBnd}]
    The proof directly results from combining the previous lemmas,
    \begin{align*}
        \privProfRem{\alloc{t}{\rand}}(\eps) & = \HockeyStickDiv{e^{\eps}}{\allocFunc{t, k}{\rand}{*}}{\allocFunc{t, k}{\rand}{\bot}}
        \\ & \eqExp{1}{=} \HockeyStickDiv{e^{\eps}}{\rem{\gamma} \sum_{k \in [t]} B_{t, \lambda}(k) \cdot \allocFunc{t, 1}{\rand}{*}}{\allocFunc{t, k}{\rand}{\bot}}
        \\ & \eqExp{2}{\le} \HockeyStickDiv{e^{\eps}}{\rem{\gamma} \sum_{k \in [t]} B_{t, \lambda}(k)\cdot \allocFunc{t, k}{\rand}{*}}{\allocFunc{t, k}{\rand}{\bot}}
        \\ & \eqExp{3}{=} \HockeyStickDiv{e^{\eps}}{\mathcal{P}_{t, \lambda}^{+}(\rand; *)}{\mathcal{P}_{t, \lambda}^{+}(\rand; \bot)}
        \\ & \eqExp{4}{=} \rem{\gamma} \HockeyStickDiv{e^{\rem{\eps}}}{\PoisFunc{t, \lambda}{\rand}{*}}{\PoisFunc{t, \lambda}{\rand}{\bot}}
        \\ & = \rem{\gamma} \cdot \privProfRem{\Pois{t, \lambda}{\rand}}(\rem{\eps}),
    \end{align*}
    where (1) results from the fact $\sum_{k \in [t]} B_{t, \lambda}(k) = 1/\rem{\gamma}$, (2) from Lemma \ref{lem:allocMontn}, (3) from the definition of $\mathcal{P}_{t, \lambda}^{+}$ and the fact that $\PoisFunc{t, \lambda}{\rand}{\bot} = \rand^{\otimes t}(\bot) = \allocFunc{t, \lambda}{\rand}{\bot}$, and (4) from the first part of Lemma \ref{lem:PoisUpper}.
    
    Repeating the same proof using the second part of Lemma \ref{lem:PoisUpper} proves the bound on $\privProfAdd{\alloc{t}{\rand}}$.
\end{proof}

\begin{proof}\Arxiv{[Proof of Corollary \ref{cor:TruncPois}]}
    Notice that,
    \begin{align*}
        \privProf{\Pois{t, \lambda, k}{\rand}}(\eps) & \eqExp{1}{\le} \privProf{\Pois{t, \lambda, k}{\rand}}(\eps)
        \\ & = \HockeyStickDiv{e^{\eps}}{\PoisFunc{t, \lambda, k}{\rand}{*}}{\PoisFunc{t, \lambda, k}{\rand}{\bot}}
        \\ & \eqExp{2}{=} \HockeyStickDiv{e^{\eps}}{\left(\sum_{i = 0}^{k-1} B_{t, \lambda}(i) \allocFunc{t, i}{\rand}{*} \right) + \left(\sum_{i = k}^{t} B_{t, \lambda}(i) \right) \allocFunc{t, k}{\rand}{*}}{\PoisFunc{t, \lambda, k}{\rand}{\bot}}
        \\ & \eqExp{3}{\le} \HockeyStickDiv{e^{\eps}}{\sum_{i = 0}^{t} B_{t, \lambda}(i) \allocFunc{t, i}{\rand}{*}}{\PoisFunc{t, \lambda}{\rand}{\bot}}
        \\ & = \privProf{\Pois{t, \lambda}{\rand}}(\eps),
    \end{align*}
    where (1) results from Lemma \ref{lem:singElem}, (2) from Lemma \ref{lem:PoisDecom} and the definition of $\Pois{t, \lambda, k}{\rand}$, and (3) from Lemma \ref{lem:allocMontn}.
\end{proof}

\section{Missing proofs from Section \ref{sec:dirAnals}} \label{apd:dirAnals}
\begin{lemma}\label{lem:dirRemBnd}
    Given $t, \alpha \in \naturals$ and two distributions $P, Q$ over some domain $\Omega$, we have
    \[
        \RenyiDiv{\alpha}{\bar{P}}{Q^{\otimes t}} = \frac{1}{\alpha-1} \ln\left(\frac{1}{t^{\alpha}} \sum_{\Pi \in \boldsymbol{\Pi}_{t}(\alpha)} \binom{t}{C(\Pi)} \binom{\alpha}{\Pi} \prod_{p \in \Pi} e^{(\alpha-1) \RenyiDiv{p}{P}{Q}} \right),
    \]
    where $\bar{P} \coloneqq \frac{1}{t} \sum_{i \in [t]} \left(Q^{\otimes (i-1)} \cdot P \cdot Q^{\otimes (t-i)} \right)$, $Q^{\otimes (i-1)} \cdot P \cdot Q^{t-i}$ denotes the distribution induced by sampling all elements from $Q$, except for the $i$th one which is sampled from $P$.
\end{lemma}
We start by proving a supporting claim.
\begin{claim} \label{clm:multiCount}
    Given $\alpha, t \in \naturals$  and a list of integers $i_{1}, \ldots, i_{t} \ge 0$ such that $i_{1} + \ldots + i_{t} = \alpha$, denote by $P(i_{1}, \ldots, i_{t})$ the integer partition of $\alpha$ associated with this list, e.g. if $i_{1} = 1, i_{2} = 0, i_{3} = 2, i_{4} = 1$, then $P = [1, 1, 2]$. Given an integer partition $P$ of $\alpha$, we have $\vert B_{P} \vert = \binom{t}{C(\Pi)}$ where, 
    \[
        B_{P} = \left\{i_{1}, \ldots, i_{t} \ge 0 ~\vert~ P(i_{1}, \ldots, i_{t}) = P \right\},
    \]
    and $C(\Pi)$ was defined in Theorem \ref{thm:dirRemBnd}.
\end{claim}

\begin{proof}
    Given a partition $P$ with unique counts $C(\Pi) = (c_{1}, \ldots, c_{j})$, and an assignments $i_{1}, \ldots, i_{t}$ such that $i_{1}, \ldots, i_{t} \ge 0$ and $P(i_{1}, \ldots, i_{t}) = P$, there are $\binom{t}{c_{1}}$ ways to assign the first value to $c_{1}$ indexes of the possible $t$, $\binom{t - c_{1}}{c_{2}}$ ways to assign the second value to $c_{2}$ indexes of of the remaining $t - c_{1}$ indexes, and so on. Multiplying these terms completes the proof.
\end{proof}

\begin{proof} [Proof of Lemma \ref{lem:dirRemBnd}]
    Given a set of integers $i_{1}, \ldots, i_{t} \ge 0$ such that $i_{1} + \ldots + i_{t} = \alpha$ 
    we have,
    \[
        \prod_{k \in [t]} \expect{\viewRV \sim Q^{\otimes t}}{\left(\frac{P(\omega)}{Q(\omega)} \right)^{i_{k}}} = \prod_{p \in \Pi} \expect{\outRV \sim Q}{\left(\frac{P(\omega)}{Q(\omega)} \right)^{p}},
    \]
    where $P$ is the integer partition of $\alpha$ defined by $i_{1}, \ldots, i_{t}$, e.g. if $i_{1} = 1, i_{2} = 0, i_{3} = 2, i_{4} = 1$, then $P = [1, 1, 2]$. This is a result of the fact $\outRV_{k}$ are all identically distributed. Notice that the same partition corresponds to many assignments, e.g. $P = [1, 1, 2]$ corresponds to $i_{1} = 0, i_{2} = 1, i_{3} = 1, i_{4} = 2$ as well. The number of assignments that correspond to a partition $P$ is $\binom{t}{C(\Pi)}$. Using this fact we get,
    \begin{align*}
        e^{(\alpha-1)\RenyiDiv{\alpha}{\bar{P}}{Q^{\otimes t}}} & \eqExp{1}{=} \expect{\viewRV \sim Q^{\otimes t}}{\left(\frac{\frac{1}{t} \sum_{i \in [t]} \left(Q^{\otimes (i-1)} \cdot P \cdot Q^{\otimes (t-i)} \right)(\viewRV)}{Q^{\otimes t}(\viewRV)} \right)^{\alpha}}
        \\ & = \expect{\viewRV \sim Q^{\otimes t}}{\left(\frac{1}{t}\sum_{i \in [t]} \frac{P(\omega_{i})}{Q(\omega_{i})} \right)^{\alpha}}
        \\ & \eqExp{2}{=} \frac{1}{t^{\alpha}} \expect{\viewRV \sim Q^{\otimes t}}{\sum_{\substack{i_{1}, \ldots, i_{t} \in [\alpha]; \\ i_{1} + \ldots + i_{t} \ge 0}} \binom{\alpha}{i_{1}, \ldots, i_{t}} \prod_{k \in [t]} \left(\frac{P(\omega_{i})}{Q(\omega_{i})} \right)^{i_{k}}}
        \\ & \eqExp{3}{=} \frac{1}{t^{\alpha}} \sum_{\substack{i_{1}, \ldots, i_{t} \ge 0; \\ i_{1} + \ldots + i_{t} = \alpha}} \binom{\alpha}{i_{1}, \ldots, i_{t}} \prod_{k \in [t]} \expect{\viewRV \sim Q^{\otimes t}}{\left(\frac{P(\omega_{i})}{Q(\omega_{i})} \right)^{i_{k}}}
        \\ & \eqExp{4}{=} \frac{1}{t^{\alpha}} \sum_{\substack{i_{1}, \ldots, i_{t} \ge 0; \\ i_{1} + \ldots + i_{t} = \alpha}} \binom{\alpha}{i_{1}, \ldots, i_{t}} \prod_{p \in \Pi(i_{1}, \ldots, i_{t})} \expect{\omega \sim Q}{\left(\frac{P(\omega)}{Q(\omega)} \right)^{p}}
        \\ & \eqExp{5}{=} \frac{1}{t^{\alpha}} \sum_{\Pi \in \boldsymbol{\Pi}_{t}(\alpha)} \binom{t}{C(\Pi)} \binom{\alpha}{\Pi} \prod_{p \in \Pi} \expect{\omega \sim Q}{\left(\frac{P(\omega)}{Q(\omega)} \right)^{p}}
        \\ & = \frac{1}{t^{\alpha}} \sum_{\Pi \in \boldsymbol{\Pi}_{t}(\alpha)} \binom{t}{C(\Pi)} \binom{\alpha}{\Pi} \prod_{p \in \Pi} e^{(\alpha-1) \RenyiDiv{p}{P}{Q}},
    \end{align*}
    where (1) results from the definition of $\bar{P}$, (2) is the multinomial theorem, (3) results from the fact $\omega_{i}$ and $\omega_{j}$ are independent for any $i \ne j$, (4) from the fact $\omega_{k}$ are all identically and independently distributed with $P(i_{1}, \ldots, i_{t})$ defined in Claim \ref{clm:multiCount}, and (5) results from Claim \ref{clm:multiCount}.
\end{proof}

\begin{lemma}\label{lem:dirAddBnd}
    Given $\lambda \in [0,1]$ and two distributions $P, Q$ over some domain $\Omega$, denote $P_{\lambda} \coloneqq \frac{P^{\lambda} Q^{1-\lambda}}{Z_{\lambda}}$ where $Z_{\lambda}$ is the normalizing factor.

    Given $t \in \naturals$, for any $\eps \in \reals$ we have $\HockeyStickDiv{e^{\eps}}{Q^{\otimes t}}{\bar{P}} \le \HockeyStickDiv{e^{\eps'}}{Q^{\otimes t}}{P_{1/t}^{\otimes t}}$, where $\bar{P} \coloneqq \frac{1}{t} \sum_{i \in [t]} \left(Q^{\otimes (i-1)} \cdot P \cdot Q^{\otimes (t-i)} \right)$, $Q^{\otimes (i-1)} \cdot P \cdot Q^{t-i}$ denotes the distribution induced by sampling all elements from $Q$, except for the $i$th one which is sampled from $P$.
\end{lemma}

\begin{proof}
    By definition,
    \[
        \loss{\omega}{P_{\lambda}}{Q} = \ln\left(\frac{P_{\lambda}(\omega)}{Q(\omega)}\right) = \ln\left(\frac{P^{\lambda}(\omega) Q^{1-\lambda}(\omega)}{Q(\omega) Z_{\lambda}}\right) = \lambda \ln\left(\frac{P(\omega)}{Q(\omega)}\right) - \ln(Z_{\lambda}) = \lambda \loss{\omega}{P}{Q} - \ln(Z_{\lambda}).
    \]

    Given $\eps \in \reals$ denote $\eps' \coloneqq \eps - t \cdot \ln(Z_{1/t})$
    \begin{align*}
        \HockeyStickDiv{e^{\eps}}{Q^{\otimes t}}{\bar{P}} & = \int_{\Omega^{t}} \left[Q^{\otimes t}(\view) - e^{\eps} \bar{P}(\view) \right]_{+} d\view
        \\ & = \int_{\Omega^{t}} \left[Q^{\otimes t}(\view) - e^{\eps} \frac{1}{t} \sum_{i \in [t]} \left(Q^{\otimes (i-1)} \cdot P \cdot Q^{\otimes (t-i)} \right)(\view) \right]_{+} d\view
        \\ & = \int_{\Omega^{t}} Q^{\otimes t}(\view) \left[1 - e^{\eps} \frac{1}{t} \sum_{i \in [t]} e^{\loss{\omega_{i}}{P}{Q}} \right]_{+} d\view
        \\ & \eqExp{1}{\le} \int_{\Omega^{t}} Q^{\otimes t}(\view) \left[1 - e^{\eps + \frac{1}{t} \sum_{i \in [t]} \loss{\omega_{i}}{P}{Q}} \right]_{+} d\view
        \\ & \eqExp{2}{=} \int_{\Omega^{t}} Q^{\otimes t}(\view) \left[1 - e^{\eps' + \sum_{i \in [t]} \loss{\omega_{i}}{P_{1/t}}{Q}} \right]_{+} d\view
        \\ & = \int_{\Omega^{t}} Q^{\otimes t}(\view) \left[1 - e^{\eps'}\left(\prod_{i \in [t]} \frac{P_{1/t}(\omega_{i})}{Q(\omega_{i})} \right) \right]_{+} d\view
        \\ & = \int_{\Omega^{t}} \left[Q^{\otimes t}(\view) - e^{\eps'} P_{1/t}^{\otimes t}(\view) \right]_{+} d\view
        \\ & = \HockeyStickDiv{e^{\eps'}}{Q^{\otimes t}}{P_{1/t}^{\otimes t}}
    \end{align*}
    where (1) results from Jensen's inequality and (2) from the definition of $P_{1/t}$ and the previous claim.
\end{proof}

\begin{proof}[Proof of Corollary \ref{cor:dirGauss}]
    From the definition of the R\'{e}nyi divergence for the Gaussian mechanism, 
    \begin{align*}
        \Renyi{\alpha} \left(\allocFunc{t}{N_{\sigma}}{1} \Vert \allocFunc{t}{N_{\sigma}}{0} \right) & = \Renyi{\alpha} \left(\allocFunc{t}{N_{\sigma}}{1} \Vert N^{\otimes t}_{\sigma}(0) \right)
        \\ & = \frac{1}{\alpha - 1}\ln \left(\frac{1}{t^{\alpha}} \sum_{\Pi \in \boldsymbol{\Pi}_{t}(\alpha)} \binom{t}{C(\Pi)} \binom{\alpha}{\Pi} \prod_{p \in \Pi} e^{\RenyiDiv{p}{N_{\sigma}(1)}{N_{\sigma}(0)}} \right)
        \\ & = \frac{1}{\alpha - 1}\ln \left(\frac{1}{t^{\alpha}} \sum_{\Pi \in \boldsymbol{\Pi}_{t}(\alpha)} \binom{t}{C(\Pi)} \binom{\alpha}{\Pi} \prod_{p \in \Pi} e^{\frac{p (p-1)}{2 \sigma^{2}}} \right)
        \\ & = \frac{1}{\alpha - 1}\ln \left(\frac{e^{ - \frac{\alpha}{2 \sigma^{2}}}}{t^{\alpha}} \sum_{\Pi \in \boldsymbol{\Pi}_{t}(\alpha)} \binom{t}{C(\Pi)} \binom{\alpha}{\Pi} e^{\sum_{p \in \Pi}\frac{p^{2}}{2 \sigma^{2}}} \right)
        \\ & = -\frac{\alpha}{\alpha - 1} \left(\frac{1}{2 \sigma^{2}} + \ln \left(t\right) \right) + \frac{1}{\alpha - 1}\ln \left(\sum_{\Pi \in \boldsymbol{\Pi}_{t}(\alpha)} \binom{t}{C(\Pi)} \binom{\alpha}{\Pi} e^{\sum_{p \in \Pi}\frac{p^{2}}{2 \sigma^{2}}} \right).
    \end{align*}

    For the add direction we notice that from the definition, for any $x \in \reals$
    \[
        P^{\lambda}_{\rand}(x \vert *) P_{\rand}^{1-\lambda}(x \vert \bot) = \frac{1}{\sqrt{2 \pi} \sigma} e^{-\lambda \frac{(x - 1)^{2}}{2 \sigma^{2}} - (1-\lambda) \frac{x^{2}}{2 \sigma^{2}}} = \frac{1}{\sqrt{2 \pi} \sigma} e^{-\frac{(x-\lambda)^{2}}{2 \sigma^{2}} - \frac{\lambda (1-\lambda)}{2 \sigma^{2}}}
    \]
    so $\rand_{\lambda}(*) = \mathcal{N}(\lambda, \sigma^{2})$ and $Z_{\lambda} = e^{\frac{\lambda (1-\lambda)}{2 \sigma^{2}}}$.

    Setting $\lambda = 1/t$ we get,
    \begin{align*}
        \privProfAdd{\alloc{t}{N_{\sigma}}}(\eps) & \eqExp{1}{\le} \HockeyStickDiv{e^{\eps'}}{\mathcal{N}^{\otimes t}(0, \sigma^{2})}{\mathcal{N}^{\otimes t}(1/t, \sigma^{2})} 
        \\ & \eqExp{2}{=} \HockeyStickDiv{e^{\eps'}}{\mathcal{N}^{\otimes t}(0, t^{2} \sigma^{2})}{\mathcal{N}^{\otimes t}(1, t^{2} \sigma^{2})}
        \\ & \eqExp{3}{=} \HockeyStickDiv{e^{\eps'}}{\mathcal{N}(0, t \sigma^{2})}{\mathcal{N}(1, t \sigma^{2})},
    \end{align*}
    where (1) results from Theorem \ref{thm:dirAddBnd}, (2) from the fact that the hockey-stick divergence between two Gaussians with the same scale depends only on the ratio of the difference between their means and their scale, and (3) from the fact that the $t$-composition hockey-stick divergence between two Gaussians with the same scale amounts to dividing their scale by $\sqrt{t}$.
\end{proof}
We remark that the expression in Corollary \ref{cor:dirGauss} for the remove direction was previously computed in \citet{LT22}, up to the improvement of using integer partitions. In this (unpublished) work the authors give an incorrect proof that datasets $(0, \ldots, 0, 1)$ and $(0, \ldots, 0)$ are a dominating pair of datasets for the shuffle scheme applied to Gaussian mechanism. Their analysis of the RDP bound for this pair of distributions is correct (even if significantly longer) and the final expression is identical to ours. 

\begin{figure}[ht]
    \centering
    \includegraphics[width=1\linewidth]{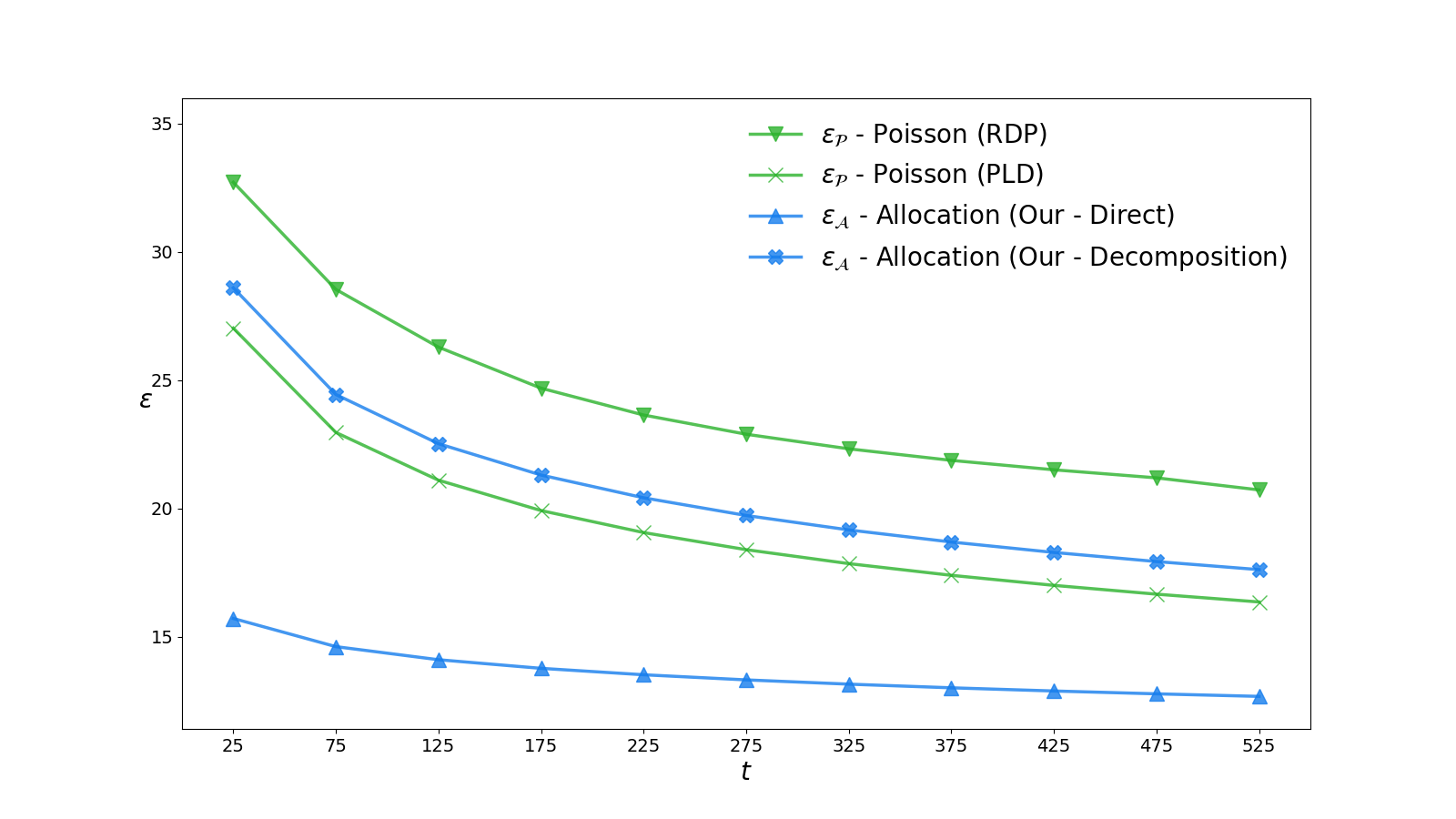}
    \caption{\small Upper bounds on privacy $\eps$ as a function of the number of steps $t$ for the Poisson and random allocation schemes, for fixed parameters $\sigma=0.3$, $\delta = 10^{-4}$.}
    \label{fig:RDP-dom}
\end{figure}

\begin{figure}[ht]
    \centering
    \includegraphics[width=1\linewidth]{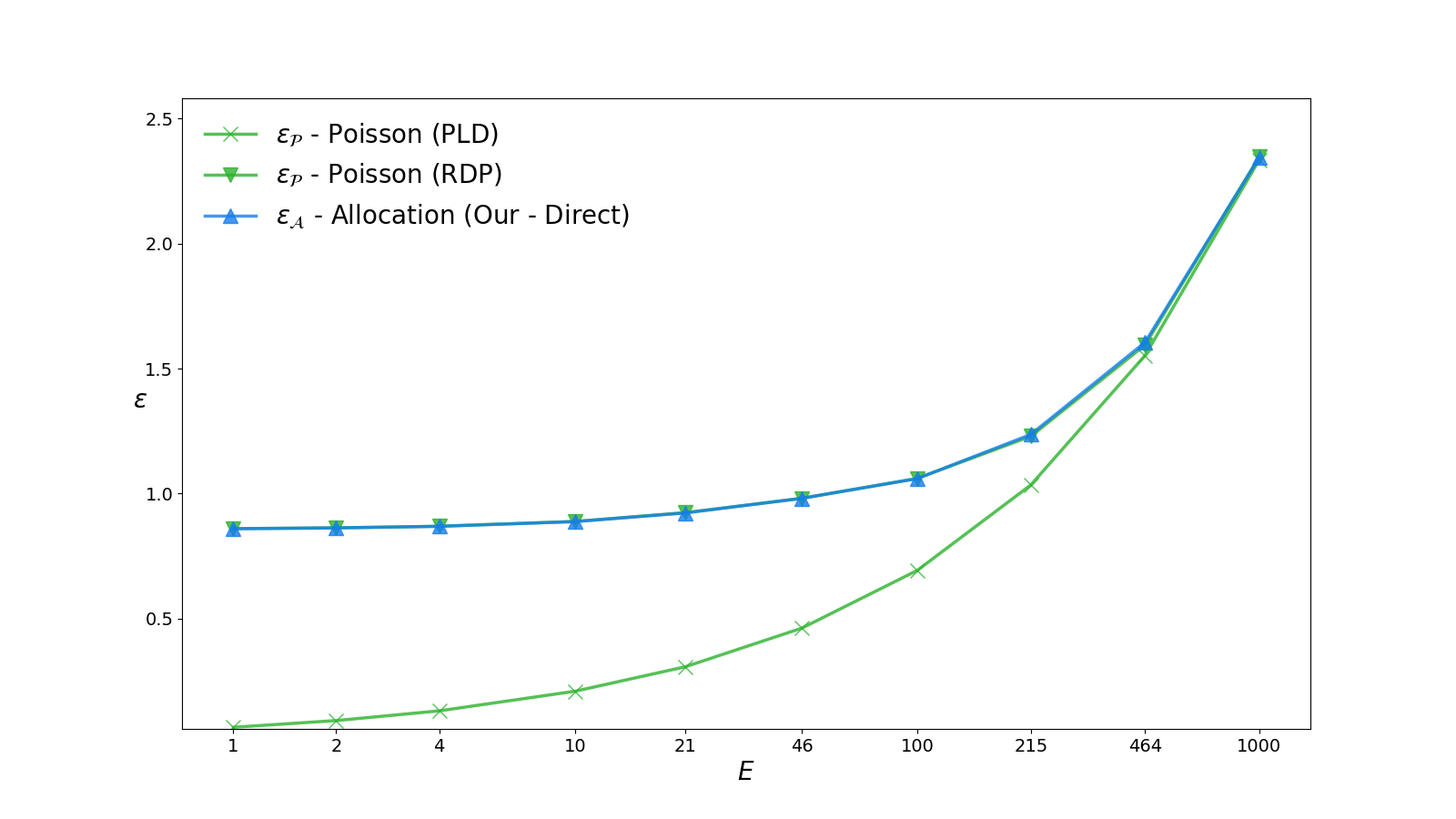}
    \caption{\small Upper bounds on privacy parameter $\eps$ for various schemes all using the Gaussian mechanism, as a function of $E$ the number of ``epochs'' - times the scheme was sequentially computed, for fixed parameters $\sigma=1$, $\delta = 10^{-8}$, $t = 10^{4}$.}
    \label{fig:multi-epoch}
\end{figure}

Figure \ref{fig:RDP-dom} provides a clear example of the advantage of direct analysis, in regimes where the privacy guarantees of the random allocation scheme are better than those of the Poisson scheme. Even though RDP-based bounds on Poisson are not as tight, the RDP-based of allocation is superior to other methods that rely on reduction to Poisson. On the other hand, figure \ref{fig:multi-epoch} indicates that the gap between our RDP-based bound and Poisson's PLD-based one in other regimes, is mainly due to the fact it relies on RDP, and not a property of random allocation. It additionally reflects the fact that the gap between PLD and RDP based analysis vanishes as the number of epochs grows.

\subsection{Implementation details}\label{apd:RDPimp}
Computation time of the naive implementation of our RDP calculation ranges between second and minutes on a typical personal computer, depending on the $\alpha$ value and other parameters, but can be improved by several orders of magnitude using several programming and analytic steps which we briefly discuss here.

On the programming side, we used vectorization and hashing to reduce runtime. To avoid overflow we computed most quantities in log form, and used and the LSE trick. While significantly reducing the runtime, programming improvements cannot escape the inevitable exponential (in $\alpha$) nature of this method. Luckily, in most settings, $\alpha^{*}$ - the $\alpha$ value which induces the tightest bound on $\eps$ is typically in the low $10$s. Unfortunately, finding $\alpha^{*}$ requires computing $\Renyi{\alpha}$, so reducing the range of $\alpha$ values for which $\Renyi{\alpha}$ is crucial.

We do so by proving an upper bound on $\alpha^{*}$ in terms of a known bound on $\eps$.
\begin{claim}
    Given $\delta \in (0,1)$ and two distributions $P, Q$ and, denote by $\varepsilon(\delta) \coloneqq \underset{x > 0}{\inf}(\delta(x) < \delta)$. Given $\eps > 0$, if $\varepsilon(\delta) \le \eps$ and $\RenyiDiv{\alpha}{P}{Q} > \eps$, then $\alpha^{*} < \alpha$.
\end{claim}
A direct implication of this Lemma is that searching on monotonically increasing values of $\alpha$ and using the best bound on $\eps$ achieved at any point to check the relevancy of $\alpha$, we don't have to compute many values of $\alpha$ greater than $\alpha^{*}$ before we stop.

\begin{proof}
    Denote by $\gamma_{\delta}(\alpha)$ the bound on $\eps$ achieved using $\RenyiDiv{\alpha}{P}{Q}$. From Lemma \ref{lem:RDPtoDP}, $\gamma_{\delta}(\alpha) = \RenyiDiv{\alpha}{P}{Q} + \phi(\alpha)$ for a non negative $\phi$ (except for the range $\alpha > 1/(2 \delta)$ which provides a vacuous bound). Since $\RenyiDiv{\alpha}{P}{Q}$ is monotonically non-decreasing in $\alpha$ we have for any $\alpha' \ge \alpha$, 
    \[
        \gamma_{\delta}(\alpha') \ge \RenyiDiv{\alpha'}{P}{Q} \ge \RenyiDiv{\alpha}{P}{Q} \ge \eps,
    \]
    so it cannot provide a better bound on $\alpha$.
\end{proof}

\section{Comparison to other techniques}
For completeness, we state how one can directly estimate the hockey-stick divergence of the entire random allocation scheme. This technique was first presented in the context of the Gaussian mechanism by \cite{CGHLKKMSZ24}.

We first provide an exact expression for the privacy profile of the random allocation scheme.
\begin{lemma}\label{lem:exctGauss}
    For any randomizer $\rand$ and $\eps > 0$ we have, 
    \[
        \privProf{\alloc{t}{\rand}}(\eps) = \expect{\viewRV \sim \rand^{\otimes t}(\bot)}{\left[\frac{1}{t} \sum_{i \in [t]}e^{\loss{\outRV_{i}}{*}{\bot}} - e^{\eps}  \right]_{+}}.
    \]
    
    Given $\sigma > 0$, if $N_{\sigma}$ is a Gaussian mechanism with noise scale $\sigma$ we have, 
    \[
        \privProf{\alloc{t}{N_{\sigma}}}(\eps) = \expect{\boldsymbol{Z} \sim \mathcal{N}(\bar{0}, \sigma^{2} I_{t})}{\left[\frac{1}{t} \sum_{i \in [t]} e^{\frac{2 Z_{i} - 1}{2 \sigma^{2}}} - e^{\eps} \right]_{+}}
    \]
\end{lemma}

We note that up to simple algebraic manipulations, this hockey-stick divergence is essentially the expectation of the right tail of the sum of $t$ independent log-normal random variables, which can be approximated as a single log-normal random variable \citep{MWMZ07}, but this approximation typically provide useful guarantees only for large number of steps.

\begin{proof}
    Denote by $I$ the index of the selected allocation. Notice that for any $i \in [t]$ we have, 
    \[
        P_{\allocFunc{t}{\rand}{*}}(\view \vert I=i) = \left(\prod_{j=1}^{i-1} P_{\rand(\bot)}(\out_{j}) \right) P_{\rand(*)}(\out_{i}) \left(\prod_{j=1}^{i-1} P_{\rand(\bot)}(\out_{j}) \right) = P_{\allocFunc{t}{\rand}{\bot}}(\view) \cdot \frac{P_{\rand(*)}(\out_{i})}{P_{\rand(\bot)}(\out_{i})}
    \]

    \[
        \Rightarrow P_{\allocFunc{t}{\rand}{*}}(\view) = \frac{1}{t} \sum_{i \in [t]} P_{\allocFunc{t}{\rand}{*}}(\view \vert I=i) = \frac{1}{t} P_{\allocFunc{t}{\rand}{\bot}}(\view) \sum_{i \in [t]}  \frac{P_{\rand(*)}(\out_{i})}{P_{\rand(\bot)}(\out_{i})} 
    \]
    
    Using this identity we get, 
    \[
        \lossAlg{\alloc{t}{\rand}}{\view}{*}{\bot} = \ln \left(\frac{P_{\allocFunc{t}{\rand}{*}}(\view)}{P_{\allocFunc{t}{\rand}{\bot}}(\view)} \right) = \ln \left(\frac{1}{t} \sum_{i \in [t]} \frac{P_{\rand(*)}(\out_{i})}{P_{\rand(\bot)}(\out_{i})} \right) = \ln \left(\frac{1}{t} \sum_{i \in [t]} e^{\lossAlg{\rand}{\outRV_{i}}{*}{\bot}} \right).
    \]
    Plugging this into the definition of the hockey-stick divergence completes the proof of the first part.

    The second part is a direct result of the dominating pair for the random allocation scheme of the Gaussian mechanism (Claim \ref{clm:domPairGauss}).
\end{proof}

\subsection{Monte Carlo simulation \citet{CGHLKKMSZ24}}\label{apd:monte-carlo}
Using Monte Carlo simulation to estimate this quantity, is typically done using the $\expect{\omega \sim P}{\left[1 - \alpha e^{-\loss{\omega}{P}{Q}} \right]_{+}}$ representation of the hockey-stick divergence, so that numerical stability can be achieved by bounding the estimates quantity $\in [0,1]$.

\begin{figure}[H]
    \centering
    \includegraphics[width=0.8\linewidth]{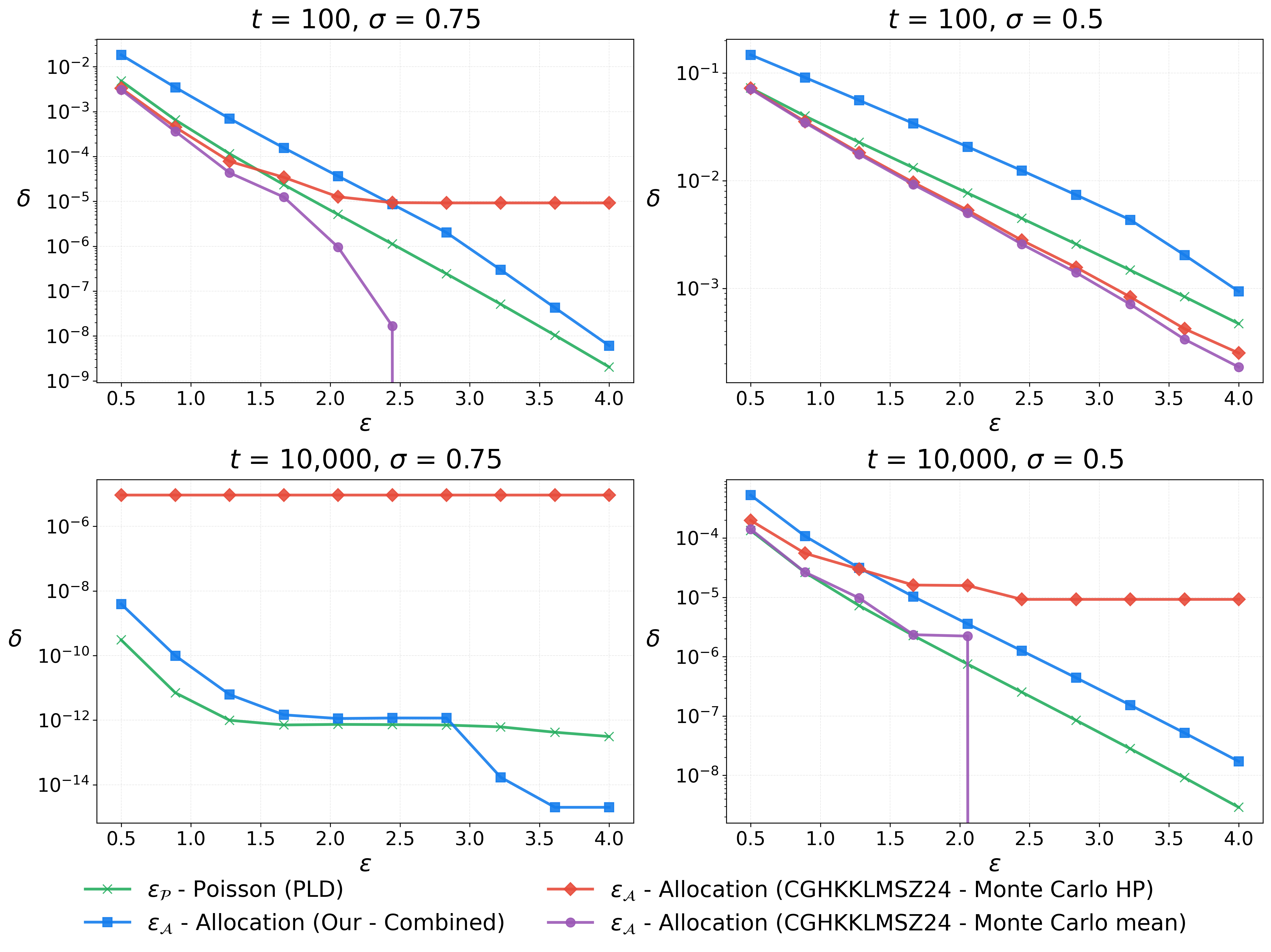}
    \caption{\small Comparison of $\delta$ bound for Poisson scheme with various bounds for the random allocation scheme, for several $\sigma$ and $t$ values; our combined methods, the high probability and the average estimations of using Monte Carlo simulation with order statistics, $5\cdot 10^{5}$ samples and $99\%$ confidence level, by \citet{CGHLKKMSZ24}.}
    \label{fig:MC-comp}
\end{figure}

A naive estimation will require an impractical number of experiments, especially in the low $\delta$ and high confidence level regimes. These challenges can be partially mitigated using importance sampling and order statistics, a new technique recently presented by \citet{CGHLKKMSZ24}. Still, this technique suffers from several limitations. It can only account for the setting of $k=1$ and does not provide a full PLD, and so cannot be composed. It can only estimate $\delta$, so plotting $\eps$ as a function of some other parameter is computationally prohibitive. Figure \ref{fig:MC-comp} illustrates simultaneously the tightness of our bounds, which are within a constant from the lower bound in the delta regime, and the limitations of the MC methods which become loose in the $\delta < 10^{-4}$ regime for the chosen parameters.

We additionally repeat the analysis using the experimental setting presented in \citet[Figure 2]{CGHLKKMSZ24}, both in the form of Figure \ref{fig:main} and Figure \ref{fig:MC-comp}. The choice of $t$ depends on the experimental settings, the dataset (Criteo pCTR or Criteo search conversion logs) and the batch size.

\begin{figure}[H]
  \centering
  \includegraphics[width=0.8\linewidth]{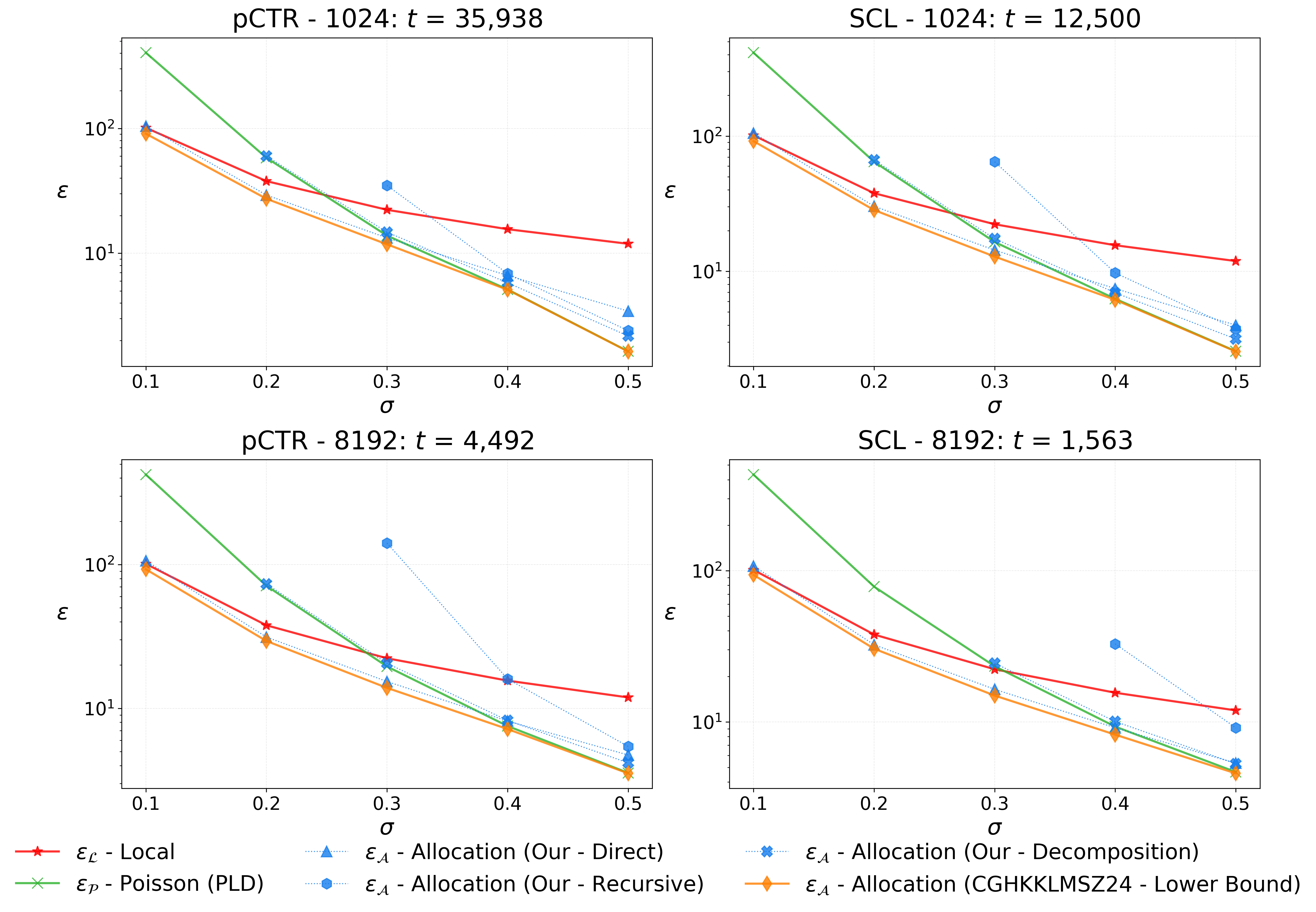}
  \caption{\small Upper bounds on privacy parameter $\eps$ as a function of the noise parameter $\sigma$ for various schemes and the local algorithm (no amplification), all using the Gaussian mechanism, with privacy parameters $\delta = 10^{-7}$ and various values of $t$, following the experimental parameters following the experimental settings of \citet[Figure 2]{CGHLKKMSZ24}. In the Poisson scheme $\lambda = 1/t$.}
  \label{fig:Chau_et_al_eps}
\end{figure}

\begin{figure}[H]
  \centering
  \includegraphics[width=0.8\linewidth]{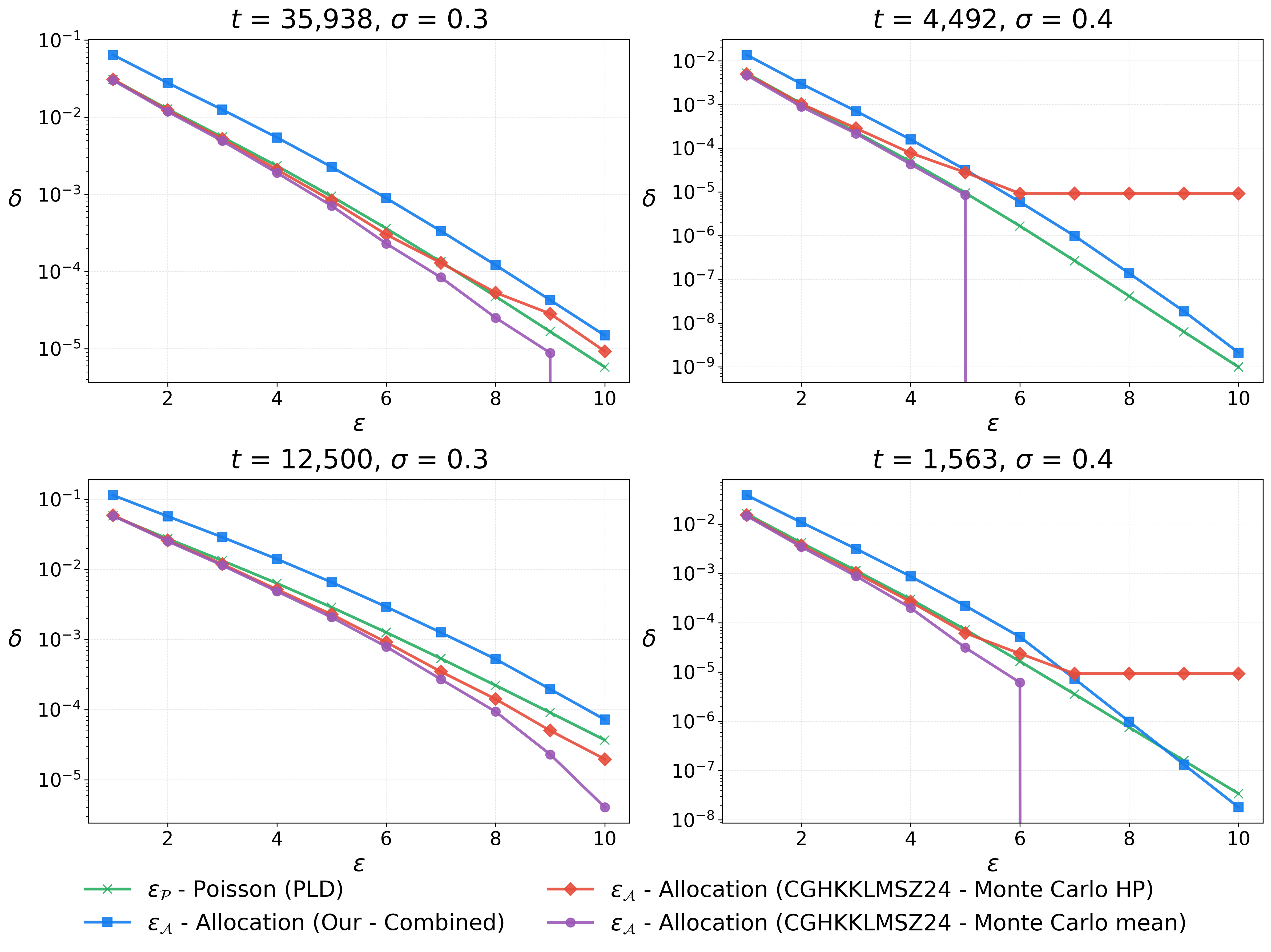}
  \caption{\small Comparison of $\delta$ bound for Poisson scheme with various bounds for the random allocation scheme, for several $\sigma$ and $t$ values; our combined methods, the high probability and the average estimations of using Monte Carlo simulation with order statistics, $5\cdot 10^{5}$ samples and $99\%$ confidence level, following the experimental settings of \citet[Figure 2]{CGHLKKMSZ24}.}
  \label{fig:Chau_et_al_delta}
\end{figure}

\subsection{RDP-based bound by \citet{DCO25}}\label{apd:comp_loose_RDP}
A recent independent work by \citet{DCO25} considered the same setting under the name Balanced Iteration Subsampling. In Theorem 3.1 they provide two RDP bounds for the remove direction and one for add, that are comparable to Theorem \ref{thm:dirRemBnd} in our work. Since the bound for the remove direction always dominated the add direction, we focus on it. The first one is tight but computationally expensive even for the case of $k=1$, as it sums over $O(t^{k \alpha})$ terms (in the case of $k=1$ their expression matches the one proposed by \citet{LT22}, which is mathematically identical to our, but requires $O(t^{\alpha})$ summands rather than our $O(2^{\alpha})$ ones.). The second bound they propose requires summing only over a linear (in $k$) number of terms which is significantly more efficient than our term, but is lossy. This gap is more pronounced in some parameter regimes, mainly when the $\alpha$ used for inducing the best bound on $\eps$ is large. On the other hand, this method allows for direct analysis of the $k > 1$ case, while our analysis relies on the reduction to composition of $k$ runs of the random allocation process with a selection of $1$ out of $t/k$ steps.

\begin{figure}[H]
    \centering
    \includegraphics[width=0.8\linewidth]{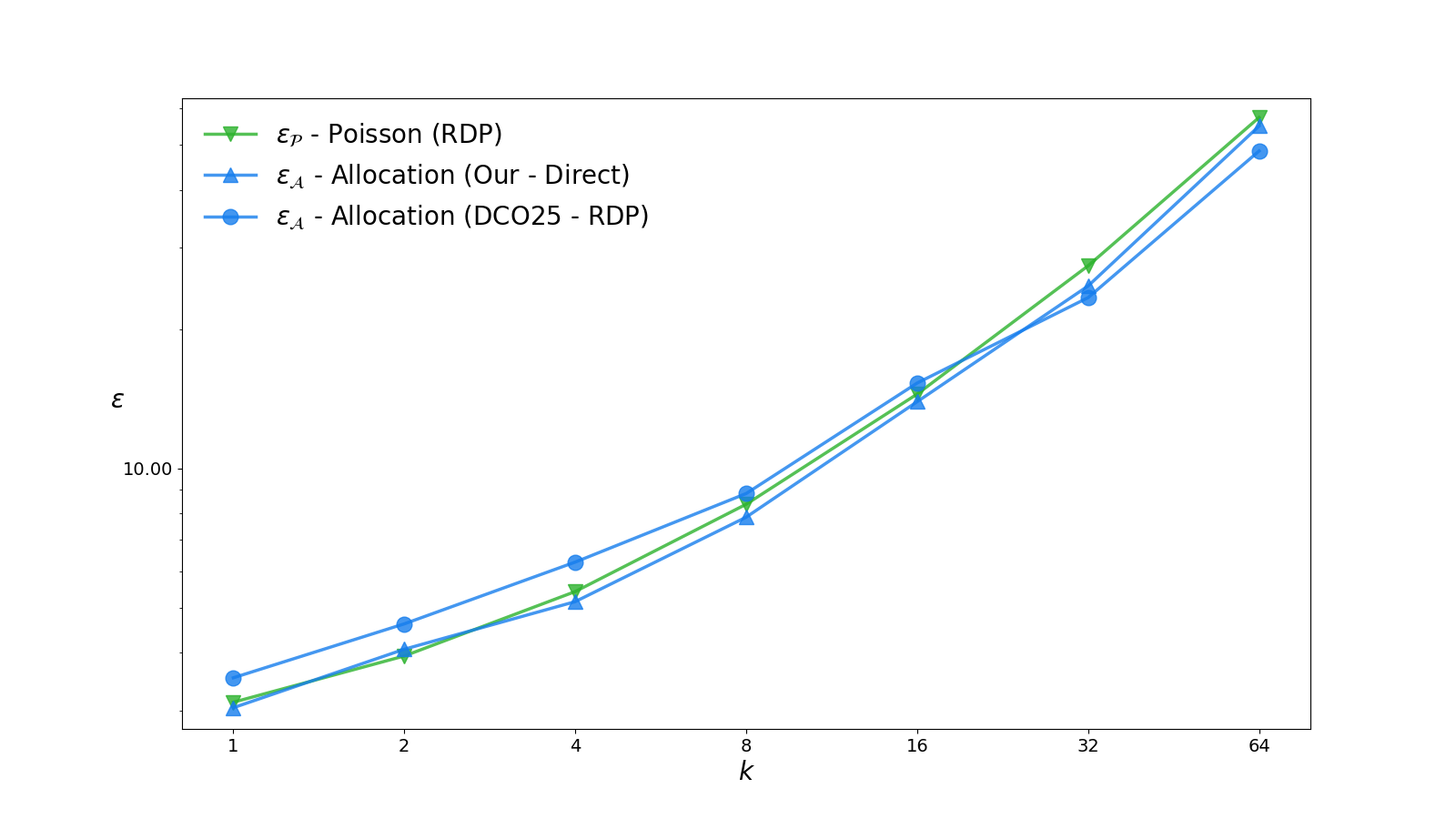}
    \caption{\small Upper bounds on privacy parameter $\eps$ as a function of the the number of allocations $k$ for the Poisson and random allocation schemes, all using the Gaussian mechanism with fixed parameters $\delta = 10^{-6}$, $t = 2^{10}$, $\sigma = 0.6$. The y-axis uses logarithmic scale to emphasize the relative performance.}
    \label{fig:DCO_comp}
\end{figure}

Figure \ref{fig:DCO_comp} depicts the spectrum of these effects. For small values of $k$, our RDP based bounds are tighter than the loose bound proposed by \citet{DCO25}, while for the large values of $k$ when $\eps$ is quite large our composition based analysis is looser.

\NeurIPS{\section{Privacy-utility tradeoff}\label{apd:util}}
\showstored{mov:util}

\ifconferencemode
\newpage
\section*{NeurIPS Paper Checklist}
\begin{enumerate}

\item {\bf Claims}
    \item[] Question: Do the main claims made in the abstract and introduction accurately reflect the paper's contributions and scope?
    \item[] Answer: \answerYes{}
    \item[] Justification: All claims are mathematically proven.

\item {\bf Limitations}
    \item[] Question: Does the paper discuss the limitations of the work performed by the authors?
    \item[] Answer: \answerYes{}
    \item[] Justification: We discuss both tightness of the results and computation limitations when applied.

\item {\bf Theory assumptions and proofs}
    \item[] Question: For each theoretical result, does the paper provide the full set of assumptions and a complete (and correct) proof?
    \item[] \answerYes{}
    \item[] Justification: All the claims are detailed and all the proofs appear in the appendices.

    \item {\bf Experimental result reproducibility}
    \item[] Question: Does the paper fully disclose all the information needed to reproduce the main experimental results of the paper to the extent that it affects the main claims and/or conclusions of the paper (regardless of whether the code and data are provided or not)?
    \item[] Answer: \answerYes{}
    \item[] Justification: The paper discusses the parameters and the supplementary material includes the code.

\item {\bf Open access to data and code}
    \item[] Question: Does the paper provide open access to the data and code, with sufficient instructions to faithfully reproduce the main experimental results, as described in supplemental material?
    \item[] Answer: \answerYes{}
    \item[] Justification: The code is provided as supplementary material. No data is involved.

\item {\bf Experimental setting/details}
    \item[] Question: Does the paper specify all the training and test details (e.g., data splits, hyperparameters, how they were chosen, type of optimizer, etc.) necessary to understand the results?
    \item[] Answer: or \answerNA{}
    \item[] Justification: The paper only contains numerical analysis without any usage of data. All relevant details were provided.

\item {\bf Experiment statistical significance}
    \item[] Question: Does the paper report error bars suitably and correctly defined or other appropriate information about the statistical significance of the experiments?
    \item[] Answer: \answerNA{}
    \item[] Justification: The paper mostly contains numerical analysis without any usage sampling. when MC based methods are considered, CI are provided.

\item {\bf Experiments compute resources}
    \item[] Question: For each experiment, does the paper provide sufficient information on the computer resources (type of compute workers, memory, time of execution) needed to reproduce the experiments?
    \item[] Answer: \answerNA{}
    \item[] The paper only contains numerical analysis, the longest of which runs for several minutes on a personal computer.
    
\item {\bf Code of ethics}
    \item[] Question: Does the research conducted in the paper conform, in every respect, with the NeurIPS Code of Ethics \url{https://neurips.cc/public/EthicsGuidelines}?
    \item[] Answer: \answerYes{}
    \item[] Justification: This is a purely theoretical work.

\newpage
\item {\bf Broader impacts}
    \item[] Question: Does the paper discuss both potential positive societal impacts and negative societal impacts of the work performed?
    \item[] Answer: \answerYes{}
    \item[] Justification: This paper provides several new tools for analyzing the privacy of machine learning algorithms. We do not anticipate any impacts beyond those typical for such results.
    
\item {\bf Safeguards}
    \item[] Question: Does the paper describe safeguards that have been put in place for responsible release of data or models that have a high risk for misuse (e.g., pretrained language models, image generators, or scraped datasets)?
    \item[] Answer: \answerNA{} 
    \item[] Justification: No data or models were released in this work.

\item {\bf Licenses for existing assets}
    \item[] Question: Are the creators or original owners of assets (e.g., code, data, models), used in the paper, properly credited and are the license and terms of use explicitly mentioned and properly respected?
    \item[] Answer: \answerYes{} 
    \item[] Justification: Details can be found in the README file.

\item {\bf New assets}
    \item[] Question: Are new assets introduced in the paper well documented and is the documentation provided alongside the assets?
    \item[] Answer: \answerNA{} 
    \item[] Justification: No new assets were introduced.

\item {\bf Crowdsourcing and research with human subjects}
    \item[] Question: For crowdsourcing experiments and research with human subjects, does the paper include the full text of instructions given to participants and screenshots, if applicable, as well as details about compensation (if any)? 
    \item[] Answer: \answerNA{} 
    \item[] Justification: No such thing was done in this paper.

\item {\bf Institutional review board (IRB) approvals or equivalent for research with human subjects}
    \item[] Question: Does the paper describe potential risks incurred by study participants, whether such risks were disclosed to the subjects, and whether Institutional Review Board (IRB) approvals (or an equivalent approval/review based on the requirements of your country or institution) were obtained?
    \item[] Answer: \answerNA{} 
    \item[] Justification:  No such thing was done in this paper.

\item {\bf Declaration of LLM usage}
    \item[] Question: Does the paper describe the usage of LLMs if it is an important, original, or non-standard component of the core methods in this research? Note that if the LLM is used only for writing, editing, or formatting purposes and does not impact the core methodology, scientific rigorousness, or originality of the research, declaration is not required.
    \item[] Answer: \answerNA{} 
    \item[] Justification: LLMs only assisted with some of the technical coding tasks.

\end{enumerate}
\fi
\end{document}